\documentclass{article} 

\PassOptionsToPackage{dvipsnames,table}{xcolor}
\usepackage{iclr2027_conference,times}
\newif\ificlrpreprint
\iclrfinalcopy
\iclrpreprinttrue

\usepackage[T1]{fontenc}
\usepackage[utf8]{inputenc}
\usepackage{latexsym}
\usepackage{microtype}

\usepackage{graphicx}
\usepackage{wrapfig}

\usepackage{amsmath}
\usepackage{amssymb}

\usepackage{booktabs}
\usepackage{longtable}
\usepackage{multirow}
\usepackage{makecell}

\usepackage{xcolor}

\usepackage{tikz}
\usetikzlibrary{backgrounds}
\usepackage{tcolorbox}
\tcbuselibrary{breakable}

\usepackage{titletoc} 

\usepackage[inline]{enumitem}

\usepackage{listings}
\lstdefinestyle{promptstyle}{
    basicstyle=\ttfamily\small,
    breaklines=true,
    breakatwhitespace=true,
    breakindent=0pt,
    breakautoindent=false,
    columns=fullflexible,
    keepspaces=true,
    showstringspaces=false,
    frame=single,
    framesep=4pt,
    xleftmargin=0pt,
    xrightmargin=0pt
}

\usepackage{url}
\usepackage{hyperref}
\definecolor{darkblue}{rgb}{0, 0, 0.5}
\hypersetup{colorlinks=true, citecolor=darkblue, linkcolor=darkblue, urlcolor=darkblue}

\newcommand{\splitcelltb}[3]{%
  \tikz[baseline=(X.base)]{
    \node[
      inner sep=0pt,
      outer sep=0pt,
      minimum width=2.8em,
      minimum height=1.6em,
      anchor=base
    ] (X) {#3};
    \begin{scope}[on background layer]
      \fill[#1] (X.south west) rectangle (X.east);
      \fill[#2] (X.west) rectangle (X.north east);
    \end{scope}
  }%
}

\newcommand{\Ours}{\textbf{\textsc{SimLife}}}
\newcommand{\OursBench}{\textbf{\textsc{SimLife-BP}}}

\title{\textsc{SimLife}: Pattern Understanding for \\Long-Horizon Human-Agent Partnership}

\author{%
\begin{tabular}[t]{@{}l*{4}{@{\hspace{1.5em}}l}@{}}
Run Peng$^1$ & Zinnia Nie$^1$ & Jing Ding$^1$ & Yinpei Dai$^1$ & Yichi Zhang$^{1}$\thanks{Work done while at the University of Michigan} \\
Zengqing Wu$^2$ & Yao Fu$^1$ & Ziqiao Ma$^{1}$\footnotemark[1] & Jiayuan Mao$^{3,4}$ & Joyce Chai$^1$
\end{tabular} \\
\normalsize
    $^1$University of Michigan\;
    $^2$Osaka University\;
    $^3$Amazon\;
    $^4$University of Pennsylvania \\
    {\tt\small \{roihn,chaijy\}@umich.edu}
}

\begin{document}

\maketitle

\ificlrfinal
  \ificlrpreprint
    \lhead{Preprint}
  \else
    \lhead{Published as a conference paper at ICLR 2027}
  \fi
\else
  \lhead{Under review as a conference paper at ICLR 2027}
\fi
\begin{abstract}
Understanding humans over long horizons requires agents to infer not only what people need in the moment, but also how routines form, why they repeat, and when they change. We introduce \Ours{}, a scalable platform for simulating long-term household life with rich visual observations, ground-truth action logs, and synthetic dialogues with audio. Built on \Ours{}, \OursBench{} evaluates long-context pattern understanding: the ability to infer latent behavioral rules from weeks or months of everyday observations. The benchmark contains 106 episodes averaging 15.49 hours and 38.57 in-game days, and 1,439 question-answer pairs. Each task probes direct, counterfactual, noisy, and inverse reasoning under different levels of rule hints. Evaluating frontier models and architectures, we find that current models often achieve surface-level prediction without comprehensive rule understanding, rely on frequency-based heuristics rather than if-then reasoning over evidence, and struggle to adapt when behavioral patterns change. These findings suggest that long-context pattern understanding remains a major bottleneck for future embodied agents, while \Ours{} opens a broader space for studying memory, personalization, adaptation, and planning in everyday human-AI interaction.\footnote{\url{https://roihn.github.io/SimLife}}
\end{abstract}
\section{Introduction}

Understanding humans’ implicit intents and motives has long been a fundamental challenge. In long-context settings, this challenge goes beyond interpreting a person’s immediate needs or beliefs~\citep{ma2023towards,peng2025communicationverificationllmagents}: an agent must also recognize recurring behaviors and infer the underlying patterns that govern them over time. Such longitudinal understanding is essential for agents that continuously interact with people across weeks or months of everyday life.

Long-term behavioral patterns cannot be understood from isolated observations alone. For example, inferring that \textit{Bob drinks coffee every weekday morning or after staying up late} requires an agent to form hypotheses (e.g., ``Bob may drink coffee every day.''), 
revisit relevant observations, and repeatedly verify or revise them as evidence accumulates. This makes pattern understanding a genuinely long-horizon reasoning problem, beyond retrieving and using a previously observed fact.

\begin{figure}[t]
    \centering
    \begin{minipage}[t]{0.48\textwidth}
    \vspace{0pt}
        \centering
        \includegraphics[width=0.7\linewidth]{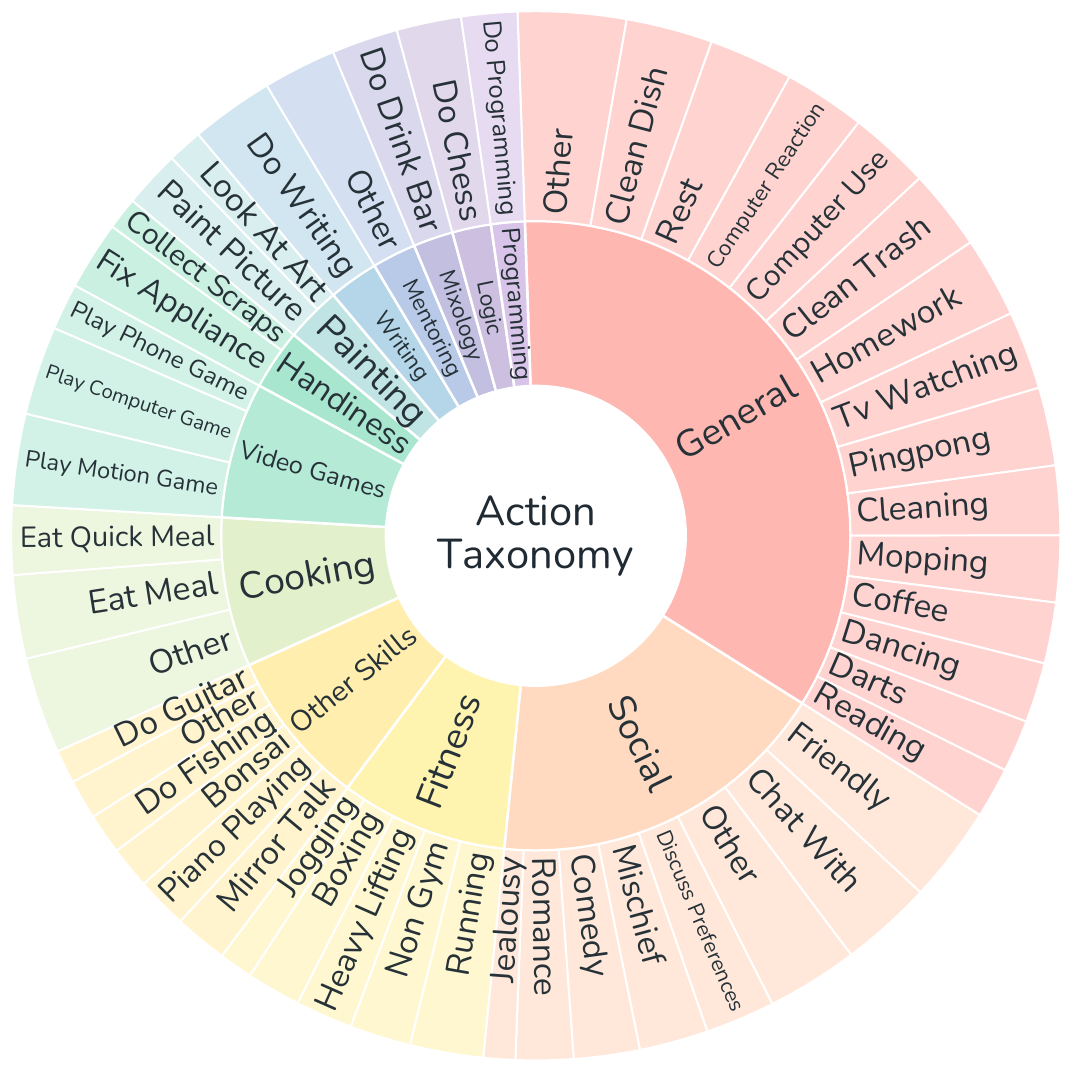}%
        \par
        \includegraphics[width=\linewidth]{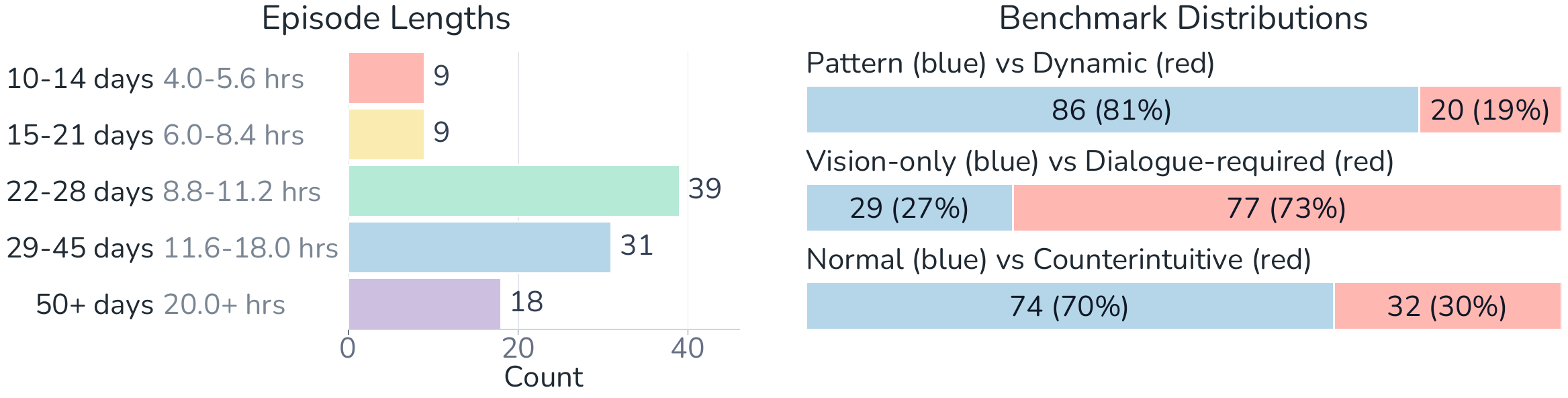}
        \vspace*{-15pt}
        \caption{
        \OursBench{} distribution overview.
        \textbf{Top:} action taxonomy with log-scaled counts.
        \textbf{Bottom-left:} episode lengths distribution, where one in-game day corresponds to 24 real minutes, or 0.4 real hours.
        \textbf{Bottom-right:} composition over three dimensions.
        }
        \label{fig:act_dist}
    \end{minipage}\hfill
    \begin{minipage}[t]{0.49\textwidth}
    \vspace{0pt}
        \centering
        \includegraphics[width=\linewidth]{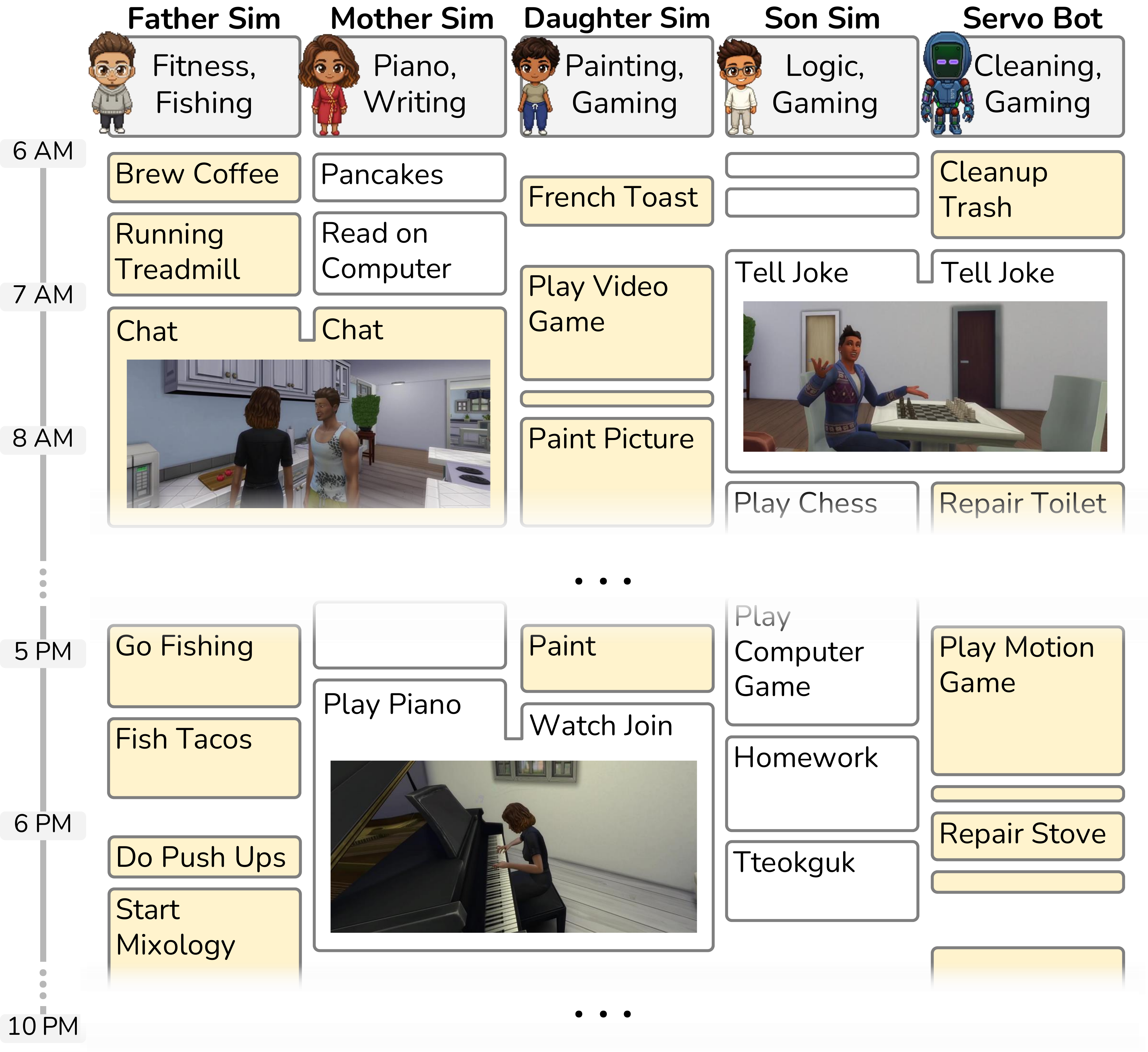}
        \vspace*{-11pt}
        \caption{Example simulated day in \Ours{}.
        Each column shows the 6 AM--10 PM schedule of one Sim, whose behavioral preferences are annotated above the timeline. Interactions between Sims are shown as shared blocks. Some time slots are collapsed for presentation.
        }
        \label{fig:dataset-example}
    \end{minipage}
    \vspace*{-8pt}
\end{figure}

To evaluate this form of long-context understanding under controlled yet realistic conditions, we need simulation environments that support extended timelines and rich everyday behaviors. Existing embodied simulation platforms provide valuable testbeds for embodied tasks~\citep{Puig_2018_CVPR, ALFRED20, puig2024habitat, ren2026simworldopenendedrealisticsimulator,zhou2026virtual}, but long-term household routines with diverse contextual cues remain difficult to simulate at scale. To address this gap, we propose \Ours{}, a scalable platform for simulating long-term everyday life in a home-like environment. Built on \textit{The Sims 4}\footnote{\url{https://www.ea.com/games/the-sims/the-sims-4} This research is not endorsed by or affiliated with Electronic Arts (EA) or its licensors.}, \Ours{} supports scripted simulations of multi-character household activities with rich visual observations, annotations, and diverse event categories (Figure~\ref{fig:act_dist}). We use this platform as a controllable foundation for constructing long-context benchmarks grounded in everyday behavior.

In this paper, we instantiate \Ours{} as \OursBench{}, a benchmark for behavioral pattern (BP) understanding: the ability to infer behavioral regularities, the rules behind them, and how these patterns evolve over time. \OursBench{} contains 106 ultra-long episodes spanning up to 48.23 hours and 120 in-game days, with 1,439 questions in total. Each episode follows a controlled, deterministic latent behavioral pattern governed by temporal or event-based rules, together with rich synchronized annotations for contextual grounding. To systematically evaluate pattern understanding, \OursBench{} combines \textit{four question types} that comprehensively probe pattern inference and application with \textit{three hint levels} that vary the amount of prior rule information. This design tests whether models merely recognize surface regularities or can infer and robustly apply the underlying behavioral rules across diverse contexts.

Our evaluation reveals a consistent picture. Models improve substantially as evidence becomes cleaner and more structured, yet even under favorable conditions, latent rule inference remains a major bottleneck. Our comprehensive question design further exposes limited robustness: models often succeed on easier forward predictions but fail to apply the same inferred pattern consistently across reasoning variants. Looking beyond final-task performance, we find that models frequently fall back on simple heuristics such as frequency counting rather than systematic if-then reasoning, and struggle to revise their understanding when the underlying pattern itself changes. Together, these results suggest that long-context pattern understanding requires more than remembering distant observations: agents must continually form, test, and revise hypotheses about behavior over time. More broadly, \Ours{} provides a foundation for studying this kind of persistent reasoning in long-horizon human-AI interaction, toward embodied agents that can build and adapt their understanding throughout extended periods of everyday life.

\vspace{-5pt}
\section{Related Work} 
\vspace{-5pt}
\paragraph{Long Context Learning}
Long-context learning has long been studied across application domains, from early work on document summarization and long-document understanding~\citep{luhn_summarization_1958, Edmundson1969NewMI, Salton_retrieval_1994} to modern long-context understanding~\citep{Beltagy2020Longformer, liu-etal-2024-lost, kamradt2023needle, pmlr-v235-jelassi24a}, dialogue and chat history modeling~\citep{bei2025memgallery}, agentic tool use~\citep{yao2023react, Acharya_agentic_survey_2025,fu2024autoguide}, multimodal understanding~\citep{long2026seeing, chen2026multimodallifelongunderstandingdataset, yan2025teleego}, and embodied AI with persistent memory~\citep{wang2024voyager, dai2026robomme}. Architecturally, this progress has been supported by increasingly large context windows~\citep{gemini152024, bai2025qwen3, peng2024yarn, gemma4}, test-time adaptation~\citep{ma2026fast, zhang2025test}, continual learning~\citep{deng2025unlocking}, and self-evolving agents~\citep{gao2026a}. Everyday assistants fall into this setting, yet the growing scale and complexity of long-context benchmarks raise a key question: what makes a task uniquely long-context, beyond simply requiring longer inputs? Echoing~\citet{hsieh2024ruler, bai-etal-2024-longbench, bai-etal-2025-longbenchv2}, \OursBench{} addresses this question by targeting intrinsically long-horizon reasoning rather than local retrieval over chunked segments, requiring models to integrate and cross-reference evidence scattered across days, weeks, and modalities to infer latent behavioral patterns.
\vspace{-5pt}

\paragraph{Long Video Understanding}

Long video understanding benchmarks have expanded across duration and modality, covering  hour-long QA~\citep{wu2024longvideobench}, extreme-length comprehension~\citep{wang2025lvbenchextremelongvideo}, and egocentric scenarios~\citep{Grauman_2022_CVPR, mangalam2023egoschema, yang2025egolifeegocentriclifeassistant, long2026seeing, chen2026multimodallifelongunderstandingdataset, yan2025teleego}. Yet many still operationalize long-video understanding as retrieval or localization. \OursBench{} instead targets intrinsically cross-temporal understanding, where answers require integrating implicit evidence across multiple days, modalities, and interactions. By combining continuous daily-life simulations with controlled scripts and sampled agent profiles, \OursBench{} enables controllable evaluation of latent behavioral pattern inference.
\vspace{-5pt}

\paragraph{Pattern Recognition}
Pattern recognition spans perceptual regularity detection, sequential pattern mining, and abstract rule discovery \citep{bongard1970pattern,klahr1988dual,josephson1996abductive,agrawal1995mining,srikant1996mining,mannila1997discovery,pei2001prefixspan}. 
Recent LLM benchmarks test rule induction from compact visual puzzles, symbolic examples, noisy observations, or programmatic input-output pairs \citep{chollet2019measure,chollet2025arc,hua2025inductionbench,li2025patterns, li2025mirage}. 
Long-context variants~\citep{yan2025mir,kuratov2024babilong} require aggregating many examples or scattered facts, but still use structured textual inputs. 
In contrast, \OursBench{} evaluates behavioral pattern recognition from unstructured ultra-long multimodal observations, requiring models to infer reusable latent rules from sparse, perceptually grounded, temporally distributed evidence.

\section{\Ours{} Benchmark}
\subsection{Platform \& Dataset}

\Ours{} is built on \textit{The Sims 4}, where we simulate multi-character household activities observed from the egocentric perspective of a serving robot called \textit{Servo Bot}. Each simulated day spans 6 AM to 10 PM in-game, corresponding to approximately 24 minutes in real time. 
Figure~\ref{fig:dataset-example} illustrates one example simulated day.
Characters (or Sims) are initialized with behavioral profiles specifying probabilistic activity preferences, which shape their daily routines and interactions. We then sample 466 in-game simulated days as the building blocks of our dataset. Each ultra-long video instance is constructed by concatenating a sequence of these day-level clips. We use a full simulated day as the basic unit because it forms a complete behavioral cycle, allowing clips to be recombined without introducing inconsistencies within a character's daily behavior.



Each recorded day is further enriched with fine-grained, synchronized annotations. We collect ground-truth game logs filtered to Servo Bot's field of view and introduce context-oriented dialogues, ranging from natural everyday conversations to utterances that reference off-screen events or evolving situations. These dialogues are further rendered as spatialized conversational audio using Sim-specific voices, allowing the observer to distinguish both who is speaking and the speaker's relative distance. Together, these components provide rich, temporally aligned observations for studying long-horizon understanding in simulated daily life.

\begin{figure*}
    \centering
    \includegraphics[width=\linewidth]{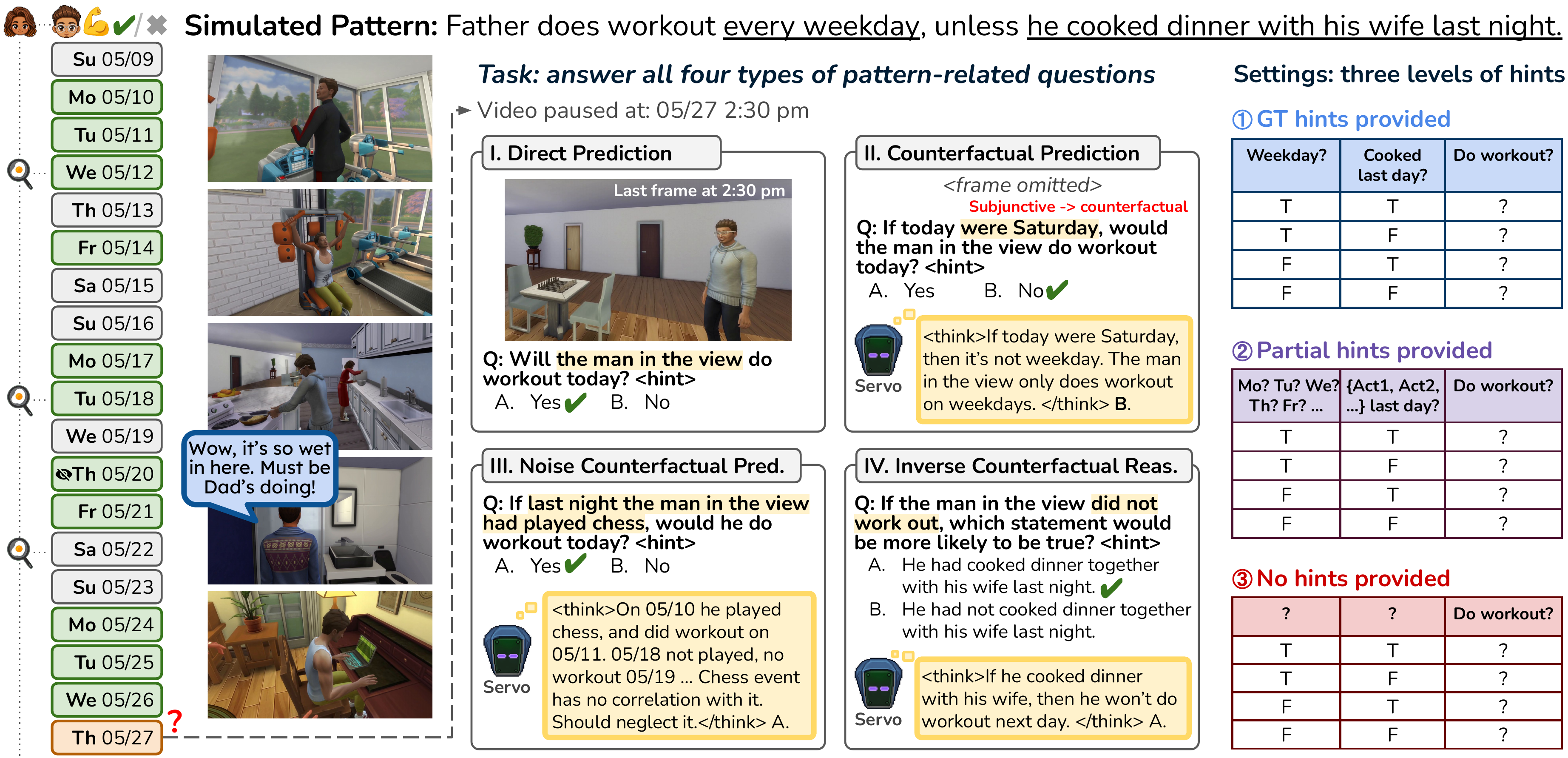}
    \vspace{-15pt}
    \caption{One example task of \OursBench{} benchmark. Given the Father Sim’s workout routine (shown at the top), we simulate three weeks of daily life that reflect this underlying pattern, recorded from the egocentric view of the Servo Bot. After sufficient observation, models are evaluated using four types of questions designed to assess their consistency in pattern understanding. We further provide three levels of hints that systematically ablate the skills required for long-term reasoning.} 
    \vspace{-10pt}
    \label{fig:example}
\end{figure*}


\subsection{Benchmark Design}
\label{sec:benchmark_design}

\paragraph{Pattern Definition}
We formulate the task as inferring a simulated character's behavioral pattern from ultra-long observations. Each pattern specifies whether and when a character performs a target activity under a set of temporal or event-based conditions. For example, the pattern \textit{``Bob drinks coffee on weekdays or after staying up late''} links coffee drinking to the day of the week and a prior-night event. To instantiate such a pattern, we filter and link simulated days into a long video sequence that repeatedly exhibits evidence of the underlying rule. The model must aggregate distributed observations to recover the rule governing the character's behavior. See Appendix~\ref{sec:appendix-phenomena} for details.

Our simulated patterns are controlled and deterministic once their underlying conditions are met. Under the rule above, for example, Bob drinks coffee on every weekday and always drinks coffee after staying up late, without randomly skipping the activity when either condition holds. Conditioning events can themselves occur stochastically at an independently specified frequency, but once such an event occurs, its effect on the target behavior is deterministic.

Note that behavioral patterns can be represented in many different forms, including graphs, procedures, policies, and probabilistic models. Within the scope of this work, we represent patterns using symbolic rules in the form of if–then propositions (e.g., if X and Y, then A). We choose this representation because such rules are highly interpretable and provide an explicit characterization of the agent's understanding of behavioral regularities. We view this formulation as an initial step rather than a restriction of the framework. Building on our simulation platform and data-generation pipeline, future work can explore alternative and potentially richer representations of behavioral patterns.
\vspace{-8pt}


\paragraph{Task Design}
We evaluate models using grouped question answering in a streaming setting. After exposing the model to a sufficiently long video history of in-house activities, we pause the video and instantiate a \textbf{task}, defined as a group of four binary multiple-choice questions\footnote{In special cases (\texttt{Hist} task in Section~\ref{sec:rq-pattern-adaptation}), there are no direct prediction questions.} about the behavioral pattern at that point. The four questions probe complementary forms of pattern understanding:

\begin{enumerate}[leftmargin=*, itemsep=3pt, parsep=0pt, topsep=0pt, partopsep=0pt]
\item \textbf{Direct Prediction:} Predict the most likely behavior given the current conditions.
\item \textbf{Counterfactual Prediction:} Predict the most likely behavior under a hypothetical change in conditions.
\item \textbf{Noisy Counterfactual Prediction:} Predict the most likely behavior under hypothetical conditions that include irrelevant information.
\item \textbf{Inverse Counterfactual Reasoning:} Infer the most plausible condition or cause given a counterfactual behavior.
\end{enumerate}

Each behavioral pattern yields multiple tasks, inserted at query points chosen to cover different combinations of the pattern conditions. In the coffee example, we query the model on days corresponding to different combinations of weekday/weekend and with/without stayed-up-late. This allows us to test whether the model has learned the complete rule rather than a partial heuristic.

In addition, we evaluate each task under three hint levels that vary how much rule information is provided. In \textbf{Full Hints}, the relevant conditions are directly specified, e.g., \textit{``the activity is related to whether it is a weekday''}. In \textbf{Partial Hints}, only the abstract type of each condition is revealed, e.g., \textit{``the activity is related to the day of the week''}; this yields a closed hypothesis space while still requiring models to eliminate inconsistent rules from evidence. In \textbf{No Hints}, the model must infer the pattern entirely from observation, which is the most realistic, but unbounded setting  (see Appendix~\ref{sec:appendix-questions}).

Finally, to avoid asking questions before sufficient evidence is available, we use the Partial Hints setting as a reference point for query timing. For example, if \textit{Bob plays ping-pong every Monday and Friday} and Partial Hints indicate that the pattern depends on the day of the week, observing one complete week is sufficient to determine his future behavior under our deterministic assumption. We call the earliest such point the \textbf{convergence point}.


For more complex conditions and different starting dates, the convergence point is less obvious. We therefore run a hypothesis-elimination procedure, starting from all candidate rules expressible over the condition space specified by Partial Hints and removing hypotheses inconsistent with each newly observed day. Formally, we define the convergence point as the earliest day after which all remaining hypotheses produce predictions consistent with the ground-truth rule for every subsequent day, and insert tasks only after this point. This ensures that each task is, in principle, answerable from the accumulated evidence with an unambiguous ground-truth prediction under the Partial Hints and Full Hints settings. While No Hints more closely reflects a realistic setting, it does not expose the ground-truth condition space required to define such a convergence point. We therefore view deterministic pattern understanding as a controlled prerequisite for studying more dynamic and probabilistic behavioral patterns. More details are provided in Appendix~\ref{app:elimination}.
\vspace{-5pt}

\paragraph{Benchmark Composition}
As is shown in Figure~\ref{fig:act_dist}, \OursBench{} contains 86 episodes for \textit{pattern understanding} and 20 episodes for \textit{dynamic understanding}. Pattern understanding episodes involve a single behavioral pattern, while dynamic understanding episodes involve multiple behavioral patterns and require models to identify distinct patterns as well as detect pattern shifts over time. Among them, 29 are solvable from vision alone, while 77 additionally require dialogue (audio).

To further probe contextual reasoning, 30\% of episodes follow \textit{counterintuitive patterns}, which associate activities with unusual or implausible conditions that cannot be inferred through common sense alone. For example, the pattern \textit{``drink coffee after playing the piano''} is individual-specific and unlikely to arise from prior general knowledge captured by large models. Models must therefore rely primarily on evidence from the video context rather than world priors when answering questions.

\section{Experiments \& Results}

We evaluate AI models' ability to understand behavioral patterns from ultra-long observations along three complementary dimensions: \textbf{whether models can infer such patterns, how their understanding develops as evidence accumulates, and how they adapt when patterns evolve.}

\subsection{RQ1: Can AI models infer patterns from ultra-long observations?} 
\label{sec:RQ1-overall}
\begin{table*}[t]
    \centering
    \setlength{\aboverulesep}{1pt}
    \setlength{\belowrulesep}{1pt}
    \setlength{\tabcolsep}{3pt}
    \definecolor{best}{HTML}{B7D8EF}
    \definecolor{second}{HTML}{E2EEF8}
    \definecolor{cD}{HTML}{0072B2}   
    \definecolor{cCF}{HTML}{E69F00}  
    \definecolor{cNC}{HTML}{009E73}  
    \definecolor{cIC}{HTML}{CC79A7}  
    
    \newcommand{\res}[2]{\ensuremath{#1^{\,\scalebox{0.7}{$\pm#2$}}}}
    \newcommand{\bars}[4]{%
    \begin{tikzpicture}[baseline=3pt, x=8pt, y=0.14pt]
      \fill[cD]  (0, 0) rectangle (0.92, #1);
      \fill[cCF] (1, 0) rectangle (1.92, #2);
      \fill[cNC] (2, 0) rectangle (2.92, #4);
      \fill[cIC] (3, 0) rectangle (3.92, #3);
      \draw[gray!40, line width=0.8pt] (-0.05, 50) -- (4.05, 50);
    \end{tikzpicture}}
    \caption{Main results on \OursBench{} (381 tasks / 1,439 questions per hint level). We report All-Correct rate (\texttt{AC}) and question-level accuracy (\texttt{Acc}) with 95\% binomial CI half-widths. Each \texttt{Sub\;Acc} cell shows per-question-type accuracy, left-to-right: \textcolor{cD}{Direct}, \textcolor{cCF}{Counterfactual}, \textcolor{cNC}{Noise Counterfactual}, \textcolor{cIC}{Inverse Counterfactual}; the gray line marks chance. Shading highlights within-row best variants, and \underline{\textbf{bold}} marks the overall best in each column.}
    \label{tab:main_results}
    \resizebox{\linewidth}{!}{%
    \begin{tabular}{@{}lcccccccccccc@{}}
    \toprule
    \multirow{2}{*}{Model} & \multirow{2}{*}{FPS} & \multirow{2}{*}{Thinking} &
    \multicolumn{3}{c}{Full Hints} & \multicolumn{3}{c}{Partial Hints} & \multicolumn{3}{c}{No Hints} \\
    \cmidrule(lr){4-6} \cmidrule(lr){7-9} \cmidrule(lr){10-12}
    & & & AC & Acc & Sub\,Acc & AC & Acc & Sub\,Acc & AC & Acc & Sub\,Acc \\
    \midrule
    \multicolumn{12}{@{}l}{\textit{Vision + Language (V+L)}} \\
    Qwen3-VL-4B-Instruct & 0.5 & CoT & \cellcolor{second}\res{10.5}{3.1} & \res{49.8}{2.6} &
  \bars{46.6}{55.1}{48.1}{48.6} & \cellcolor{best}\res{11.5}{3.2} & \cellcolor{second}\res{50.0}{2.6} &
  \bars{48.9}{52.8}{47.8}{50.4} & \res{9.4}{2.9} & \cellcolor{best}\res{50.3}{2.6} &
    \bars{50.5}{53.3}{47.8}{49.6} \\
    Qwen3-VL-8B-Instruct & 0.5 & CoT & \cellcolor{best}\res{12.6}{3.3} & \cellcolor{best}\res{50.1}{2.6} &
  \bars{46.3}{54.1}{51.1}{48.3} & \cellcolor{second}\res{9.7}{3.0} & \cellcolor{second}\res{48.6}{2.6} &
  \bars{44.3}{51.4}{48.9}{49.1} & \res{7.6}{2.7} & \res{48.0}{2.6} &
    \bars{48.2}{49.9}{45.7}{48.3} \\
    Qwen3-VL-32B-Instruct & 0.5 & CoT & \res{11.0}{3.1} & \res{51.9}{2.6} & \bars{54.7}{53.5}{49.2}{50.7} &
  \cellcolor{best}\res{13.6}{3.4} & \cellcolor{best}\res{53.2}{2.6} & \bars{53.1}{53.0}{53.8}{53.0} &
  \cellcolor{second}\res{12.1}{3.3} & \cellcolor{second}\res{53.0}{2.6} &
    \bars{51.8}{55.6}{51.4}{52.8} \\
    Qwen3.5-9B & 0.5 & R & \res{12.3}{3.3} & \res{48.9}{2.6} & \bars{47.2}{52.0}{45.9}{50.1} &
  \cellcolor{best}\res{14.2}{3.5} & \cellcolor{best}\res{51.4}{2.6} & \bars{51.1}{53.8}{48.4}{52.0} &
  \cellcolor{second}\res{13.4}{3.4} & \cellcolor{second}\res{51.0}{2.6} &
    \bars{52.4}{54.1}{47.0}{50.7} \\
    Qwen3.5-35B-A3B & 0.5 & R & \cellcolor{second}\res{13.1}{3.4} & \res{52.2}{2.6} & \bars{50.2}{56.7}{48.1}{53.3}
  & \cellcolor{best}\res{14.2}{3.5} & \cellcolor{best}\res{53.3}{2.6} & \bars{53.7}{55.6}{53.2}{50.7} &
  \res{12.3}{3.3} & \cellcolor{second}\res{52.5}{2.6} &
    \bars{53.4}{55.4}{53.5}{47.8} \\
    Gemma-4-26B-A4B-it & 0.5 & R & \cellcolor{second}\res{7.6}{2.7} & \cellcolor{second}\res{48.8}{2.6} &
  \bars{50.8}{47.8}{51.4}{45.7} & \res{7.6}{2.7} & \res{48.1}{2.6} & \bars{50.5}{50.9}{45.7}{45.7} &
  \cellcolor{best}\res{10.8}{3.1} & \cellcolor{best}\res{49.3}{2.6} &
    \bars{50.2}{53.0}{48.4}{45.7} \\
    Gemini-3-flash-preview & 0.5 & R & \cellcolor{best}\res{14.7}{3.6} & \res{53.0}{2.6} &
  \bars{53.7}{59.3}{46.5}{52.2} & \res{13.9}{3.5} & \cellcolor{second}\res{54.1}{2.6} &
  \bars{56.7}{57.2}{48.6}{54.3} & \cellcolor{second}\res{14.2}{3.5} & \cellcolor{best}\res{54.8}{2.6} &
    \bars{53.7}{56.2}{53.0}{55.9} \\
    VideoTree (Mixtral-8x7B) & 1 & CoT & \cellcolor{best}\res{11.0}{3.1} & \cellcolor{best}\res{54.0}{2.6} &
  \bars{45.9}{56.2}{57.8}{54.6} & \res{9.7}{3.0} & \res{51.7}{2.6} &
  \bars{46.3}{53.0}{50.0}{56.4} & \cellcolor{second}\res{10.8}{3.1} & \cellcolor{second}\res{52.6}{2.6} &
  \bars{49.8}{52.0}{53.5}{54.6} \\
    \midrule
    \multicolumn{12}{@{}l}{\textit{Vision + Audio (V+A)}} \\
    Qwen3-Omni-30B-A3B \small{+ Qwen3.5-9B} & 0.5 & R & \cellcolor{second}\res{10.8}{3.1} & \res{48.1}{2.6} &
  \bars{45.3}{50.9}{45.9}{49.6} & \cellcolor{best}\res{11.0}{3.1} & \cellcolor{best}\res{48.6}{2.6} &
  \bars{48.9}{49.9}{44.1}{51.7} & \res{10.8}{3.1} & \cellcolor{second}\res{48.6}{2.6} &
  \bars{47.6}{50.1}{48.9}{47.8} \\
    M3-Agent & 1 & R & \res{10.0}{3.0} & \res{53.4}{2.6} & \bars{46.3}{55.4}{60.0}{50.7} &
  \cellcolor{second}\res{10.2}{3.0} & \cellcolor{best}\res{55.0}{2.6} & \bars{49.2}{57.0}{58.4}{54.3} &
  \cellcolor{best}\res{12.1}{3.3} & \cellcolor{second}\res{54.6}{2.6} & \bars{48.2}{55.9}{57.0}{55.9} \\
    \midrule
    \multicolumn{12}{@{}l}{\textit{Language Only (L)}} \\
    Qwen3.5-9B & -- & R & \res{9.2}{2.9} & \res{50.9}{2.6} & \bars{46.9}{62.5}{48.1}{45.4} &
  \cellcolor{second}\res{13.6}{3.4} & \cellcolor{best}\res{54.6}{2.6} & \bars{55.0}{61.4}{48.1}{53.8} &
  \cellcolor{best}\res{14.2}{3.5} & \cellcolor{second}\res{54.6}{2.6} &
    \bars{50.2}{61.9}{51.6}{53.8} \\
    Qwen3.5-35B-A3B & -- & R & \cellcolor{second}\res{18.1}{3.9} & \cellcolor{best}\res{61.1}{2.5} &
  \bars{54.1}{75.1}{60.3}{53.5} & \cellcolor{best}\res{18.4}{3.9} & \cellcolor{second}\res{59.3}{2.5} &
  \bars{59.3}{71.4}{54.6}{51.7} & \res{16.0}{3.7} & \res{58.7}{2.5} &
    \bars{49.5}{74.0}{55.9}{53.5} \\
    Gemma-4-26B-A4B-it & -- & R & \cellcolor{best}\res{11.3}{3.2} & \res{53.9}{2.6} &
  \bars{52.8}{61.7}{53.0}{47.8} & \res{11.0}{3.1} & \cellcolor{second}\res{54.4}{2.6} &
  \bars{58.0}{55.6}{52.2}{52.5} & \cellcolor{second}\res{11.3}{3.2} & \cellcolor{best}\res{54.8}{2.6} &
  \bars{56.4}{62.2}{48.4}{52.2} \\
    Gemini-2.5-flash & -- & R & \cellcolor{second}\res{17.9}{3.8} & \cellcolor{second}\res{58.6}{2.6} &
  \bars{57.3}{69.8}{57.0}{49.9} & \cellcolor{best}\res{21.0}{4.1} & \cellcolor{best}\res{60.2}{2.5} &
  \bars{59.3}{70.1}{56.5}{54.6} & \res{16.3}{3.7} & \res{56.4}{2.6} &
  \bars{51.1}{68.5}{54.9}{50.1} \\
    Gemini-3-flash-preview & -- & R & \cellcolor{second}\res{27.8}{4.5} & \cellcolor{second}\res{69.2}{2.4} &
  \bars{72.3}{81.1}{61.4}{62.5} & \cellcolor{best}\res{\textbf{\underline{33.3}}}{4.7} &
  \cellcolor{best}\res{69.4}{2.4} & \bars{71.0}{81.1}{60.5}{64.8} & \res{27.8}{4.5} &
    \res{66.5}{2.4} & \bars{60.9}{81.1}{63.8}{59.1} \\
    Gemini-3.1-flash-lite & -- & R & \cellcolor{second}\res{20.7}{4.1} & \cellcolor{second}\res{59.9}{2.5} &
  \bars{59.9}{68.0}{54.1}{57.5} & \cellcolor{best}\res{21.0}{4.1} & \cellcolor{best}\res{61.2}{2.5} &
  \bars{57.0}{73.5}{57.8}{55.4} & \res{18.6}{3.9} & \res{59.1}{2.5} &
    \bars{52.4}{71.9}{56.8}{53.8} \\
    \midrule
    \multicolumn{12}{@{}l}{\textit{Language only, noiseless (L-N)}} \\
    Qwen3.5-9B & -- & R & \cellcolor{best}\res{23.6}{4.3} & \cellcolor{best}\res{65.5}{2.5} &
  \bars{70.4}{79.3}{50.5}{62.2} & \cellcolor{second}\res{18.4}{3.9} & \cellcolor{second}\res{61.2}{2.5} &
  \bars{64.5}{76.9}{46.2}{57.5} & \res{16.0}{3.7} & \res{56.5}{2.6} &
    \bars{59.0}{74.0}{41.9}{51.2} \\
    Qwen3.5-35B-A3B & -- & R & \cellcolor{best}\res{34.9}{4.8} & \cellcolor{best}\res{73.0}{2.3} &
  \bars{76.9}{84.0}{60.5}{70.9} & \cellcolor{second}\res{32.0}{4.7} &
  \cellcolor{second}\res{\textbf{\underline{70.3}}}{2.4} & \bars{71.7}{84.0}{58.9}{66.4} & \res{21.5}{4.1} &
    \res{62.7}{2.5} & \bars{61.2}{81.9}{50.8}{56.2} \\
    Gemma-4-26B-A4B-it & -- & R & \cellcolor{best}\res{30.2}{4.6} & \cellcolor{best}\res{72.1}{2.3} &
  \bars{82.4}{86.9}{55.7}{64.8} & \cellcolor{second}\res{26.0}{4.4} & \cellcolor{second}\res{68.4}{2.4} &
  \bars{75.6}{84.3}{52.2}{62.5} & \res{22.3}{4.2} & \res{64.1}{2.5} &
  \bars{69.4}{82.4}{54.9}{50.5} \\
    Gemma-4-31B-it & -- & R & \cellcolor{best}\res{\textbf{\underline{36.2}}}{4.8} &
  \cellcolor{best}\res{\textbf{\underline{74.6}}}{2.2} & \bars{83.4}{86.6}{54.9}{74.8} &
  \cellcolor{second}\res{30.2}{4.6} & \cellcolor{second}\res{69.6}{2.4} & \bars{75.2}{86.4}{51.1}{66.4} &
    \res{\textbf{\underline{29.4}}}{4.6} & \res{\textbf{\underline{67.6}}}{2.4} & \bars{72.6}{84.8}{51.9}{61.7} \\
    Gemini-2.5-flash & -- & R & \cellcolor{best}\res{31.5}{4.7} & \cellcolor{best}\res{71.8}{2.3} &
  \bars{80.8}{82.9}{52.2}{72.4} & \cellcolor{second}\res{22.3}{4.2} & \cellcolor{second}\res{65.3}{2.5} &
  \bars{70.0}{80.8}{46.8}{64.0} & \res{22.1}{4.2} & \res{63.7}{2.5} &
  \bars{64.8}{80.3}{52.4}{57.0} \\
    Gemini-3-flash-preview & -- & R & \cellcolor{second}\res{24.1}{4.3} & \cellcolor{best}\res{68.2}{2.4} &
  \bars{76.9}{81.1}{42.2}{73.5} & \cellcolor{best}\res{24.9}{4.3} & \cellcolor{second}\res{67.3}{2.4} &
  \bars{74.6}{80.1}{46.8}{68.5} & \res{23.1}{4.2} & \res{63.9}{2.5} &
    \bars{65.5}{80.3}{45.7}{63.8} \\
    \bottomrule
    \end{tabular}%
    }

    \vspace{-10pt}
  
  \end{table*}

We first assess the overall capability of models to recognize and reason about behavioral patterns accumulated over extended observation periods.

\subsubsection{Experimental Setup}

We designed experiments varying both the input modality and the degree of provided context. We evaluate models under four input conditions:
\vspace{-5pt}

\begin{itemize}[leftmargin=*, itemsep=2pt, parsep=0pt, topsep=0pt, partopsep=2pt]
    \item \textbf{Vision + Language (V+L):} Video frames (without audio) and dialogue transcripts.
    \item \textbf{Vision + Audio (V+A):} Video frames plus raw audio recordings of dialogue instead of transcripts.
    \item \textbf{Language Only (L):} Event logs (action \& dialogue) concatenated into a single input. This bypasses the vision process and uses ground-truth activities instead.
    \item \textbf{Language Only, noiseless (L-N):} Filtered logs with main and directly related events.
\end{itemize}
\vspace{-5pt}

\paragraph{Model Choice} We evaluate a broad set of models, including the Qwen3-VL family (4B, 8B, and 32B)~\citep{bai2025qwen3} with Chain-of-Thought (CoT) prompting; reasoning models with hybrid attention such as the Qwen3.5~\citep{qwen3.5} and Gemma4~\citep{gemma4} families, covering both dense and mixture-of-experts (MoE) variants; Qwen3-Omni~\citep{xu2025qwen3omnitechnicalreport} with an audio channel, and the Gemini~\citep{gemini3flashpreview, comanici2025gemini} family as representative proprietary models. 
For vision-based input conditions, we follow the Socratic Models framework~\citep{zeng2023socratic}: we divide each day's video into four clips of approximately six minutes each, generate a summary for each clip, and concatenate all summaries with the task questions. For language-only variants, we concatenate the logs and questions into a single input pass. Implementation details are provided in Appendices~\ref{sec:appendix:vision_implementation_details} and~\ref{sec:appendix:text_implementation_details}.

We also include representative agentic baselines designed for temporal search, i.e. VideoTree~\citep{Wang_2025_CVPR}, and memory management, i.e. M3-Agent~\citep{long2026seeing}. 
\vspace{-5pt}

\paragraph{Hint Levels} We evaluate models with three hint levels as introduced in Section~\ref{sec:benchmark_design}.
\vspace{-5pt}

\paragraph{Metrics} We report three primary metrics: \textit{All-Correct Rate} (\texttt{AC}), the proportion of tasks for which the model answers all sub-questions correctly, reflecting the consistency and completeness of its pattern understanding; \textit{Average Accuracy Rate} (\texttt{Acc}), the mean accuracy across all sub-questions within a task, as well as \textit{Sub-questions Accuracy} (\texttt{Sub Acc}), providing a finer-grained measure.


\subsubsection{Results}

\paragraph{Evidence Quality Drives Performance}
Table~\ref{tab:main_results} shows an overall trend of improved performance as inputs become cleaner and more structured. In the V+L setting, most models remain near chance on \texttt{Acc}, with only Gemini 3 Flash consistently performing above chance across hint levels. In contrast, models consistently perform above chance in L-N, where Gemma-4-31B-it reaches $36.2\%$ \texttt{AC} and $74.6\%$ \texttt{Acc} with Full Hints. The substantial improvement from V+L to L-N highlights the difficulty of extracting relevant evidence from long multimodal observations, while the remaining gap even with oracle-filtered inputs indicates that latent rule inference itself remains challenging.
\vspace{-5pt}


\paragraph{Structured Reasoning Emerges in L and L-N} 


Across hint levels, V+L shows no consistent benefit from stronger hints, while a signal of structured reasoning begins to emerge in L and becomes clearest in L-N, where Full Hints generally perform best. Reasoning-trace analysis helps explain this progression: Full Hints direct models toward the relevant variables, but using them effectively depends on whether the corresponding evidence can be recovered. In V+L, extracting the deciding evidence from long visual observations is already challenging, and it may be absent from the generated diaries; in L, the evidence is available but can be buried in noisy logs; in L-N, it is both complete and structured, allowing the hints to translate more reliably into correct rule use. Together, these results suggest that structured pattern reasoning begins to emerge once relevant evidence becomes accessible, but remains strongly constrained by evidence extraction and organization.
\vspace{-5pt}

\paragraph{All-Correct Rate (\texttt{AC}) Guards Against Shortcutting}
The per-question-type breakdown (\texttt{Sub Acc}) reveals a consistent asymmetry: normal counterfactual questions (CF) are much easier than inverse (IC) and noise-injected (NC) variants, and this gap widens as inputs become cleaner: under No Hints, the median CF-IC gap increases from +4.2\% in V+L to +16.2\% in L and +32.1\% in L-N, while IC accuracy changes little.
In addition, although Gemma-4-31B-it (L-N) achieves 86.6\% accuracy on CF questions, only 36.2\% of tasks are solved consistently across all question variants.
These results support the design of \texttt{AC}, which requires models to answer all variants consistently and therefore prevents high performance on easier forward questions alone from being mistaken for genuine pattern understanding. In this way, \texttt{AC} better distinguishes systematic rule inference from answers obtained through question-specific shortcuts, common-sense priors, or chance. We show the full results in Table~\ref{tab:main_results_full}.



\subsection{RQ2: How do AI models develop pattern understanding over time?}

Having examined models' overall pattern-understanding performance, we next investigate how this understanding develops as observations accumulate over time, and whether their evolving beliefs align with human judgments. Instead of focusing only on the final prediction, we probe how models incrementally update their beliefs about the underlying pattern across successive observations.

\subsubsection{Experimental Setup}
We evaluate four representative models (Gemini-2.5-flash, Gemini-3-flash-preview, Gemma-4-26B-A4B-it, and Qwen3.5-35B-A3B) under the \textit{Language Only}, \textit{Partial Hints} setting. 
For each day $d$ in an episode, the model predicts whether the target activity will occur, yielding a daily belief trajectory $P_d(\text{yes})$.
For models supporting token log-probabilities, we derive $P_d(\text{yes})$ by summing and normalizing the probability mass of \textit{yes}/\textit{no} variants at the answer token; for Gemini-3-flash-preview, which does not expose log-probabilities, we map its emitted yes/no answer to 1/0.
Rather than evaluating only the final prediction, this day-by-day trajectory captures how a model's belief shifts from uncertainty to confidence as evidence accumulates.

Since a model's internal decision process is not directly observable, we compare its trajectory against interpretable reference baselines as behavioral probes: if a model's predictions closely track a baseline, this suggests that it may be following a similar decision strategy. We consider five reference baselines. Two are shallow heuristics that do not perform rule inference: (B1) \textit{frequency-based}, predicting from the historical base rate, and (B2) \textit{last-day-based}, extrapolating from the previous day's outcome. Three are advanced references: (B3) \textit{ground-truth} oracle, (B4) \textit{human predictions} (Appendix~\ref{sec:appendix:human_study_details}), and (B5) \textit{hypothesis-elimination}, which prunes candidate rules against accumulated evidence (Appendix~\ref{app:elimination}).
\vspace{-5pt}

\paragraph{Metrics} We analyze the models' daily predicted probabilities for yes/no answers. We use the \textbf{Brier score}~\citep{1950brier} to measure the discrepancy between predicted probabilities $P_d(\mathrm{yes})$ and ground-truth outcomes $y_d$, computed as
\begin{equation*}
    \mathrm{Brier}=\frac{1}{D}\sum_{d=1}^{D}\left(P_d(\mathrm{yes})-y_d\right)^2,
    \vspace*{-4pt}
    \label{eq:brier}
\end{equation*}
where $y_d\in\{0,1\}$; lower scores indicate better probabilistic predictions. Accuracy (\texttt{Acc}), balanced accuracy (\texttt{B-Acc}), positive F1-score (\texttt{F1$^+$}), and averaged $P(\mathrm{yes})$ (\texttt{$\bar p_{\mathrm{yes}}$}) are also reported for reference. In addition, we report Spearman correlation between each model's daily prediction trajectory and the five reference baselines described above.


\subsubsection{Results}

Table~\ref{tab:trajectory-spearman} reports the pairwise Spearman correlations among the $P_d(\text{yes})$ trajectories across all sources, measuring how similarly each trajectory evolves over the chain. For Qwen, Gemma, and Gemini 2.5 Flash, the strongest non-self correlation is with the frequency-based baseline (28.8\%, 32.2\%, 38.5\% respectively). These results suggest that, given partial hints, models tend to \textbf{default to simple heuristics, like counting instances of the target activity, rather than evaluating candidate rules against the accumulated evidence.} 

By construction, the hypothesis-elimination baseline captures a sequential process of pruning candidate rules against observed evidence. Human predictions are strongly correlated with this baseline (78.1\%), suggesting that it captures an important aspect of how humans update their beliefs over time. Against this reference, Gemini 3 Flash stands out: its closest non-self baseline is hypothesis elimination (40.5\%), and it also shows the strongest alignment with human predictions (46.7\%). Together, these results suggest that \textbf{Gemini 3 Flash more closely tracks hypothesis-elimination behavior and produces more human-like trajectories over the observation chain}. In contrast, the other models appear more strongly anchored by frequency and recency heuristics, with weaker evidence of systematic hypothesis elimination.

\vspace{-5pt}
\begin{figure*}[!t]
    \centering
    \includegraphics[width=\linewidth]{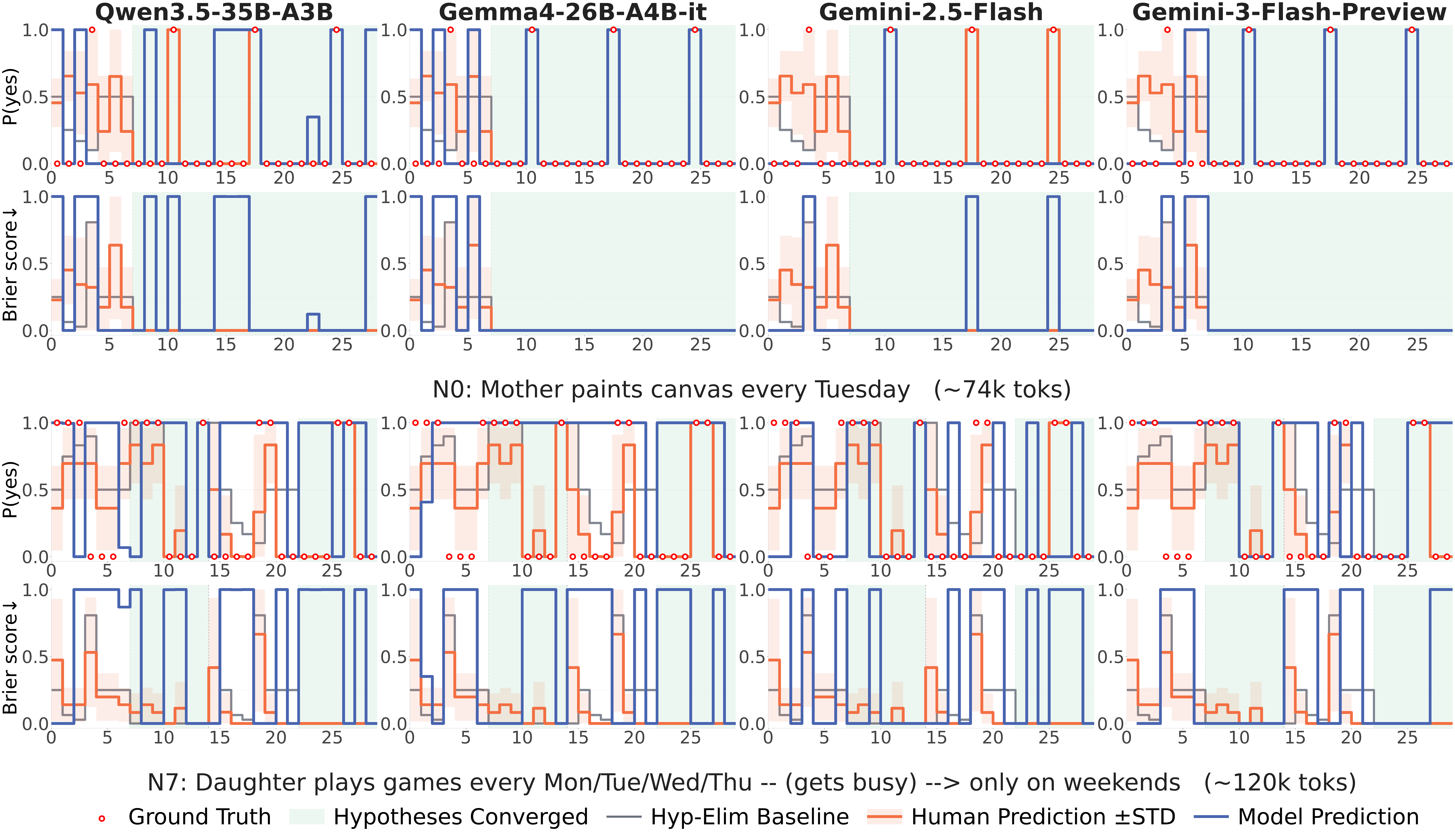}
    \vspace{-20pt}
    \definecolor{hypelimbg}{HTML}{EDF8F2}
    \caption{%
  Per-day trajectories of model predictions and human judgments. 
  Bottom rows show the per-day Brier score $\left(P_d(\mathrm{yes})-y_d\right)^2$. 
  A rapid convergence of Brier score of model prediction toward zero after \colorbox{hypelimbg}{hypothesis elimination} indicates successful pattern understanding and retention.%
}
\vspace{-18pt}
    \label{fig:trajectory_probing_curves}
\end{figure*}

\begin{table*}[!t]
\centering
\setlength{\aboverulesep}{1pt}
\setlength{\belowrulesep}{1pt}
\setlength{\tabcolsep}{3pt}
\renewcommand{\arraystretch}{0.95}
\definecolor{best}{HTML}{B7D8EF}
\definecolor{second}{HTML}{E2EEF8}
\definecolor{humanbest}{HTML}{F4C088}
\definecolor{humansecond}{HTML}{FBE6CC}
\caption{Spearman rank correlation of $P_d(\text{yes})$ trajectories (\%). 
Columns [1]--[9] correspond to the indexed source rows in the leftmost column.
Blue/light blue: best/second-best non-self correlation per row ([1]--[5]). 
Orange/light orange: best/second-best per column ([5], Acc, B-Acc, F1$^+$, Brier).
}
\label{tab:trajectory-spearman}
\resizebox{\linewidth}{!}{%
\begin{tabular}{@{}l|ccccccccc|cccccc@{}}
\toprule
Source & [1] & [2] & [3] & [4] & [5] & [6] & [7] & [8] & [9] & $n$ & Acc & B-Acc & F1$^+$ & Brier$\downarrow$ & $\bar p_{\mathrm{yes}}$ \\
\midrule
\multicolumn{1}{@{}l|}{\textit{Language Only (L)}} & \multicolumn{9}{c|}{} & \multicolumn{6}{c@{}}{} \\
{[1]} Qwen3.5-35B-A3B & 100.0 & 22.0 & 20.4 & 21.4 & 15.2 & \cellcolor{best}28.8 & \cellcolor{second}26.2 & 19.6 & 23.7 & 4067 & 59.7 & 61.3 & 46.1 & 39.8 & 48.1 \\
{[2]} Gemma-4-26B-A4B-it & 22.0 & 100.0 & 18.8 & 31.4 & \cellcolor{humansecond}22.0 & \cellcolor{best}32.2 & 12.8 & 23.6 & \cellcolor{second}25.0 & 4017 & 64.0 & 64.0 & 48.7 & 36.0 & 43.5 \\
{[3]} Gemini-2.5-flash & 20.4 & 18.8 & 100.0 & 21.4 & 21.7 & \cellcolor{best}38.5 & 22.1 & 21.4 & \cellcolor{second}29.2 & 4068 & 71.6 & 59.6 & 39.0 & 28.4 & 20.1 \\
{[4]} Gemini-3-flash-preview & 21.4 & 31.4 & 21.4 & 100.0 & \splitcelltb{humanbest}{best}{46.7} & 30.4 & 9.6 & 38.8 & \cellcolor{second}40.5 & 4053 & \cellcolor{humansecond}74.0 & \cellcolor{humansecond}70.7 & \cellcolor{humansecond}56.5 & \cellcolor{humansecond}26.0 & 33.4 \\
\multicolumn{1}{@{}l|}{\textit{Human Study}} & \multicolumn{9}{c|}{} & \multicolumn{6}{c@{}}{} \\
{[5]} Human-prediction & 15.2 & 22.0 & 21.7 & 46.7 & 100.0 & 17.6 & 4.8 & \cellcolor{second}74.3 & \cellcolor{best}78.1 & 194$\times$3 & \cellcolor{humanbest}84.5 & \cellcolor{humanbest}83.3 & \cellcolor{humanbest}79.2 & \cellcolor{humanbest}9.8 & 39.4 \\
\multicolumn{1}{@{}l|}{\textit{Reference Baselines}} & \multicolumn{9}{c|}{} & \multicolumn{6}{c@{}}{} \\
{[6]} Frequency-based & 28.8 & 32.2 & 38.5 & 30.4 & 17.6 & 100.0 & 52.0 & 37.0 & 50.2 & 4068 & 73.0 & 63.3 & 45.6 & 17.0 & 29.2 \\
{[7]} Last-day-based & 26.2 & 12.8 & 22.1 & 9.6 & 4.8 & 52.0 & 100.0 & 12.7 & 19.3 & 4068 & 65.3 & 56.5 & 36.7 & 33.7 & 26.9 \\
{[8]} Ground-truth-based & 19.6 & 23.6 & 21.4 & 38.8 & 74.3 & 37.0 & 12.7 & 100.0 & 79.2 & 4068 & 100.0 & 100.0 & 100.0 & 0.0 & 26.6 \\
{[9]} Hypothesis-elimination-based & 23.7 & 25.0 & 29.2 & 40.5 & 78.1 & 50.2 & 19.3 & 79.2 & 100.0 & 4068 & 88.7 & 88.0 & 80.2 & 6.5 & 27.7 \\
\bottomrule
\end{tabular}%
}

\vspace{-10pt}
\end{table*}

\paragraph{Representative Examples} Figure~\ref{fig:trajectory_probing_curves} unpacks the aggregate correlations on two chains selected from the human study (see Appendix~\ref{sec:appendix:human_study_details}). The first case N0 (\textit{``Mother paints every Tuesday''}) is the easiest pattern that hypothesis-elimination resolves within the first week (green area). Both Gemini 3 Flash and Gemma stay close to the human curve here, while Gemini 2.5 Flash picks up the signal early on but reverts to ``no'' for the end of the chain. 

The second case, N7 (\textit{``Daughter plays games every Mon/Tue/Wed/Thu, but then plays only on weekends after she gets busy''}), contains a pattern shift at day 14 (as the pivot). Humans and Gemini 3 Flash adapt quickly: Brier scores on post-pivot weekdays initially spike due to the old pattern, but they return to near zero within a week. In contrast, Gemini 2.5 Flash, Gemma, and Qwen remain locked onto the original rule and continue predicting “yes” on weekdays for the rest of the chain.\footnote{We note one fairness asymmetry: humans receive an explicit notice (“the behavior pattern may change after this point”), whereas models must infer the shift through dialogue, thus the post-pivot gap reflects both the scaffolding difference and a failure to recognize that dialogue indicates a change.}


Overall, Gemini 3 Flash is the one model whose reasoning traces consistently show explicit cycle detection and post-pivot dialogue use, matching its stronger correlation with both hypothesis-elimination baseline and human prediction. 





\subsection{RQ3: How do AI models adapt to evolving patterns?}
\label{sec:rq-pattern-adaptation}
Having examined how models update their understanding as evidence accumulates, we next ask how they respond when the \textbf{underlying pattern itself changes}. We examine whether models can detect such changes, adapt their understanding accordingly, and keep knowledge of prior patterns without forgetting.


\subsubsection{Experimental Setup}

We evaluate dynamic understanding across 20 episodes, each consisting of two sequential behavioral patterns, using three task types: \texttt{P1}, questions about Pattern 1 while active; \texttt{P2}, questions about Pattern 2 after pattern shift and sufficient exposure to the new pattern; and \texttt{Hist}, questions about the earlier Pattern 1 after observing Pattern 2. Representative results appear in Table~\ref{tab:rq3_adaptation_main}, using \texttt{Acc}.







\subsubsection{Results}
\label{sec:result-pattern-adaptation}

\begin{wraptable}{r}{0.48\linewidth}
  \vspace{-\intextsep}
  \setlength{\abovecaptionskip}{0pt}
  \centering
  \caption{Pattern adaptation of representative baselines (partial-hint, question-level accuracy in \%). Stars indicate task-level Mann-Whitney significance ($^{*}p\!<\!.05$, $^{**}p\!<\!.01$, $^{***}p\!<\!.001$). See full results in Table~\ref{tab:rq3_adaptation_full}.}
  \label{tab:rq3_adaptation_main}

  \vspace{2pt}
  \setlength{\tabcolsep}{1pt}
  \footnotesize
  
  \resizebox{\linewidth}{!}{%
  \begin{tabular}{@{}c@{\hspace{3pt}}lcccccc@{}}
  \toprule
  & Model & P1 & P2 & Hist & \makecell{P1-P2} & \makecell{Hist-P1} & \makecell{Hist-P2} \\
  \midrule
  \multirow{2}{*}{\rotatebox{90}{\scriptsize L}}
    & Qwen3.5-35B-A3B  & 63.4 & 48.7 & 65.3 & +14.8$^{*}$   & +1.8         & +16.6$^{*}$ \\
    & Gemini-3-flash   & 86.9 & 61.3 & 74.5 & +25.6$^{***}$ & -12.4$^{**}$ & +13.2$^{*}$ \\
  \midrule
  \multirow{2}{*}{\rotatebox{90}{\scriptsize L-N}}
    & Qwen3.5-35B-A3B  & 85.5 & 65.3 & 77.3 & +20.2$^{***}$ & -8.2$^{**}$  & +12.0 \\
    & Gemini-3-flash   & 79.3 & 65.3 & 76.4 & +14.0$^{*}$   & -2.9         & +11.1 \\
  \midrule
  \multirow{3}{*}{\rotatebox{90}{\scriptsize V/Agent}}
    & Gemini-3-flash (V+L)   & 57.9 & 56.7 & 55.6 & +1.3 & -2.4  & -1.1 \\
    & VideoTree (retrieval)  & 48.3 & 51.3 & 59.3 & -3.1 & +11.0 & +7.9 \\
    & M3-Agent (memory)      & 60.7 & 56.7 & 57.9 & +4.0 & -2.8  & +1.2 \\
  \bottomrule
  \end{tabular}%
  }
    \vspace{-10pt}

\end{wraptable}

Under L and L-N settings, \texttt{P2} is consistently weakest; three of the four L/L-N settings follow \texttt{P1} > \texttt{Hist} > \texttt{P2}.
The \texttt{P1}$\rightarrow$\texttt{P2} gap is significant for every model, reaching +25.6$\%$ ($p\!<\!.001$) for Gemini 3 Flash, showing that \textbf{models can identify a schedule, but struggle to update once it changes}. The limitation is online integration of the new pattern. This gap shrinks but does not close for stronger models (see Table~\ref{tab:rq3_adaptation_full}).

\vspace{-5pt}

\paragraph{\texttt{Hist} Retrieval Shortcut}
Notably, we expected \texttt{Hist} to be the hardest setting with the worst performance, because it requires retaining the earlier pattern while resisting interference from the current one; yet models generally perform better on \texttt{Hist} than \texttt{P2}. 
We attribute this to a shortcut our design intended to avoid but that persists in full-context models: because pre-switch evidence remains verbatim in the context, models can answer \texttt{Hist} through local pattern-matching rather than by reasoning about the pattern transition. 
Accuracy is consistent with this reading: \texttt{Hist} sits far above \texttt{P2}, yet stays below the clean \texttt{P1} (-12.4\%, $p\!<\!.01$ for Gemini 3 Flash, L) condition, likely because post-shift content dilutes but does not prevent retrieval. Temporal anchors were intentionally vague (e.g., ``roughly one month ago'') to discourage this shortcut, but wording had no effect on accuracy. 



\section{Conclusion \& Future Work}
We presented \Ours{}, a platform for simulating long-term everyday life, and \OursBench{}, a benchmark for evaluating pattern understanding from ultra-long multimodal observations. Our results show that current models often rely on surface heuristics, fail to apply inferred rules robustly and struggle to adapt to pattern shifts. These findings highlight long-context pattern understanding as an open challenge for embodied AI agents and \Ours{} provides a foundation for studying memory, personalization, adaptation, and long-horizon planning in everyday human-AI interaction. 

It is true that human behaviors are often probabilistic and may exhibit substantial variability and exceptions. However, as the goal of our research is to evaluate models' ability to infer behavioral patterns from observations, we opt to use deterministic rules to provide a well-defined ground truth for systematic and controlled evaluation. This design allows us to isolate the model's ability to discover the underlying behavioral regularities without introducing additional uncertainty about the behaviors themselves. We view this as a controlled starting point. Extending the framework to probabilistic patterns and behaviors with exceptions represents an important direction for future work.


\subsection*{AI use statement}

In this work, we used generative AI tools for generating synthetic data sets as part of our benchmark (text-to-audio). We have not used generative AI tools for: \textit{``design or provide feedback on research  methodology or experiments, implement methods, assist with translation, clean and reformat dataset, support qualitative and thematic data analysis, interpret results.''},
and \textit{``help develop theoretical models or conceptual frameworks, formulate mathematical claims, provide critical ingredients for proving mathematical claims, assist in the writing of proofs, propose or refine hypotheses''} are not applicable to this work.
Additionally, we used generative AI tools for creating and editing software code, summarizing existing literature,  brainstorming, and paper editing for better readability. We have reviewed all AI-assisted work. We take responsibility for the final content of this work, including text, claims or artifacts produced with the aid of generative AI.
\subsection*{Ethics statement}
\label{app:code_of_ethics}

\paragraph{Human-study Ethics}
An institutional review board determined that this study was exempt from ongoing review.
We recruited 12 adult English-reading participants through personal outreach, and compensated each participant \$20 for a study expected to take approximately one hour, following institutional participant-compensation guidelines.
Participants were shown a consent statement describing the study purpose, duration, task format, recorded data, and research use.
Only task responses, including symbolic inputs, slider responses, and text-based answers, were recorded.
We did not store personally identifiable information, collect detailed demographic or geographic attributes, or collect sensitive personal information.
Human-study results are reported only in aggregate, and individual response files are not publicly released.

\paragraph{Risks and Intended Use}
The study uses simulated household activity schedules rather than real personal behavior traces, and participants were not asked to disclose personal experiences, private routines, demographic attributes, or sensitive information.
We therefore consider the study to pose minimal risk to participants.

\paragraph{Dataset Collection}
Our dataset and benchmark are constructed from recorded gameplay in \textit{The Sims 4}, used as a controllable simulation environment for non-commercial academic research.
Data collection is controlled through custom scripts and community-developed modifications, with recordings captured from the perspective of an in-game robot agent and organized into video observations, event sequences, behavioral annotations, and learning targets.
The original game narrative, artistic presentation, character design, and gameplay experience are not the object of study, and no models are trained on the collected data.
Data collection is limited to simulated daily activity; the \textit{Discover University} downloadable content and community-developed modifications are used only to support and control in-game robot-agent behavior.             

\ificlrfinal
\subsubsection*{Acknowledgments}
We gratefully acknowledge the Google Cloud Research Credits Program for providing credits that supported the Gemini API usage in this work. We also thank Yutong Wang for valuable assistance during the preliminary exploration of this project.
\fi

\newpage
\bibliographystyle{iclr2027_conference}
\bibliography{custom}

\newpage
\appendix
\section*{Appendix}
\label{sec:appendix}

\startcontents[appendices]
\printcontents[appendices]{}{1}{}

\newpage

\section{Limitations}

While \Ours{} and \OursBench{} have comprehensively revealed models' pattern understanding ability, several limitations remain. First, the behavioral patterns in \OursBench{} are deterministic abstractions of real human behavior. Real-world habits are often stochastic, multi-factorial, and shaped by latent causes that may not be directly observable. We adopt deterministic rules not to claim that human behavior is rule-based, but to provide a controlled and verifiable setting in which pattern understanding can be evaluated with reliable ground truth. We view this as an intermediate step toward more realistic long-term human modeling, where the same core abilities: discovering regularities, ruling out inconsistent explanations, and adapting to change remain necessary.

At the same time, although \Ours{} provides rich multimodal observations
and controllable long-term behavior, simulated household routines cannot fully capture the variability, ambiguity, and social complexity of real human lives. An important next step is to study whether models trained or evaluated on \OursBench{} transfer to real-world long-term observation datasets, and whether simulated data can serve as useful supervision for building more adaptive assistants.

Additionally, our human study is not a perfectly symmetric comparison between humans and models. To reduce cognitive load, human participants are shown a calendar-like interface and simplified structured summaries rather than raw ultra-long logs or videos (see Appendix~\ref{sec:appendix:human_study_details} for more details). In dynamic-pattern settings, humans are also explicitly informed when a pattern may change, whereas models receive the corresponding signal only through the observation stream. These design choices make the human study more interpretable and feasible, but they also mean that human-model comparisons should be read as diagnostic rather than strictly controlled head-to-head evaluations.

\section{Controlled Generation of Behavioral Phenomena}
\label{sec:appendix-phenomena}

\Ours{} inverts the usual record-then-annotate pipeline into a controlled generation procedure: we specify a latent behavioral rule first, then assemble a long video that provably exhibits it by selecting and linking recordings from a finite pool of reusable pre-recorded days. This allows efficient benchmark construction with minimal annotation and filtering effort, and guarantees some extent of reproducibility.~\footnote{Reproducibility is bounded by the simulator. A deterministic environment (e.g. with a controllable random seed) gives full reproducibility; \textit{The Sims 4} does not — we can ensure Bob drinks coffee around 8 am but not the exact time, nor fine-grained details such as which hand holds the cup.}

The abstractions below build around one central object, the \textbf{Phenomenon}: a complete specification of a behavioral rule, and the blueprint every generated episode follows. Below it, an action, a condition, and a profile are the successive units that compose a phenomenon out of smaller parts. Above it, a schedule and a video chain are the successive steps that turn a phenomenon into an actual sequence of recordings. We walk through these six layers in that order, then trace one complete example end to end in \ref{sec:appendix-example}.

\subsection{Actions}
\label{sec:appendix-actions}


An \textbf{action} is the atomic unit of behavior in \Ours{}. Each action belongs to one of ten categories and can be performed by one or more characters (``Sims''), each in a designated \textbf{role} (e.g., \texttt{cook} or \texttt{doer}) that the matcher in \ref{sec:appendix-activity-spec} selects on. 


\paragraph{Characters.}
The household consists of five characters: \textit{Father Sim}, \textit{Mother Sim}, \textit{Son Sim}, \textit{Daughter Sim}, and \textit{Servo Bot} (a household robot). All the videos are recorded from the egocentric view of \textit{Servo Bot}. 

\paragraph{Action categories.}
Table~\ref{tab:action-categories} summarizes the ten action categories. Cooking recipes additionally carry dietary tags (\textit{lactose-free}, \textit{vegetarian}) that enable dietary-aware condition design.

\begin{table}[h]
\centering
\small
\resizebox{\columnwidth}{!}{
\begin{tabular}{llcl}
\toprule
\textbf{Category} & \textbf{Example Actions} & \textbf{Count} & \textbf{Collab.} \\
\midrule
Cooking   & \{\texttt{eggs\_and\_toast}, \texttt{spaghetti}, \texttt{butter\_chicken}\} & 33 & \checkmark \\
Piano     & \{\texttt{practice\_piano}, \texttt{play\_blues\_songs\_on\_piano}\} & 5  & -- \\
Guitar    & \{\texttt{practice\_guitar}, \texttt{play\_rock\_songs\_on\_guitar}\} & 5  & -- \\
Violin    & \{\texttt{practice\_violin}, \texttt{play\_country\_songs\_on\_violin}\} & 5  & -- \\
Fitness   & \{\texttt{endurance\_run}, \texttt{box}, \texttt{heavy\_lifting}, \texttt{jogging}\} & 5  & -- \\
Fishing   & \{\texttt{fish}, \texttt{fish\_with\_bait}\} & 2  & \checkmark \\
Writing   & \{\texttt{practice\_writing}, \texttt{write\_fantasy\_book}, \texttt{write\_biography}\} & 8  & -- \\
Video Games & \{\texttt{play\_sims\_forever}, \texttt{play\_blicblock}\}              & 5  & -- \\
Logic     & \{\texttt{play\_chess}\}                                                    & 1  & \checkmark \\
General   & \{\texttt{brush\_teeth}, \texttt{drink\_coffee}, \texttt{homework}, \texttt{watch\_tv}\} & 10 & Mixed \\
\bottomrule
\end{tabular}
}
\caption{Action categories in \Ours{}. ``Collab.'' indicates whether the action can involve multiple participants.}
\label{tab:action-categories}
\end{table}

\paragraph{Time periods.}
A simulated day spans from 6 AM to 10 PM.

\subsection{Conditions}
\label{sec:appendix-conditions}

A \textbf{condition} turns an abstract behavioral rule into per-day ground truth: it determines \emph{on which days} a target activity should or should not occur, and the resulting per-day labels are exactly what the benchmark later queries. Each condition targets an \emph{activity}, an action performed within a time window by a specified set of participants, which we denote $\mathcal{A}$ and formalize fully, as an \textbf{Activity Specification}, once it is needed for matching against recordings (\ref{sec:appendix-schedule}). Given a calendar of $N$ days, a condition produces a binary label vector $\mathbf{l} \in \{0,1\}^N$, where $l_d = 1$ indicates that the activity is required on day $d$, and $l_d = 0$ means the activity is required not to happen on day $d$.

There are two families of conditions, corresponding to the two ways a rule can be driven: \emph{temporal} conditions are driven by calendar structure (e.g., the day of the week), and \emph{activity-based} conditions are driven by another activity (e.g., an event the day before). Because real behavioral patterns are rarely a single rule but a conjunction, exclusion, or alternation of simpler ones, conditions targeting the same activity can be merged; each condition therefore carries an \texttt{operation} field governing how it combines with others (\ref{sec:appendix-merging}).

\subsubsection{Temporal Conditions}
\label{sec:appendix-temporal}

Temporal conditions determine activity occurrence based purely on time, frequency, or calendar structure (see Table~\ref{tab:temporal-conditions}). These conditions can be used to create structured daily life patterns that depend on the time of day. 

\begin{table}[h]
\centering
\small
\resizebox{\columnwidth}{!}{
\begin{tabular}{lp{5cm}p{7cm}}
\toprule
\textbf{Type} & \textbf{Template} & \textbf{Example} \\
\midrule
\texttt{every\_selected\_weekday}
  & On $\{$list of weekdays$\}$ 
  & ``Practice piano on Monday, Wednesday, Friday'' \\
\texttt{repeat\_every\_other}
  & Repeat $\{$count$\}$ times every other $\{$value$\}$ $\{$unit$\}$
  & ``Go fishing 1 time every other week'' \\
\texttt{on\_date}
  & On $\{$date\_expression$\}$
  & ``Write screenplay on the last Friday of each month'' \\
\bottomrule
\end{tabular}
}
\caption{Temporal condition types. Parameters in braces are configurable.}
\label{tab:temporal-conditions}
\end{table}

\paragraph{Duration units.}
The $\{$unit$\}$ parameter supports six values, each defining a window length for grouping calendar dates as is shown in Table~\ref{tab:duration-units}.

\begin{table}[h]
\centering
\begin{tabular}{lcl}
\toprule
\textbf{Unit} & \textbf{Length (days)} & \textbf{Scope} \\
\midrule
\texttt{day}     & 1   & Every day \\
\texttt{week}    & 7   & Calendar weeks \\
\texttt{month}   & 30  & Calendar months \\
\texttt{year}    & 365 & Calendar years \\
\texttt{weekday} & 5   & Monday--Friday only \\
\texttt{weekend} & 2   & Saturday--Sunday only \\
\bottomrule
\end{tabular}
\caption{Supported duration units and their effective lengths.}
\label{tab:duration-units}
\end{table}

\paragraph{Date expressions.}
The \texttt{on\_date} type supports patterns such as:
\begin{itemize}[leftmargin=*, itemsep=3pt, parsep=0pt, topsep=0pt, partopsep=3pt]
  \item \texttt{the \{n\}-th day of each month} (e.g., the 15th)
  \item \texttt{the \{n\}-th \{weekday\} of each month} (e.g., the 2nd Tuesday)
  \item \texttt{the last day of each month}
  \item \texttt{the last \{weekday\} of each month} (e.g., the last Friday)
\end{itemize}

\subsubsection{Activity-Based Conditions}
\label{sec:appendix-activity-conditions}

Activity-based conditions make the occurrence of a \emph{main activity} $\mathcal{A}_{\text{main}}$ depend on a \emph{trigger activity} $\mathcal{A}_{\text{trigger}}$ (see Table~\ref{tab:activity-conditions}). The trigger activity is pre-sampled independently across all calendar dates with a configurable frequency $f \in (0,1]$ (default $f = 0.1$). These conditions specify conditional behavioral patterns with activities that only occur when another event does. 

\begin{table}[h]
\centering
\small
\resizebox{\columnwidth}{!}{
\begin{tabular}{lp{5cm}p{7cm}}
\toprule
\textbf{Type} & \textbf{Template} & \textbf{Example} \\
\midrule
\texttt{right\_after}
  & $\mathcal{A}_m$ the day after $\mathcal{A}_t$
  & ``Drink coffee the day after a late-evening run'' \\
\texttt{within\_same\_day}
  & $\mathcal{A}_m$ same day as $\mathcal{A}_t$
  & ``Shower on the same day as heavy lifting'' \\
\texttt{no\_activity\_right\_after}
  & $\mathcal{A}_m$ must \emph{not} occur the day after $\mathcal{A}_t$
  & ``No video games the day after failing homework'' \\
\texttt{no\_activity\_within\_same\_day}
  & $\mathcal{A}_m$ must \emph{not} occur same day as $\mathcal{A}_t$
  & ``No coffee on jogging days'' \\
\bottomrule
\end{tabular}
}
\caption{Activity-based condition types. $\mathcal{A}_m$ = main activity, $\mathcal{A}_t$ = trigger activity.}
\label{tab:activity-conditions}
\end{table}

\subsubsection{Condition Merging}
\label{sec:appendix-merging}

When multiple conditions target the same activity, their binary labels are merged day-by-day using the \texttt{operation} field as is shown in Table~\ref{tab:operations}. Conditions are applied sequentially; later conditions in the list take higher priority.

\begin{table}[h]
\centering
\begin{tabular}{lll}
\toprule
\textbf{Operation} & \textbf{Merge Rule} & \textbf{Use Case} \\
\midrule
\texttt{or}  & $m_d \leftarrow m_d \lor l^{(i)}_d$  & Additive: activity \emph{should} happen \\
\texttt{and} & $m_d \leftarrow m_d \land l^{(i)}_d$  & Restrictive: activity should \emph{not} happen \\
\bottomrule
\end{tabular}
\caption{Condition operations and their semantics. $l^{(i)}_d$ is the label from condition $i$ for day $d$; $m_d$ is the running merged label.}
\label{tab:operations}
\end{table}

\paragraph{Running example.}
Consider the phenomenon ``Father Sim drinks coffee every weekday morning, but not on jogging days.'' This is implemented with two conditions on the same $\mathcal{A}_{\text{coffee}}$:
\begin{enumerate}[leftmargin=*, itemsep=3pt, parsep=0pt, topsep=0pt, partopsep=3pt]
  \item \texttt{every\_selected\_weekday}: on Monday, Tuesday, Wednesday, Thursday, Friday, \texttt{operation}=\texttt{or} $\;\Rightarrow\;$ labels coffee on each weekday.
  \item \texttt{no\_activity\_within\_same\_day}: trigger = $\mathcal{A}_{\text{jogging}}$, frequency = $0.1$, \texttt{operation}=\texttt{and} $\;\Rightarrow\;$ samples jogging events with 10\% frequency, and then removes coffee on days when jogging occurs.
\end{enumerate}
The merged schedule reflects the conjunction: coffee appears on weekdays \emph{except} those with a jog.

\subsection{Profile}
\label{sec:appendix-profile}

A \textbf{Human Profile} bundles the conditions that define one stable behavioral regime for a character: a static pattern is a single profile, and a dynamic phenomenon is several profiles linked together (\ref{sec:appendix-phenomenon}). Formally, it is a named behavioral state observed over a specified duration:

\begin{equation}
H = (\textit{desc},\; N,\; T)
\label{eq:profile}
\end{equation}

\noindent
where $N$ is the observation duration in days and $T$ is a list of conditions, each either temporal or activity-based (\ref{sec:appendix-temporal} and \ref{sec:appendix-activity-conditions}).

While the underlying system allows a profile's conditions to target different activities (i.e., a single profile could encode several independent activity patterns at once), \textsc{SimLife-BP} restricts every condition within a profile to share the same target activity. This keeps each task focused on inferring the rule governing one activity, rather than requiring a model to first disentangle which evidence belongs to which of several interleaved patterns.

\paragraph{Example: ``Father Sim coffee drinker'' ($\textit{length} = 21$ days), all conditions targeting $\mathcal{A}_{\text{coffee}}$.}
\begin{itemize}[leftmargin=*, itemsep=3pt, parsep=0pt, topsep=0pt, partopsep=3pt]
  \item \texttt{every\_selected\_weekday}(weekdays=$\{$Monday, Tuesday, Wednesday, Thursday, Friday$\}$, op=\texttt{or})
  \item \texttt{right\_after}(trigger=$\mathcal{A}_{\text{late\_run}}$, op=\texttt{or})
\end{itemize}
This profile labels coffee on every weekday \emph{and} on any day following a late-evening run.

\subsection{Phenomenon}
\label{sec:appendix-phenomenon}

A \textbf{Phenomenon} is the top-level behavioral narrative that profiles compose into and the object each generated episode ultimately corresponds to:

\begin{equation}
P = (\mathbf{H},\; \mathbf{E}, \; \textbf{start date})
\label{eq:phenomenon}
\end{equation}

\noindent
where $\mathbf{H} = [H_1, \ldots, H_k]$ is an ordered list of profiles and $\mathbf{E} = [E_1, \ldots, E_{k-1}]$ is a list of evolution rules connecting consecutive profiles. The \textbf{start date} indicates the calendar date that the phenomenon starts to simulate. This structure enables us to assemble video chain instances for our dataset. 

We support two phenomenon types (Table~\ref{tab:phenomenon-types}):

\begin{table}[h]
\centering
\begin{tabular}{lccl}
\toprule
\textbf{Type} & $|\mathbf{H}|$ & $|\mathbf{E}|$ & \textbf{Description} \\
\midrule
Pattern & 1 & 0 & Static behavioral pattern; no change over time \\
Dynamic & $\geq 2$ & $\geq 1$ & Two or more behavioral phases connected by (a) life event(s) \\
\bottomrule
\end{tabular}
\caption{Phenomenon types and their structural constraints.}
\label{tab:phenomenon-types}
\end{table}

\subsubsection{Evolution Rules}
\label{sec:appendix-evolution}

An \textbf{Evolution Rule} is needed only when a phenomenon is dynamic: it describes a transition between two profiles, triggered by a life event:

\begin{equation}
\begin{split}
E = (&\textit{desc},\; H_{\text{from}},\; H_{\text{to}},\;\\
&\mathcal{A}_{\text{must\_have}},\; \mathcal{A}_{\text{must\_not\_have}})
\end{split}
\label{eq:evolution}
\end{equation}

\noindent
The transition occupies exactly one day, during which the activities in $\mathcal{A}_{\text{must\_have}}$ are required and those in $\mathcal{A}_{\text{must\_not\_have}}$ are forbidden.

\paragraph{Example phenomenon for Dynamic Understanding.}

\begin{quote}
``Father Sim drinks coffee every weekday morning. After a health scare, he stops drinking coffee and switches to morning jogging.''
\end{quote}

\begin{itemize}[leftmargin=*, itemsep=3pt, parsep=0pt, topsep=0pt, partopsep=3pt]
  \item $H_1$ = ``coffee drinker'' (21 days): \texttt{every\_selected\_weekday}(M,T,W,Th,F) on $\mathcal{A}_{\text{coffee}}$
  \item $E_1$ = ``health scare'': $H_1 \rightarrow H_2$
  \item $H_2$ = ``healthy lifestyle'' (14 days):
      \texttt{no\_activity\_within\_same\_day}(\\trigger=$\mathcal{A}_{\text{jogging}}$, freq=1) on $\mathcal{A}_{\text{coffee}}$
\end{itemize}

\noindent
The resulting observation spans $21 + 1 + 14 = 36$ days.

\subsubsection{Phenomena Generation with LLMs}
\label{sec:appendix-generation}

Phenomena are generated using Gemini 2.5 Flash with structured prompts that provide:
\begin{enumerate}[leftmargin=*, itemsep=3pt, parsep=0pt, topsep=0pt, partopsep=3pt]
  \item The full action catalog (Table~\ref{tab:action-categories}).
  \item The activity-matching schema used to express conditions, formalized as an Activity Specification in \ref{sec:appendix-schedule}.
  \item All condition trigger templates (Tables~\ref{tab:temporal-conditions} and~\ref{tab:activity-conditions}) with examples.
  \item The output JSON schema specifying profiles, conditions, and evolution rules.
  \item Constraints: minimum conditions per profile ($\geq 2$), number of profiles, and number of evolution rules.
\end{enumerate}

\noindent
For \textit{pattern} phenomena, the prompt requests one profile with no evolution rules. For \textit{dynamic} phenomena, the prompt requests exactly two profiles and one evolution rule connecting them. Generated JSON is validated against the schema before manual verification.

While we use LLM generation for some ideas for new phenomena to add to our dataset, most of the data was created through manual annotation. A portion of the phenomena for our data was also generated automatically using simulated day activity distributions to ensure we had enough simulated day videos to create the video chain instances. 

With a phenomenon specified, the remaining two sections describe how it becomes data: a schedule compiles it into a concrete, date-by-date requirement matrix (\ref{sec:appendix-schedule}), and a video chain instantiates that matrix as an ordered sequence of recordings (\ref{sec:appendix-video-chain}).

\subsection{Schedule Generation}
\label{sec:appendix-schedule}

A \textbf{schedule} compiles a declarative phenomenon into a concrete, date-by-date requirement: it turns ``this rule holds'' into ``here is what day 12 must contain.'' The same phenomenon compiled from a different start date, or with different independently-sampled trigger days, yields a different schedule. This makes it possible to generate multiple valid schedules with one phenomenon, as well as doing rejection sampling until getting a valid schedule.
Each requirement in a schedule is expressed as an \textbf{Activity Specification}, which filters the pool of recorded days and returns those that contain a matching activity.

\subsubsection{Activity Specification}
\label{sec:appendix-activity-spec}

An \textbf{Activity Specification} (ActivitySpec) selects a class of concrete activities along three orthogonal dimensions:

\begin{equation}
\begin{split}
\mathcal{A} = (&\textit{actions},\; \textit{time\_zones}, \\
&\textit{allowed\_participant\_sets})
\end{split}
\label{eq:activity-spec}
\end{equation}

\noindent
We detail each dimension in Table~\ref{tab:activity-spec-components}. A concrete activity matches $\mathcal{A}$ if and only if \emph{all three} dimensions are satisfied simultaneously; any recorded day containing such an activity is a candidate to satisfy this requirement.

\begin{table}[h]
\centering
\small
\resizebox{\columnwidth}{!}{
\begin{tabular}{lp{7cm}p{5cm}}
\toprule
\textbf{Component} & \textbf{Description} & \textbf{Matching} \\
\midrule
\texttt{actions}       & List of action names (e.g., \texttt{drink\_coffee}, \texttt{jogging}) & Any one matches (OR) \\
\texttt{time\_zones}   & List of \texttt{TimeZone} (TZ) objects, each with \texttt{start\_time}, \texttt{end\_time}, and \texttt{type} $\in$ \{\texttt{must\_have}, \texttt{must\_not\_have}\} & \texttt{must\_have}: OR across zones; \texttt{must\_not\_have}: AND (avoid all) \\
\texttt{allowed\_participant\_sets} & List of participant combinations, each a list of (character, role) pairs & Any one set matches (OR) \\
\bottomrule
\end{tabular}
}
\caption{Components of an Activity Specification and their matching semantics.}
\label{tab:activity-spec-components}
\end{table}

\paragraph{Day-level scope.} An ActivitySpec matches within a single recorded day and carries no information about which day it is or where it falls on the calendar; its \texttt{time\_zones} are clock windows (e.g., 06:00--10:00), not weekdays or dates. Whether a requirement should apply on a given day, such as only on weekdays, is determined separately by schedules which are defined with conditions (\ref{sec:appendix-temporal}).

\paragraph{Example 1: Father Sim drinks coffee in the morning.}

\begin{equation}
\begin{split}
\mathcal{A}_{\text{coffee}} = \Big(
  \big[\texttt{drink\_coffee}\big],\;
  \\ \big[\text{TZ}(\text{06:00},\, \text{10:00},\, \texttt{must\_have})\big],\;
  \\ \big[\big[(\text{Father Sim},\, \texttt{doer1})\big]\big]
\Big)
\end{split}
\label{eq:activity-spec-coffee}
\end{equation}

\paragraph{Example 2: Mother Sim or Daughter Sim does any fitness activity in the evening.}

\begin{equation}
\begin{split}
\mathcal{A}_{\text{fitness}} = \Big(
  &\big[\texttt{endurance\_run},\, \texttt{jogging},\, \texttt{box}\big], \\
  &\big[\text{TZ}(\text{15:00},\, \text{22:00},\, \texttt{must\_have})\big], \\
  &\big[\big[(\text{Mother Sim},\, \texttt{doer1})\big],\; \\
  &\;\,\big[(\text{Daughter Sim},\, \texttt{doer1})\big]\big]
\Big)
\end{split}
\label{eq:activity-spec-fitness}
\end{equation}

\noindent
This specification matches any of the three fitness actions, performed by either Mother Sim or Daughter Sim (but not both simultaneously), between 15:00 and 22:00.

Given a phenomenon $P$ with profiles $\mathbf{H}$ and a start date, the schedule generator produces a mapping from calendar dates to activity requirements:

\begin{equation}
\mathcal{S}: \textit{date} \rightarrow \{(\mathcal{A}_j,\; b_j)\}_{j=1}^{m}
\label{eq:schedule}
\end{equation}

\noindent
where $m$ is the number of distinct ActivitySpecs referenced by the phenomenon, one for each profile's target activity plus one for each trigger activity used by an activity-based condition, and $b_j \in \{0, 1\}$ indicates whether activity specification $\mathcal{A}_j$ is required on that date. We use these schedules to select simulated day videos and assemble a unique, long video for each phenomenon that shows enough evidence of the pattern the phenomenon describes.

\paragraph{Algorithm.}
For each profile $H_i$:
\begin{enumerate}[leftmargin=*, itemsep=3pt, parsep=0pt, topsep=0pt, partopsep=3pt]
  \item \textbf{Pre-sample trigger activities.} For each unique trigger activity spec across all activity-based conditions, independently sample each date as active with probability $f$ (\ref{sec:appendix-activity-conditions}).
  \item \textbf{Generate per-condition labels.} Dispatch each condition to its handler (Tables~\ref{tab:temporal-conditions}--\ref{tab:activity-conditions}), producing binary labels.
  \item \textbf{Merge labels.} Apply each condition's \texttt{operation} (Table~\ref{tab:operations}) sequentially to produce the final schedule.
\end{enumerate}

\paragraph{Constraint matrix.}
The schedule is converted into a boolean matrix $\mathbf{M} \in \{0,1\}^{N \times m}$, where $M_{d,j} = 1$ if day $d$ requires activity spec $\mathcal{A}_j$. Each row defines a unique constraint pattern that Video Chain construction (\ref{sec:appendix-video-chain}) matches against the pool of pre-recorded simulated days.

\subsection{Video Chain}
\label{sec:appendix-video-chain}

A \textbf{Video Chain} instantiates a schedule: it realizes the abstract requirement matrix $\mathbf{M}$ as an ordered sequence of actual recordings, and is the final data instance inserted into the dataset.

\begin{equation}
\mathcal{V} = [s_1, s_2, \ldots, s_N]
\label{eq:video-chain}
\end{equation}

\noindent
where each $s_d$ is a pre-recorded simulated day whose activities satisfy the constraint pattern at position $d$.

\paragraph{Construction.}
For each unique constraint pattern (row of $\mathbf{M}$), the system queries the database for simulated days whose recorded activities match all required and forbidden activity specifications (\ref{sec:appendix-activity-spec}), filtered by weekday/weekend status. Days are sampled without replacement; if the pool is exhausted, it resets for round-robin reuse.

\paragraph{Multi-profile chains.}
For dynamic phenomena, the video chain is built segment by segment:
\begin{equation}
\mathcal{V} = [\underbrace{s_1, \ldots, s_{|H_1|}}_{\text{Profile } H_1},\; \underbrace{s_{|H_1|+1}}_{\text{Evolution}},\; \underbrace{s_{|H_1|+2}, \ldots, s_N}_{\text{Profile } H_2}]
\end{equation}

\noindent
The evolution rule day must satisfy $\mathcal{A}_{\text{must\_have}}$ and avoid $\mathcal{A}_{\text{must\_not\_have}}$.

Each day in the chain carries two metadata records:
\begin{itemize}[leftmargin=*, itemsep=3pt, parsep=0pt, topsep=0pt, partopsep=3pt]
  \item \textbf{Day Activity Occurrence Set}: which activity specs were required/forbidden.
  \item \textbf{Day Condition Set}: which conditions were satisfied (boolean flags per condition). This is used for generating counterfactual questions.
\end{itemize}

\subsection{Illustrative End-to-End Example}
\label{sec:appendix-example}

We trace the pipeline described above, end to end, for one pattern phenomenon:

\begin{tcolorbox}[
    width=\linewidth,
    colback=blue!3!white,
    colframe=blue!60!black,
    fonttitle=\bfseries,
]
Son Sim practices guitar every weekday evening. On days when Daughter Sim plays violin, Son Sim also plays guitar within the same day.
\end{tcolorbox}

\paragraph{Step 1: Activity Specifications.}
\begin{align*}
\mathcal{A}_{\text{guitar}} &= \big([\texttt{practice\_guitar}], \\
&\quad [\text{TZ}(\text{15:00}, \text{22:00}, \texttt{must\_have})], \\
&\quad [[(\text{Son Sim}, \texttt{doer1})]]\big) \\
\mathcal{A}_{\text{violin}} &= \big([\texttt{practice\_violin}], \\
&\quad [\text{TZ}(\text{15:00}, \text{22:00}, \texttt{must\_have})], \\
&\quad [[(\text{Daughter Sim}, \texttt{doer1})]]\big)
\end{align*}

\paragraph{Step 2: Conditions.}
\begin{enumerate}[leftmargin=*, itemsep=3pt, parsep=0pt, topsep=0pt, partopsep=3pt]
  \item $C_1$ = \texttt{every\_selected\_weekday}(weekdays=\\$\{$Monday, Tuesday, Wednesday, Thursday, Friday$\}$, op=\texttt{or}) on $\mathcal{A}_{\text{guitar}}$: guitar every weekday.
  \item $C_2$ = \texttt{within\_same\_day}(trigger=$\mathcal{A}_{\text{violin}}$, freq=0.1, op=\texttt{or}) on $\mathcal{A}_{\text{guitar}}$: guitar on violin days too.
\end{enumerate}

\paragraph{Step 3: Profile.}
\begin{align*}
\mathcal{H} = (\text{``Son guitar practice''},\; 30,\; [C_1, C_2]).
\end{align*}

\paragraph{Step 4: Schedule.}
Over 30 days, the merged schedule labels guitar on all weekdays (from $C_1$) plus any weekend day when violin is triggered (from $C_2$, with 10\% probability per day).

\paragraph{Step 5: Video chain.}
The constraint matrix distinguishes three day types: (a)~guitar required, (b)~both guitar and violin required, and (c)~neither required. For each type, matching simulated days are sampled from the database and assembled into a 30-day video chain.

\section{Task Design}
\label{sec:appendix-questions}

Each \OursBench{} task poses four types of multiple-choice questions over the same long video: \emph{direct prediction}, \emph{counterfactual prediction}, \emph{noisy counterfactual prediction}, and \emph{inverse counterfactual reasoning}. All four are built on a single abstraction, the \emph{Day Condition Set} (Appendix~\ref{sec:appendix-video-chain}), and together probe whether a model has recovered the rule governing a character's behavior.

The design factors into a shared foundation and three orthogonal axes. The foundation is the \emph{Day Condition Set} (Appendix~\ref{sec:appendix-q-dcs}): the truth table of the behavioral rule a task targets. The three axes are \emph{what} is asked (the four question types, Appendix~\ref{sec:appendix-q-types}), \emph{how much} of the rule is revealed (hint levels, Appendix~\ref{sec:appendix-q-hint}), and \emph{which} behavioral phase is targeted (standard vs.\ history questions, Appendix~\ref{sec:appendix-q-history}). Appendix~\ref{sec:appendix-q-anatomy} composes the three axes into one rendered question, and Appendices~\ref{sec:appendix-q-noise} and~\ref{sec:appendix-q-impl} document two implementation details: noise-condition selection and the natural-language rendering pass.

\subsection{Setup: the Day Condition Set}
\label{sec:appendix-q-dcs}

For every day in a video chain we record two things: (i)~the boolean flag for each condition that governs the \emph{main activity} (e.g., did ``today is Saturday'' hold? or did ``Daughter Sim played violin the day before'' hold?), and (ii)~the \textbf{outcome} $M \in \{0,1\}$, whether the main activity actually occurred that day. We call this record the \textbf{Day Condition Set} (DCS): a condition configuration together with its outcome. We write $M(d)$ for the outcome of a DCS $d$.

Across a chain only a few distinct DCSs appear; collecting them yields the \emph{truth table} of the phenomenon's underlying rule. A task is anchored to one \textbf{observed} DCS $d_{\text{obs}}$. The four question types are different ways of interrogating this truth table.

\paragraph{Running example}
We use one phenomenon throughout this section:
\begin{tcolorbox}[width=\linewidth, colback=blue!3!white, colframe=blue!60!black]
Son Sim plays chess in the evening (3PM--10PM) on weekends, and also on any day right after Daughter Sim plays violin.
\end{tcolorbox}
\noindent
The main activity $\mathcal{A}_{\text{chess}}$ is governed by two conditions, merged with \texttt{or}:
\begin{itemize}[leftmargin=*, itemsep=3pt, parsep=0pt, topsep=0pt, partopsep=3pt]
  \item $C_1$ (temporal): \textit{today is Saturday or Sunday}.
  \item $C_2$ (activity, \texttt{right\_after}): \textit{Daughter Sim played violin the day before}.
\end{itemize}
Across the chain the days fall into four distinct DCSs, giving the truth table in Table~\ref{tab:question-truth-table}.

\begin{table}[!ht]
\centering
\begin{tabular}{cccc}
\toprule
\textbf{DCS} & \textbf{$C_1$ (weekend)} & \textbf{$C_2$ (violin yesterday)} & \textbf{$M$ (chess occurs)} \\
\midrule
$d_0$ & F & F & 0 \\
$d_1$ & T & F & 1 \\
$d_2$ & F & T & 1 \\
$d_3$ & T & T & 1 \\
\bottomrule
\end{tabular}
\caption{Truth table for the chess phenomenon ($M = C_1 \lor C_2$).}
\label{tab:question-truth-table}
\end{table}

\subsection{The Four Question Types}
\label{sec:appendix-q-types}

The four types are different interrogations of the same truth table (Table~\ref{tab:question-types}). We illustrate each on the running example, taking the observed DCS to be $d_{\text{obs}} = d_2$: a weekday following a violin day, on which chess \emph{did} occur ($M(d_2) = 1$).

\begin{table*}[!ht]
\centering
\small
\resizebox{\columnwidth}{!}{
\begin{tabular}{llll}
\toprule
\textbf{Type} & \textbf{Direction} & \textbf{Answer} & \textbf{What it tests} \\
\midrule
Direct                 & ---                       & $M(d_{\text{obs}})$                & reading the actual outcome \\
Counterfactual         & cause $\rightarrow$ effect & $M(d_{\text{cf}})$                 & causal propagation \\
Noise counterfactual   & cause $\rightarrow$ effect & $M(d_{\text{obs}})$ (unchanged)    & resistance to spurious cues \\
Inverse counterfactual & effect $\rightarrow$ cause & condition matching $M(d_{\text{inv}})$ & abductive (backward) inference \\
\bottomrule
\end{tabular}
}
\caption{The four question types. $d_{\text{obs}}$ is the observed DCS; $d_{\text{cf}}$ and $d_{\text{inv}}$ are alternative DCSs chosen for the counterfactual and inverse types.}
\label{tab:question-types}
\vspace{-10pt}
\end{table*}

\paragraph{Direct Prediction}
The baseline. It asks about the observed day exactly as recorded, and the answer is the true outcome $M(d_{\text{obs}})$; no hypothetical reasoning is required. This measures whether the model can read off what actually happened, and serves as the reference point against which the counterfactual variants are scored. \emph{Example:} \textit{``Will Son Sim play chess this evening?''} $\rightarrow$ \textbf{Yes} ($M(d_{\text{obs}}) = 1$).

\paragraph{Counterfactual Prediction}
We select a \emph{different} DCS $d_{\text{cf}}$ from the same truth table and ask what would happen under it; the answer is $M(d_{\text{cf}})$. This requires models to situate themselves into the counterfactual condition and predict the occurrence of the target activity accordingly. \emph{Example:} Let $d_{\text{cf}} = d_0$ ($M(d_0) = 0$). \textit{``If Daughter Sim had not played violin the day before, would Son Sim play chess this evening?''} $\rightarrow$ \textbf{No}. Note that we only require $d_{\text{obs}} \neq d_{\text{cf}}$, thus sometimes the outcome of direct prediction would be the same as counterfactual prediction. 

\paragraph{Noise Counterfactual Prediction}
Structurally a counterfactual, but the introduced condition is \emph{causally irrelevant} to the main activity, such as an unrelated action by another character. Because the noisy condition does not enter the truth table, the answer stays at $M(d_{\text{obs}})$. This requires models to reason that the given condition is independent to the occurrence of the target activity, thus answer the question based on the current observation. \emph{Example:} \textit{``If Father Sim had washed the dishes yesterday, would Son Sim play chess this evening?''} $\rightarrow$ \textbf{Yes}, unchanged ($M(d_{\text{obs}}) = 1$).

\paragraph{Inverse Counterfactual Reasoning}
We assert the negated outcome as a premise ($\text{premise} = \neg M(d_{\text{obs}})$) and ask which \emph{conditions} are consistent with it, choosing between a condition statement drawn from an alternative DCS $d_{\text{inv}}$ and its opposite. The answer follows from matching the premised outcome against $M(d_{\text{inv}})$. Unlike the forward types, this requires models to reason abductively, inferring a plausible cause from a stated effect rather than propagating a known cause to its effect. \emph{Example:} $d_{\text{inv}} = d_0$ ($M(d_0) = 0$) \textit{``If Son Sim had not played chess this evening, which statement is more likely to be true? (A)~Daughter Sim did not play violin the day before; (B)~Daughter Sim played violin the day before.''} $\rightarrow$ \textbf{(A)}. Under the \texttt{or} rule on a non-weekend day, chess can fail only if the violin trigger was also absent, so the model reasons from effect back to cause.

Each type is rendered in two surface formats: a \textbf{yes/no} form (``Will $X$ \ldots today?'') and a \textbf{two-option} form (``which statement is more likely to be true?''), whose templates are collected in Table~\ref{tab:question-bodies}.

\begin{table*}[!ht]
\centering
\small
\resizebox{\columnwidth}{!}{
\begin{tabular}{llp{10.0cm}}
\toprule
\textbf{Type} & \textbf{Format} & \textbf{Body template} \\
\midrule
Direct                 & yes/no     & \texttt{Will \{sim\} \{action\} today\{time\}?} \\
                       & two-option & \texttt{Which statement is more likely to be true?} \\
Counterfactual         & yes/no     & \texttt{If \{conditions of d\_cf\}, will \{sim\} \{action\} today\{time\}?} \\
                       & two-option & \texttt{If \{conditions of d\_cf\}, which statement is more likely to be true?} \\
Noise counterfactual   & yes/no     & \texttt{If \{noise\}, will \{sim\} \{action\} today\{time\}?} \\
                       & two-option & \texttt{If \{noise\}, which statement is more likely to be true?} \\
Inverse counterfactual & yes/no     & \texttt{If \{M\}, is the statement "\{conditions of d\_inv\}" likely to be true?} \\
                       & two-option & \texttt{If \{premise\}, which statement is more likely to be true?} \\
\bottomrule
\end{tabular}
}
\caption{Question Body templates by type and format. \texttt{\{time\}} is an optional clock range (e.g.\ \textit{between 3PM and 10PM}). \emph{Yes/no} options are \texttt{``Yes''}/\texttt{``No''}; \emph{two-option} choices are \texttt{``\{sim\} will (not) \{action\} today\{time\}''} for the first three types, and the two condition statements \texttt{\{icf\_body\}} vs.\ \texttt{\{icf\_body\_opposite\}} for inverse counterfactual. Templates are emitted in the indicative; the LLM rendering pass (Appendix~\ref{sec:appendix-q-impl}) keeps direct questions indicative and converts the three counterfactual types to the subjunctive.}
\label{tab:question-bodies}
\end{table*}

\subsection{Hint Levels}
\label{sec:appendix-q-hint}

The four question types fix \emph{what} is asked; the \textbf{hint level} fixes \emph{how much of the governing rule is revealed} before the model must answer. We expose three hint levels as detailed in Table~\ref{tab:hint-levels}.

\begin{table}[ht]
\centering
\begin{tabular}{lll}
\toprule
\textbf{Level} & \textbf{Rendered as} & \textbf{What the model is told} \\
\midrule
No hint      & \texttt{<body>}                  & nothing about the rule \\
Partial hint & \texttt{<body> (<hint$_p$>)}     & condition categories + logic \\
Full hint    & \texttt{<body> (<hint$_f$>)}     & exact condition bindings \\
\bottomrule
\end{tabular}
\caption{The three hint levels. \texttt{<body>} refers to the four question types introduced in Appendix~\ref{sec:appendix-q-types}.}
\label{tab:hint-levels}
\end{table}

\noindent
Each governing condition contributes one clause, rendered differently at the partial and full levels (Table~\ref{tab:hint-clauses}). Here \texttt{\{participant\}} is a character (e.g.\ \textit{Daughter Sim}), \texttt{\{activity\}} the natural-language action (e.g.\ \textit{playing violin}), and \texttt{\{window\}} an optional clock range \texttt{``from HH AM/PM to HH AM/PM''} omitted when the activity spans the default observable day (6AM--10PM).

\begin{table*}[!ht]
\centering
\small
\resizebox{\linewidth}{!}{
\begin{tabular}{llp{5.0cm}p{5.0cm}}
\toprule
\textbf{Family} & \textbf{Condition Type} & \textbf{Partial Hint} & \textbf{Full Hint} \\
\midrule
Temporal & \texttt{every\_selected\_weekday} & the day of the week & weekday names, e.g.\ \texttt{Monday, Wednesday}; all seven $\rightarrow$ \texttt{from Sunday to Saturday} \\
Temporal & \texttt{on\_date} (date form)     & the date in the month            & the \texttt{\{ordinal\}} of each month; \texttt{the last day of each month} \\
Temporal & \texttt{on\_date} (weekday form)  & a specific weekday of each month & the \texttt{\{ordinal\} \{Weekday\}} of each month \\
Temporal & \texttt{repeat\_every\_other}     & related to whether a specific action is done by \texttt{\{participant\}\{window\}} the day before & related to whether \texttt{\{participant\}} does \texttt{\{activity\}\{window\}} the day before \\
Activity & \texttt{right\_after}             & related to a specific action done by \texttt{\{participant\}} the day before\texttt{\{window\}}    & related to \texttt{\{participant\}} doing \texttt{\{activity\}\{window\}} the day before \\
Activity & \texttt{within\_same\_day}        & related to a specific action done by \texttt{\{participant\}} today\texttt{\{window\}}              & related to \texttt{\{participant\}} doing \texttt{\{activity\}\{window\}} within the same day \\
Activity & \texttt{no\_activity\_right\_after}      & related to a specific action done by \texttt{\{participant\}} the day before\texttt{\{window\}} & related to \texttt{\{participant\}} doing \texttt{\{activity\}\{window\}} the day before \\
Activity & \texttt{no\_activity\_within\_same\_day} & related to a specific action done by \texttt{\{participant\}} today\texttt{\{window\}}           & related to \texttt{\{participant\}} doing \texttt{\{activity\}\{window\}} within the same day \\
\bottomrule
\end{tabular}
}
\caption{Per-condition hints at the partial and full levels.}
\label{tab:hint-clauses}
\end{table*}

\noindent
For the running example (Appendix~\ref{sec:appendix-q-dcs}), the three levels of the direct question read:
\begin{itemize}[leftmargin=*, itemsep=3pt, parsep=0pt, topsep=0pt, partopsep=3pt]
  \item \textbf{No Hint:} \textit{``Will Son Sim play chess today from 3PM to 10PM?''}
  \item \textbf{Partial Hint:} \textit{``\ldots{} (It is related to the day of the week and to a specific action done by Daughter Sim the day before.)''}
  \item \textbf{Full Hint:} \textit{``\ldots{} (It is related to whether today is Saturday or Sunday and to whether Daughter Sim played violin the day before.)''}
\end{itemize}

\subsection{History Questions}
\label{sec:appendix-q-history}

\OursBench{} also includes \textbf{history} questions (or \texttt{Hist} in~\ref{sec:result-pattern-adaptation}), which ask about an \emph{earlier} behavioral phase of a dynamic phenomenon: a period governed by the rule that held \emph{before} a life event changed it (Appendix~\ref{sec:appendix-evolution}). Because these questions concern a past period rather than ``today,'' each is prepended with a temporal \textbf{anchor} that re-situates the model in the corresponding phase. We design two types of temporal anchors (Table~\ref{tab:history-anchors}) that coarsely identify the relevant historical phase, encouraging models to reason about the previous pattern rather than retrieve a specific past day.
History questions inherit the three counterfactual variants from standard questions\footnote{We use standard questions to refer to questions about the current pattern.}, but omit direct prediction, since asking about an imagined past is counterfactual by default.

\begin{table}[ht]
\centering
\begin{tabular}{lp{5.4cm}}
\toprule
\textbf{Anchor} & \textbf{Prefix template / example} \\
\midrule
Evolution-description & \texttt{Back before \{event\}, <body>} \newline \textit{``Back before Son Sim started a new school term, \ldots''} \\
Vague-days & \texttt{\{vague\_phrase\}, <body>} \newline \textit{``About a month earlier, \ldots''} \\
\bottomrule
\end{tabular}
\caption{History anchors. \texttt{\{vague\_phrase\}} is one phrase sampled deterministically per task (shared across its questions) from a table bucketed by how far back the phase sits, e.g. 22--28 $\rightarrow$ ``About three weeks earlier,'' 29--35 $\rightarrow$ ``About a month earlier.''}
\label{tab:history-anchors}
\end{table}

\subsection{Question Anatomy}
\label{sec:appendix-q-anatomy}

Every question is issued in one of two surface formats: a \textbf{yes/no} form (``\ldots will $X$ \ldots today?'') and a \textbf{two-option} form (``which statement is more likely to be true?''). The three axes above---question type, hint level, and task kind---then compose into a single template. Every rendered question, in either surface format, is the concatenation of three slots:
\[
\underbrace{\langle\text{history prefix}\rangle}_{\text{history tasks only}}\;\;
        \underbrace{\langle\text{question body}\rangle}_{\text{always}}
        \underbrace{(\langle\text{hint}\rangle)}_{\text{partial / full only}}\;
\]

\noindent
The \textbf{body} is selected by question type and answer format (Table~\ref{tab:question-bodies}); the \textbf{history prefix} is supplied for history tasks (Table~\ref{tab:history-anchors}, empty otherwise); and the \textbf{hint} is appended at the partial or full level (Tables~\ref{tab:hint-levels}--\ref{tab:hint-clauses}, empty at no-hint).

\noindent
A fully populated example---a \emph{history}, \emph{counterfactual}, \emph{partial-hint} question---exercises all three slots:
\begin{tcolorbox}[width=\linewidth, colback=blue!3!white, colframe=blue!60!black]
\textbf{\textcolor{blue!60!black}{[prefix]}} Back before Son Sim started a new school term,
\textbf{\textcolor{blue!60!black}{[body]}} if Daughter Sim had not played violin the day before, would Son Sim play chess this evening?
\textbf{\textcolor{blue!60!black}{[hint (partial)]}} (It is related to the day of the week and to a specific action done by Daughter Sim the day before.)
\end{tcolorbox}

\subsection{Selecting a Noise Condition}
\label{sec:appendix-q-noise}

A noise counterfactual is only meaningful if the introduced condition is genuinely uninformative about the main activity. We do not pick such a condition by hand or by semantic intuition; instead we require it to be \emph{empirically independent} of the main activity $A$ within the specific video chain, and we verify this with a $2\times2$ contingency table.

\paragraph{Four-cell independence test.}
First, we use the occurrence of an action as the noise condition. For a candidate action $X$ we walk every day of the chain and tally it by two booleans: whether $X$ occurred in the relevant window (under the same temporal semantics as the ground-truth condition, e.g. \texttt{right\_after} checks day $T\!-\!1$), and whether the main activity $A$ occurred that day. This yields the table in Table~\ref{tab:noise-four-cells}.

\begin{table}[ht]
\centering
\small
\begin{tabular}{lcc}
\toprule
 & \textbf{$A$ occurred} & \textbf{$A$ did not} \\
\midrule
\textbf{$X$ occurred}     & $\geq 1$ & $\geq 1$ \\
\textbf{$X$ did not occur} & $\geq 1$ & $\geq 1$ \\
\bottomrule
\end{tabular}
\caption{Independence test for a candidate noise action $X$. All four cells must be observed at least once for $X$ to qualify.}
\label{tab:noise-four-cells}
\end{table}

\noindent
$X$ qualifies only when \emph{all four} cells are non-empty: the chain must contain days where $X$ co-occurs with $A$ happening, with $A$ not happening, and likewise days without $X$ under both outcomes. A populated table demonstrates that the presence of $X$ carries no information about $A$, so $X$ is a true distractor rather than an accidental correlate. \emph{Edge case:} when $A$ is invariant across the chain (it structurally always or never occurs, so only one outcome row is reachable), independence cannot be shown from a single row; the test then relaxes to requiring $X$ itself to vary: both $X$ occurred and $X$ did not occur must appear in the reachable row.

\paragraph{Candidate pool.}
The candidate $X$ is drawn from the same condition-slot machinery used by hypothesis elimination, so the noise action is checked under exactly the same lookup semantics as a real condition would be. When the ground truth contains an activity condition, we reuse the first one's slot parameters (participant, time zones, action-hierarchy level, and condition type) and search its action pool, excluding the true cause itself. When the ground truth is purely temporal, we synthesize a fixed condition \texttt{right\_after}, the same participant as the main activity, a full observable-day window (06:00--22:00), and subcategory level. Socially entangled action categories are excluded from the pool, since their occurrence is too context-dependent to serve as clean noise. The search is exhaustive over the pool and returns the first $X$ that passes the test; if none does, no noise counterfactual is generated for that task.

\subsection{Generation and Rendering}
\label{sec:appendix-q-impl}

Questions are produced in two stages. First, a templating stage emits each question with structured metadata: the differing conditions for a counterfactual (\texttt{if\_clause}), the irrelevant action for a noise counterfactual (\texttt{noise\_framing}), and the negated outcome plus candidate condition lists for an inverse counterfactual (\texttt{premise}, \texttt{icf\_body}, \texttt{icf\_body\_opposite}). Second, an LLM rendering pass converts these templates into fluent, mood-correct natural language: indicative for direct questions and subjunctive for the three counterfactual types, preserving all logical content and the exact character names and clock-time spans.

\section{Hypothesis Elimination For Task Insertion}
\label{app:elimination}

A benchmark task is inserted at a given point only if every plausible reading of the hint that is still consistent with what has been observed so far agrees on the answer. This section formalizes that criterion: we enumerate every hypothesis a hint permits, discard those contradicted by the recorded days, and query the model only once the survivors all agree. We use the partial-hint setting as the reference, since observations, hint, and question together guarantee this criterion is satisfiable under both the partial-hint and full-hint settings. In the no-hint setting, sufficiency cannot be formally defined because the hypothesis space is unbounded; we use it instead to evaluate open-ended pattern reasoning under conditions closer to the real world.

We ground each definition below in one running example: a partial hint stating that a target activity ``painting'' is related to the day of the week, with the true rule (unknown to the reasoner) being ``happens on Mondays and Fridays.''

\subsection{Hypothesis space}
\label{app:elim:candidates}

\paragraph{Condition instantiations.}
Consider a target activity governed by $K$ latent conditions $C_1,\dots,C_K$. Each condition is either temporal or activity-based. For each condition $C_i$, the hint level defines a finite hypothesis space $\Omega_i$ of the values that condition could take. For our running example, $K=1$ and the hint restricts $\Omega_1$ to the $2^7-1=127$ non-empty subsets of weekdays $W$.

A candidate condition tuple is
\[
  \omega = (\omega_1,\dots,\omega_K)
  \in \Omega_1 \times \cdots \times \Omega_K .
\]
On day $t$, this tuple induces a binary condition vector
\[
  \mathbf{b}(\omega,t)
  =
  (b_1(\omega_1,t),\dots,b_K(\omega_K,t))
  \in \{0,1\}^K
\]
where $b_i(\omega_i,t)=1$ iff the corresponding condition is satisfied. In our example, $\omega=W$ and $b_1(W,t)=\mathbf {1}[t\in W]$: a single bit saying whether day $t$'s weekday falls in the candidate subset $W$.

\paragraph{Boolean outcome rules.}
A hint tells the reasoner \emph{which conditions} might matter, but not \emph{how} they combine with the outcome. That mapping is itself unknown and must be enumerated. We represent it as a Boolean rule 
\[
  g: \{0,1\}^K \rightarrow \{0,1\},
\]
mapping condition configurations to activity outcomes. Enumerating over all such $g$, rather than assuming the hinted conditions are positively associated with the activity, is what lets this framework stay agnostic to how \Ours{} actually generates data: the benchmark's condition merging uses only \texttt{or}/\texttt{and} (\ref{sec:appendix-merging}), but a reasoner without that knowledge must consider every possibility, including rules where the condition suppresses the activity, has no effect, or the activity always or never occurs. For $K=1$, there are $2^{2^1}=4$ such rules; applied to our example, these are ``always paint,'' ``never paint,'' ``paint iff $t\in W$,'' and ``paint iff $t\notin W$.''

\paragraph{Full hypotheses.}
A full candidate hypothesis is
\[
  \theta = (\omega,g),
\]
with prediction
\[
  \hat{o}_t(\theta)
  =
  g\bigl(\mathbf{b}(\omega,t)\bigr).
\]
The full candidate set is
\[
  \Theta =
  \left\{
  (\omega,g):
  \begin{aligned}
  &\omega \in \Omega_1 \times \cdots \times \Omega_K,\\
  &g:\{0,1\}^K \rightarrow \{0,1\}
  \end{aligned}
  \right\},
\]
of size
\[
  |\Theta|
  =
  \Bigl(\prod_{i=1}^{K} |\Omega_i|\Bigr) \cdot 2^{2^K},
\]
where the first factor enumerates condition instantiations and the second enumerates Boolean outcome rules. For our running example, $|\Theta| = 127 \times 4 = 508$. More generally, since our benchmark uses at most two conditions, the Boolean-rule factor is $4$ for $K=1$ and $16$ for $K=2$.

\subsection{Elimination and label uniqueness}
\label{app:elim:procedure}

\paragraph{Sequential hypothesis elimination.}
Hypothesis elimination proceeds cumulatively over the observed days. Let $o_t\in\{0,1\}$ indicate whether the target activity occurs on day $t$. Starting from the full candidate set $\Theta$, we eliminate any hypothesis whose prediction disagrees with the observed outcome. The remaining set after day $t$ is
\[
  \mathcal{R}_t
  =
  \{\theta \in \mathcal{R}_{t-1}: \hat{o}_t(\theta)=o_t\},
  \qquad
  \mathcal{R}_0=\Theta .
\]
Because the true generating rule produced every observation by construction, it is never contradicted and therefore never eliminated: $\mathcal{R}_t$ always contains it and is never empty. Checking that every hypothesis in $\mathcal{R}_t$ agrees is therefore equivalent to checking that they all agree with the true rule, even without knowing which surviving hypothesis that is.

\paragraph{Convergence day.}

Let $\mathcal{D}$ denote the set of Day Condition Sets (DCSs) associated with the behavioral pattern, and let $d_t\in\mathcal{D}$ denote the DCS instantiated on day $t$. For a sequence of $T$ days, we define the \emph{convergence day} $Y$ as the earliest day satisfying both
\[
\mathcal{R}_t=\mathcal{R}_Y
\qquad
\forall t\geq Y,
\]
and
\[
\forall d\in\mathcal{D},\quad
\exists t\in{Y,\ldots,T}
\text{ such that } d_t=d.
\]
The first condition requires the hypothesis set to have stabilized: after observing day $Y$, no additional hypothesis is eliminated. The second requires the remaining sequence to contain at least one instance of every DCS.

Together, these conditions guarantee that every DCS admits a unique prediction under the surviving hypotheses. For any $d\in\mathcal{D}$, consider a post-convergence day $t$ on which $d_t=d$. Because no hypothesis is eliminated after $Y$, every $\theta\in\mathcal{R}_Y$ must correctly predict the observed outcome on that day:
\[
\hat{o}_t(\theta)=o_t
\qquad
\forall\theta\in\mathcal{R}_Y.
\]


\subsection{Example: weekday-based activity}
\label{app:elim:example}

\begin{tcolorbox}[
    width=\linewidth,
    colback=blue!3!white,
    colframe=blue!60!black,
    fonttitle=\bfseries,
]
\small
\textbf{True pattern}: Daughter Sim paints a picture between 6 AM and 12 PM on Mondays and Fridays.

\textbf{Question}: Will Daughter Sim paint a picture today between 6 AM and 12 PM?

\textbf{Partial hint}: It is related to the day of the week.
\end{tcolorbox}

\paragraph{Elimination trace.}
Table~\ref{tab:elim:trace} tracks $|\mathcal{R}_t|$ as days are observed, starting from the $508$ hypotheses established above. Each observed day removes hypotheses whose prediction disagrees with whether Daughter Sim painted that morning. For example, day~1 is a Wednesday with no painting, so hypotheses predicting painting on Wednesday are eliminated. Day~3 is a Friday with painting, so hypotheses predicting no painting on Friday are eliminated.

\begin{table}[h!]
  \centering
  \small
  \begin{tabular}{@{}rllcr@{}}
    \toprule
    Day $t$ & Date & Weekday & Painted? & Remaining hypotheses \\
    \midrule
    --      & ---        & ---  & ---  & $508$ \\
    $1$     & Oct.~15 & Wed  & no   & $254$ \\
    $2$     & Oct.~16 & Thu  & no   & $190$ \\
    $3$     & Oct.~17 & Fri  & yes  & $32$ \\
    $4$     & Oct.~18 & Sat  & no   & $16$ \\
    $5$     & Oct.~19 & Sun  & no   & $8$ \\
    $6$     & Oct.~20 & Mon  & yes  & $4$ \\
    $7$     & Oct.~21 & Tue  & no   & $2$ \\
    $8$--$14$ & Oct.~22--28 & Wed\,\ldots\,Tue
      & Fri, Mon: yes & $2$ \\
    \bottomrule
  \end{tabular}
  \caption{Elimination trace for the example chain. Daughter Sim paints exactly
  on Mondays and Fridays; the remaining set shrinks from $508$ to $2$ by day~7. The convergence day is therefore day 7.}
  \label{tab:elim:trace}
\end{table}

After all weekdays have been observed once, two hypotheses remain:
\[
\theta_A:
W=\{\text{Mon},\text{Fri}\},\quad
\text{paint on days in } W,
\]
and
\begin{align*}
\theta_B:\quad
W &= \{\text{Tue},\text{Wed},\text{Thu},\text{Sat},\text{Sun}\},\\
&\text{paint on days outside } W.
\end{align*}
These hypotheses use different symbolic descriptions, but they make identical predictions: both imply that Daughter Sim paints exactly on Mondays and Fridays. Therefore, when the direct question is asked after convergence, e.g. on Friday, Oct.~24, the unique answer is \emph{yes}.

\section{Dialogue Generation and Adaptation}
\label{app:dialogue}

The Sims 4 recordings used in SimLife contain silent ``simlish'' mouth animations rather than audible speech. We therefore synthesize English dialogue after recording. The dialogue pipeline has two purposes:
\begin{itemize}[leftmargin=*, itemsep=2pt, parsep=0pt, topsep=0pt, partopsep=2pt]
    \item \textbf{Original dialogue generation}: makes each chat interaction sound natural and temporally aligned with the video.
    \item \textbf{Prompted dialogue adaptation}: modifies generated dialogue to introduce phenomenon-relevant topics, mention activities that are not directly visible, or indicate the change-point event on the pivot day of a dynamic episode.
\end{itemize}

\subsection{Original Dialogue Generation}
\label{app:dialogue:original}

\paragraph{Chatting session detection.}
Original dialogue generation starts from the trimmed gameplay video and game action log. The action log is used to identify chatting sessions, defined as contiguous intervals in which two or more Sims are plausibly engaged in the same log-recorded chat event. Failed, canceled, idle, and menu-level actions are removed, and overlapping chat actions among nearby Sims are grouped into sessions. Sessions are merged only when participants remain temporally and spatially close, and split when a participant leaves or begins a separate chat with someone outside the session.

\paragraph{Visual speech refinement.}
Because log-derived action boundaries are often coarser than visible speech, we refine each session using mouth-motion cues from the video. Faces are detected in sampled frames, assigned to characters using reference images, and analyzed with lip landmarks. We compute a mouth aspect ratio,
\begin{equation}
\mathrm{MAR}
=
\frac{
\frac{1}{3}\sum_{(i,j)\in P}
\lVert \mathbf{x}_i - \mathbf{x}_j \rVert
}{
\lVert \mathbf{x}_{\mathrm{right}} - \mathbf{x}_{\mathrm{left}} \rVert
},
\label{eq:mar}
\end{equation}
where $P$ contains inner-lip landmark pairs and the denominator is the mouth-corner distance. Frames with sufficiently large mouth opening are treated as visual speech, while wide-smile cases are suppressed using mouth-width geometry. The visual speech intervals are then fused with the action-log session boundaries.

\paragraph{Speaker timeline construction.}
For each fused session, we construct a speaker timeline: an ordered sequence of speaker slots with start time, end time, speaker identity, and word budget. Visual speech intervals receive highest priority, while action-log rows provide softer evidence for who is likely speaking. Remaining gaps are filled by rotating among active participants, and very short slots are constrained to brief interjections.

\paragraph{Dialogue generation.}
We generate dialogue in two model-assisted steps using Qwen3.5-35B-A3B. First, the model captions the session clip, describing visible participants, facial expressions, mouth motion, interaction context, environment, and mood from Servo Bot's first-person viewpoint. Second, the model is prompted with the caption, speaker timeline, activity context, and character profiles to generate exactly one utterance per slot.

\paragraph{Speech synthesis and audio mixing.}
Each utterance is synthesized using Qwen3-TTS-12Hz-1.7B-CustomVoice\footnote{\url{https://huggingface.co/Qwen/Qwen3-TTS-12Hz-1.7B-CustomVoice}} with character-specific voice assignments and mixed into the video. Speech is speed-adjusted when necessary to preserve temporal alignment. Speaker volume is modulated by the distance to Servo Bot:
\begin{equation}
v(d) = \mathrm{clamp}\left(1 - \frac{d}{25}, 0, 1\right),
\label{eq:volume}
\end{equation}
with a soft lower gain floor to keep distant speech weak but audible.

\subsection{Prompted Dialogue Adaptation}
\label{app:dialogue:adaptation}

After producing the original dialogue track, we adapt selected sessions to support phenomena whose target rules cannot be fully inferred from visible frames alone. Dialogue adaptation is controlled through additional prompt instructions in the dialogue generation stage.

\paragraph{Invisible activity completion.}
Some phenomenon-relevant activities are not directly visible because they occur outside the camera view or during occlusion. In these cases, dialogue can make the missing activity inferable by referring to activities that have already happened, are currently being planned, or are about to happen. Such references must be consistent with the underlying action log and schedule, and are restricted to same-day activities to avoid revealing underlying patterns.

\paragraph{Dynamic Episodes.}
For \emph{dynamic episodes}, the household follows one pattern before a pivot day and another pattern after it. The pivot day contains a change-point event that explains or motivates this shift. During prompted dialogue adaptation, we inject this event into one of the pivot-day conversations without explicitly stating the full before--after rule. For example, a pivot-day dialogue session involving Mother Sim may mention that she is preparing for an art market, causing her to switch from painting every other day to painting every day.

\subsection{Constraints on Adapted Dialogue}
\label{app:dialogue:constraints}

Prompted adaptation must remain temporally aligned with the original speaker timeline, visually and causally consistent with the recorded video, action log, and character profiles, and must not directly reveal the benchmark answer. The dialogue may provide clues, reminders, or references to relevant events, but the target phenomenon must still be inferred from the full multimodal context.

\subsection{Quality Control}
\label{app:dialogue:quality}

We apply quality control at both the model-generation stage and the human-inspection stage. During original dialogue generation, the session caption is first generated from the video clip. The dialogue-generation model then receives this caption together with the speaker timeline, activity context, and game log information. Before generating the final utterances, the model is prompted to check whether the caption is consistent with the log-derived session context, including visible participants, activities, and temporal ordering. 

After dialogue generation, we further inspect generated sessions using our visualization platform. We conduct manual spot checks to verify that the dialogue is temporally aligned, visually plausible, and consistent with the underlying activity logs. This inspection is used to identify obvious captioning errors, speaker mismatches, timing issues, or dialogue that reveals more information than intended.

\section{\Ours{} Web Interface}
\label{sec:appendix-web-interface}

The following sections describe the features and use of the \Ours{} platform. 

We include three web interfaces to facilitate the data creation pipeline for \OursBench{}:

\begin{itemize}
    \item \textbf{Simulated Day Timeline:} Assists in parsing raw JSON schedules by combining individual characters' assigned schedules into a unified, side-by-side family timeline. This visual verification step ensures game mods correctly enforce scheduled actions.
    \item \textbf{Phenomenon Generation:} Provides an end-to-end platform to construct behavioral benchmarks. This page runs simulated day sampling, evaluates activity visibility each day, and verifies hypothesis elimination before adding video chains to the database.
    \item \textbf{Task Annotation:} A platform for human validation of data and ground-truth question insertion. Video playback is shown alongside character game logs to identify the right frame to insert the benchmark questions.
\end{itemize}

\subsection{Simulated Day Timeline}

\begin{figure*}[h]
    \centering
    \includegraphics[width=\linewidth]{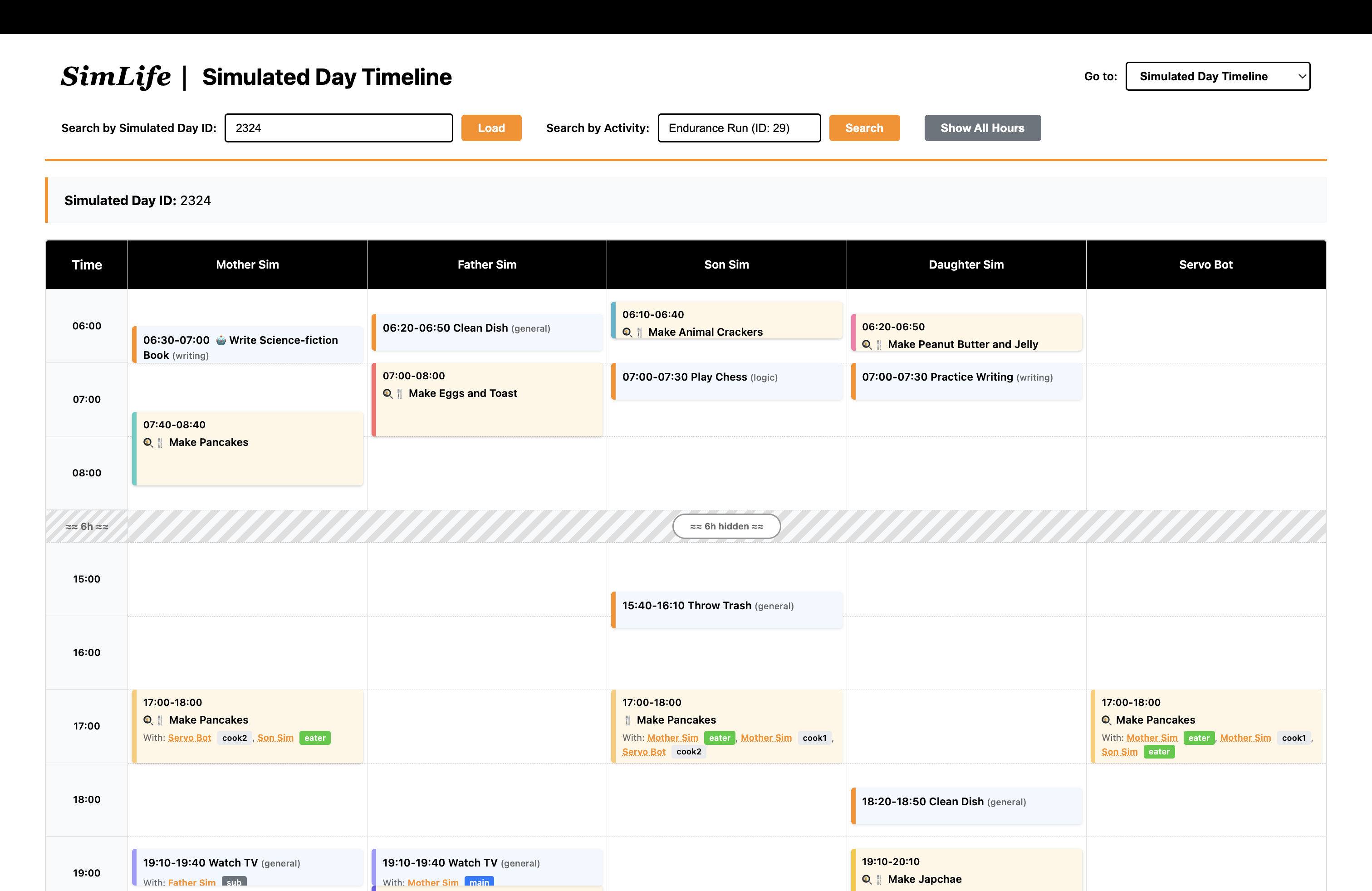}
\caption{
Timeline interface that displays the scheduled actions for a specified simulated day for each character in the game. This page will display the schedules the Sims characters will perform when we run the specified simulated day in the game.
}
\vspace*{-5pt}
    \label{fig:webinterface_simulated_days_schedule}
\end{figure*}

We created the ``Simulated Day Timeline'' page, as shown in Figure~\ref{fig:webinterface_simulated_days_schedule}, to view the simulated day schedules we generated for each character. Each block on the day schedule is an activity that we assigned to the character and we use game mods to ensure that the activity will happen around the time it is scheduled.  


Since we store the schedules in JSON files with each character's assigned activities as a list, it is difficult to visualize what the family as a whole will be doing in a single day when we run the game recording pipeline. This web interface allows us to verify if activities happen when we want them to, which is within the scheduled start and end time block, and also have a better idea of how the schedules of each family member will interact.

\subsection{Phenomenon Generation}

\begin{figure*}[h!]
    \centering
    \includegraphics[width=\linewidth]{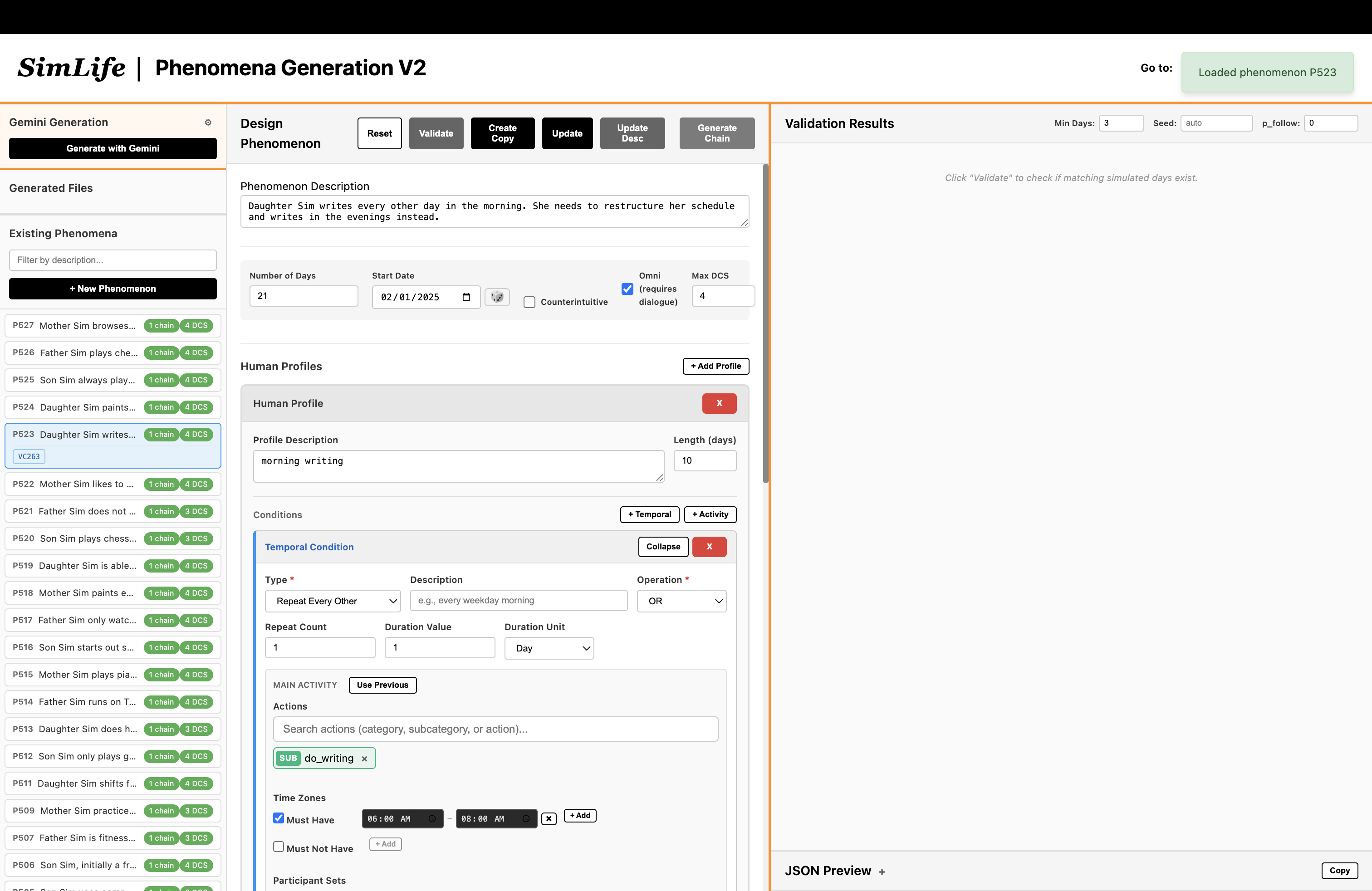}
    \caption{Phenomenon Generation page from the \Ours{} platform. This is what is shown when selecting a phenomenon from the left sidebar, before validating it. }
    \label{fig:webinterface_phenomenon_no_validation}
    \vspace{-5pt}
\end{figure*}

We use the Phenomenon Generation page to create the phenomenon, as described in Appendix~\ref{sec:appendix-phenomena}, and then generate video chains that fulfill the requirements of the phenomenon. This page allows us to build a phenomenon piece by piece while validating that it is possible to create a video chain data instance and adding new video chains to the dataset through an automatic simulated day video selection process.

\paragraph{Validating Phenomenon}

\begin{figure*}[h]
    \centering
    \includegraphics[width=0.47\linewidth]{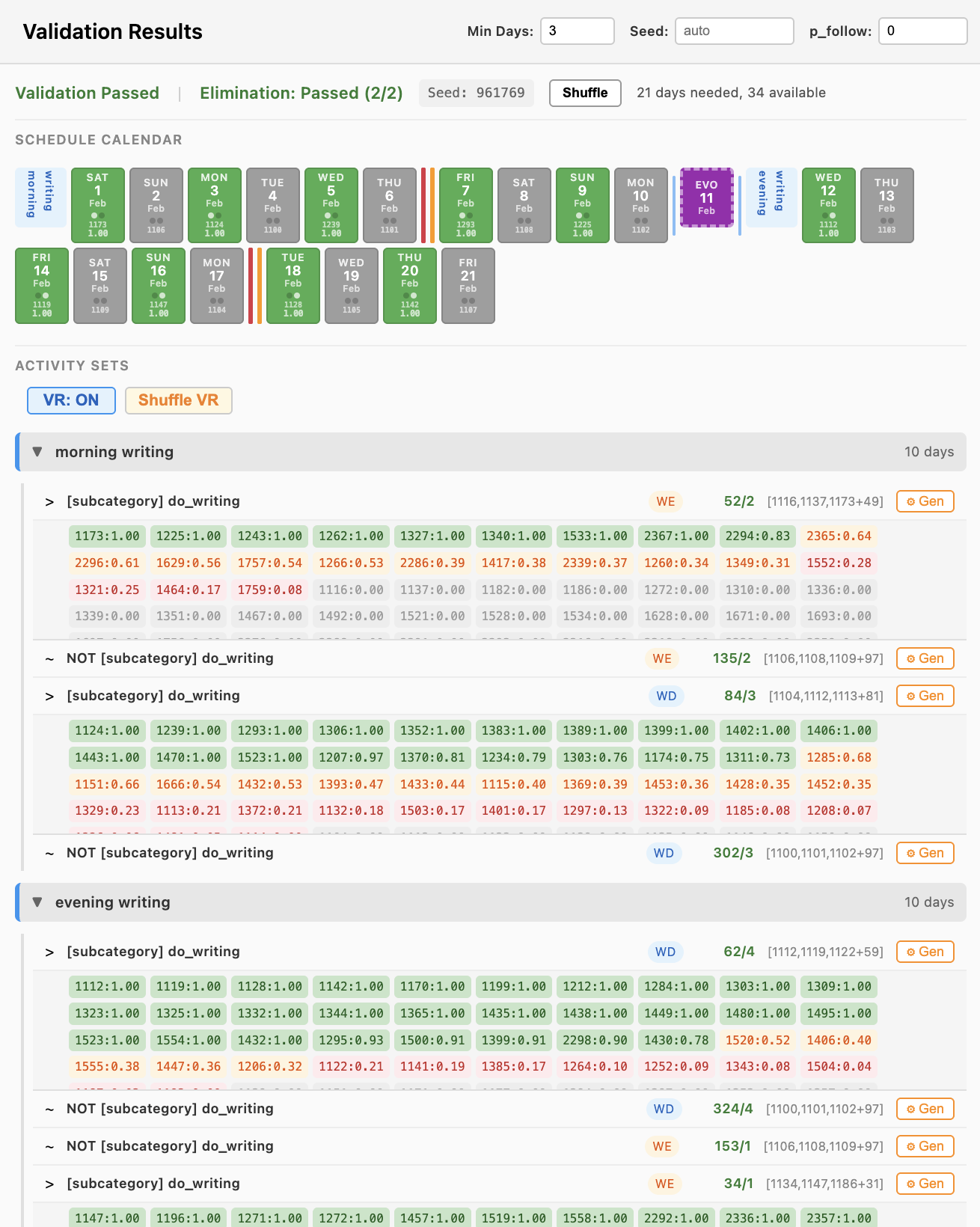}
    \includegraphics[width=0.47\linewidth]{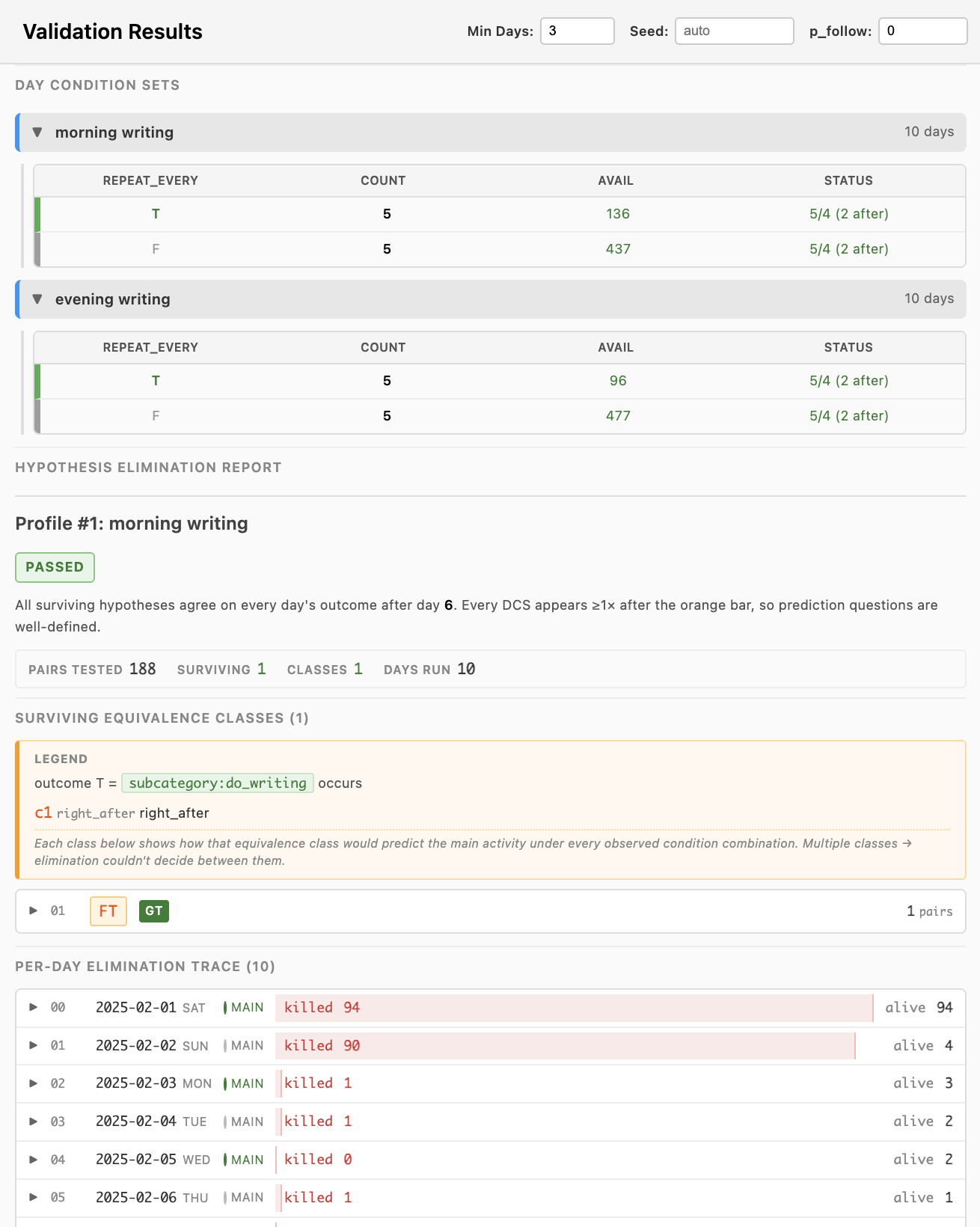}
\caption{
Output displayed on the Phenomenon Generation page after validation of a phenomenon. 
\textbf{Left:} The calendar for the specific video chain and available simulated days for a condition set. 
\textbf{Right:} Scrolling down will show the truth tables for day condition sets and hypothesis elimination trace.
}
    \label{fig:webinterface_phenomenon_validation}
\end{figure*}

The main utility from this page comes after clicking the ``Validate'' button for a phenomenon, where the backend will first perform the simulated day sampling outlined in Appendix~\ref{sec:appendix-phenomena} to create a video chain with the number of days and start date from the phenomenon. The days will be shown in a ``Schedule Calendar''.
Figure~\ref{fig:webinterface_phenomenon_validation} is the schedule calendar for a dynamic understanding phenomenon, where you can see that there is a purple ``EVO'' day, which represents a day separating the two human profiles and the day that a trigger event should be inserted. 

The activity sets shown below are the simulated day IDs that fulfill the condition requirements and the visibility score of each day. The first schedule calendar shown will always select the days with the highest visibility first. 



There are also two lines per human profile shown on the schedule calendar: the red line indicates when there are at least 3 occurrences for each day condition set, while the orange line indicates when hypothesis elimination passes. Either of these processes can fail, for example, there may not be enough simulated days to fill out the entire schedule calendar or hypothesis elimination isn't possible within the number of days for this schedule. The ``Generate Chain'' button will only be available when both pass and there is a red and orange line with at least one occurrence of each day condition set truth table assignment after that point. 

Then, clicking on ``Generate Chain'' will add the exact video chain with the simulated day IDs shown in the schedule calendar to the database.

\subsection{Task Annotation}

\begin{figure*}[h!]
    \centering
    \includegraphics[width=\linewidth]{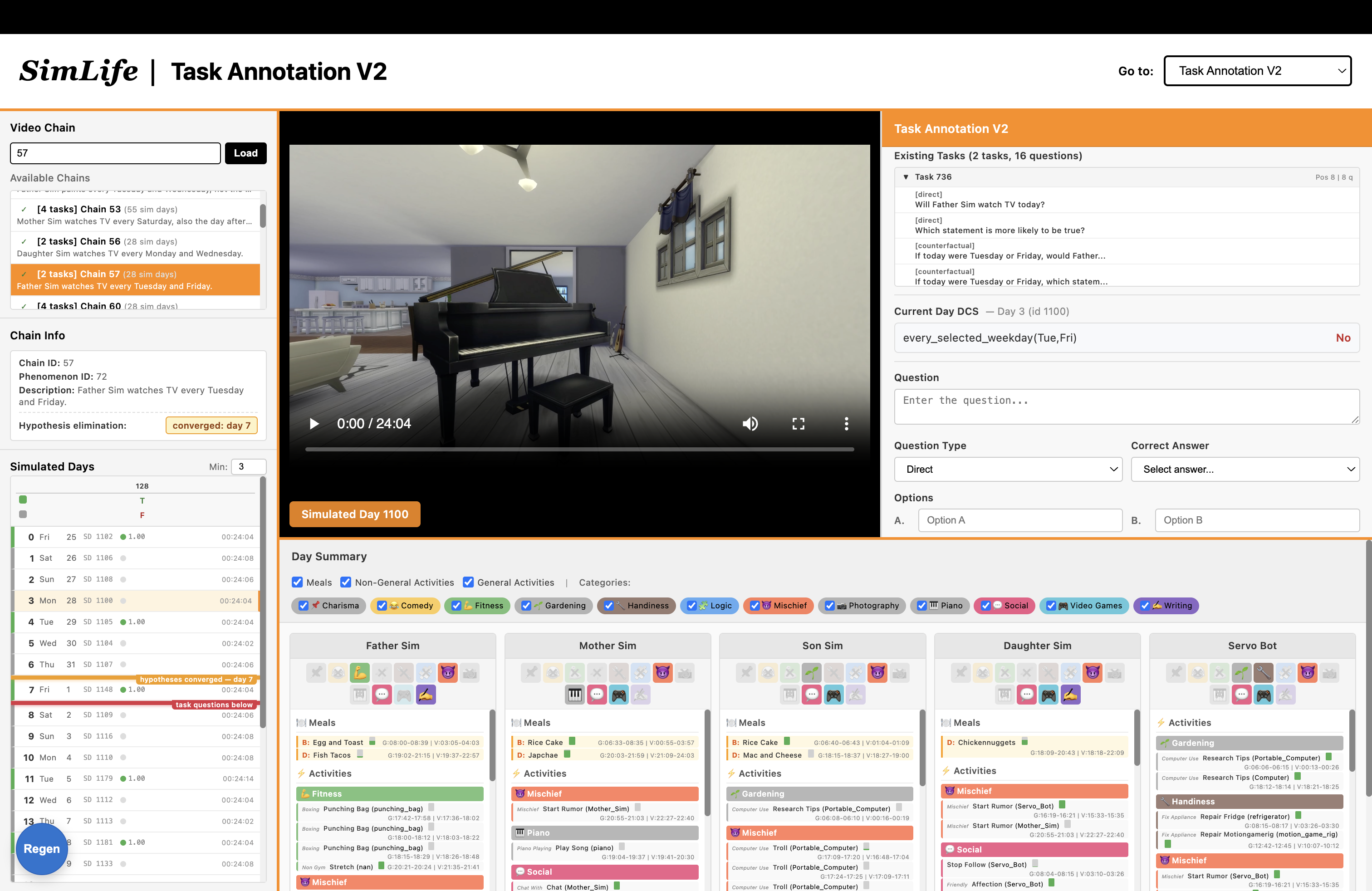}
    \caption{Task Annotation page from the \Ours{} platform. This is the appearance after selecting a video chain and a specific simulated day within the video chain in the left sidebar. }
    \label{fig:webinterface_task_generation}
    \vspace{-5pt}
\end{figure*}

The task annotation page allows us to insert tasks at a specific time in a video chain while watching the video to ensure visibility of activities. This page enables human validation of the video chain data instances that are generated from the phenomenon and easy insertion of tasks since all information is consolidated in one view. 


\paragraph{Day Summary}

The day summary includes all actions from the game logs, split by character. Each one is named and grouped based on the category of action. The green bar next to an action name is the visibility score of each activity. Clicking an activity will jump directly to the start time of that action in the video. This function can be used to identify the point in time in the video that is after a certain activity to find a place to insert a task. 

\paragraph{Inserting Tasks}

The ``Task Annotation'' section on the right of the video in Figure~\ref{fig:webinterface_task_generation} allows us to write a task question, select the type of question and correct answer, and label it with a simulated day and timestamp of the video which will be the end of the context provided to a model before the task question will be presented. 



Scrolling down on this section will show an interface to fill out information about the task question to be inserted. 




\begin{table*}[h!]
\centering
\small
\resizebox{\columnwidth}{!}{
\begin{tabular}{c l p{8.8cm} c}
\toprule
\textbf{Code} & \textbf{Type} & \textbf{Description} & \textbf{Days} \\
\midrule

\texttt{A0} & Anchor (easy) & Mother Sim plays piano in the evening every other day. & 10 \\

\texttt{A1} & Anchor (hard) & Daughter Sim writes everyday without painting in the same day. & 20 \\
\midrule

\texttt{N0} & Intuitive (temporal) & Mother Sim paints a canvas every Tuesday. & 28 \\

\texttt{N1} & Intuitive (activity) & Father Sim always drinks coffee the day after playing chess at night. & 21 \\

\texttt{N2} & Counterintuitive (activity) & Father Sim will do fitness and fishing in the same day. & 21 \\

\texttt{N3} & Counterintuitive (mixed) & Daughter Sim writes screenplays, is inspired immediately after watching TV. & 16 \\

\texttt{N4} & Dynamic & Father Sim runs on Tuesday and Thursday. He wants to participate in a half marathon and starts training every other day. & 29 \\

\texttt{N5} & Dynamic & Mother Sim paints every other day, then in order to prep for an art market, begins to paint every single day. & 29 \\

\texttt{N6} & Dynamic & Son Sim plays chess late at night every day. As he needs to wake up early in the morning due to a new club on weekdays, he only practices chess late at night on weekends. & 25 \\

\texttt{N7} & Dynamic & Daughter Sim is able to play motion games more often at the beginning of the school year, but as she gets more busy, plays motion games only on the weekends. & 29 \\

\bottomrule
\end{tabular}
}
\caption{Overview of the episodes used in the human study.}
\label{tab:human_study_video_chains}
\end{table*}

\begin{figure*}[h!]
    \centering
    \includegraphics[width=\linewidth]{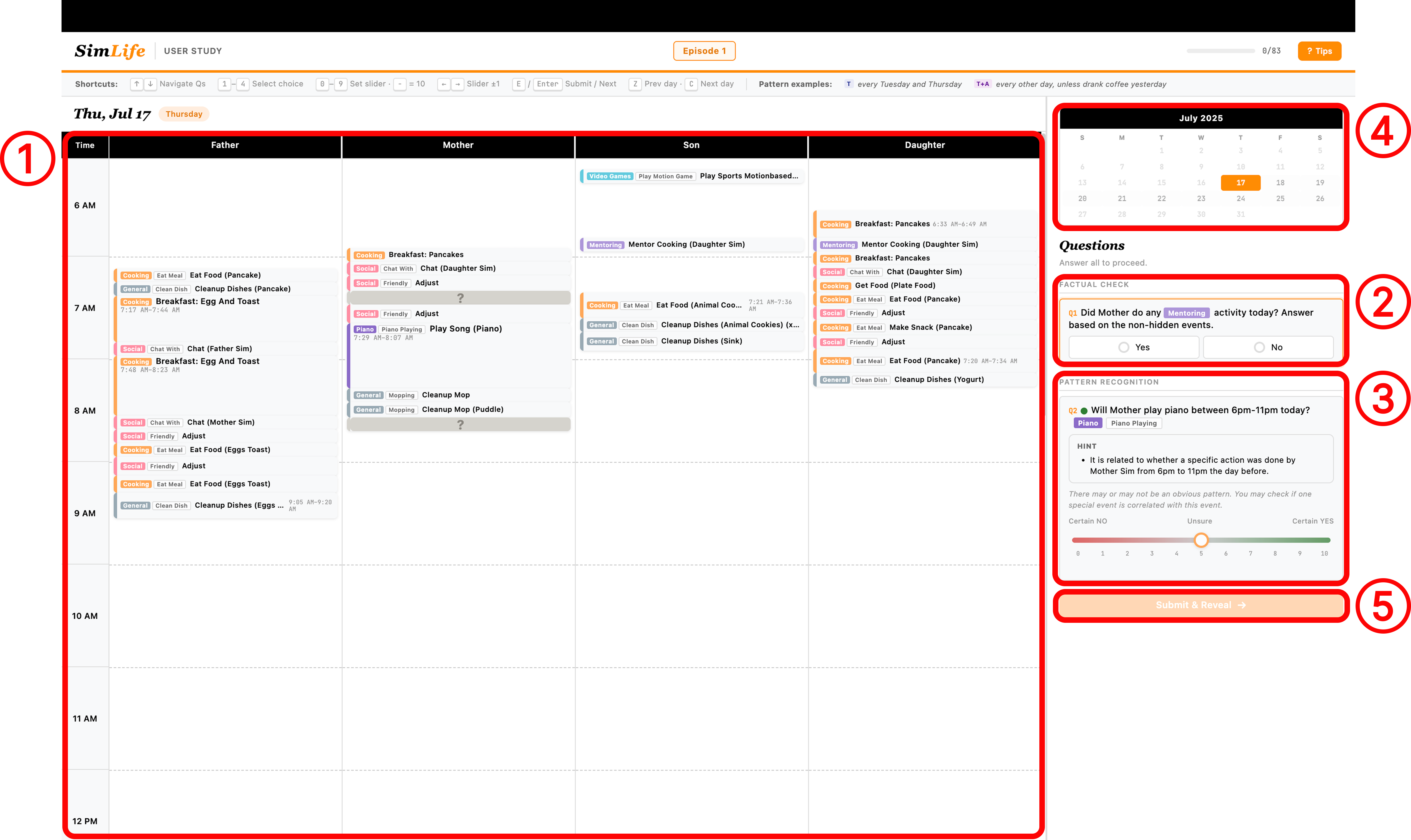}
    \caption{Overview of the human study interface. Some activity blocks are hidden from participants and shown as gray blocks.}
    \label{fig:human_study_interface}
    \vspace{-5pt}
\end{figure*}

\section{Human Study}
\label{sec:appendix:human_study_details}

We conduct the human study through a web-based interface. The study uses a subset of 10 video chains, listed in Table~\ref{tab:human_study_video_chains}. Each participant completes two shared anchor episodes, \texttt{anchor0} and \texttt{anchor1}, for calibration (Section~\ref{app:calibration}); the remaining episodes are distributed so that each is evaluated by three participants.

Each episode presents a household of four Sims over multiple simulated days. On each day, participants view a daily activity calendar with visible and hidden activity blocks, answer factual questions from the visible calendar, and rate their confidence about whether a target activity occurred. Participants first complete an interactive tutorial introducing the task, interface, and response format.

\subsection{User Interface}
Figure~\ref{fig:human_study_interface} shows the main interface. The daily calendar (marked \texttt{1} in Figure~\ref{fig:human_study_interface}) displays each Sim's activities for the current day, with hidden activities shown as unknown blocks. These hidden blocks may include the target pattern activity or other contextual activities.

For each day, participants answer factual questions (\texttt{2}) and pattern-prediction questions (\texttt{3}). Factual questions check attention to the visible schedule. Pattern-prediction responses are collected on an 11-point scale from 0 (``certainly no'') to 10 (``certainly yes''). Each prediction question is accompanied by a partial hint, described in Section~\ref{sec:benchmark_design}, which identifies a relevant class of evidence without revealing the full rule. For dynamic chains, a pop-up window notifies participants of the pattern change on the pivot day. A compact history view in the form of a calendar (\texttt{4}) summarizes whether the target activity occurred on previous days. After submission (\texttt{5}), hidden blocks are revealed and the target activity is highlighted when present, allowing participants to update their hypotheses over the episode.

\subsection{Anchor-Based Participant Calibration}
\label{app:calibration}

Participants may differ in learning speed and slider usage. To adjust for these participant-level effects, each participant completes two fixed anchor tasks designed to differ in difficulty: an easy anchor $A_0$ (\textit{``Mother plays piano every other evening''}) and a hard anchor $A_1$ (\textit{``Daughter writes on days when she does not paint''}).

On each day $d$, participant $p$ reports an integer slider response $\mathrm{slider}_{p,d}\in\{0,\dots,10\}$ indicating whether the target activity occurs on that day. We interpret this response as a probability $\mathrm{slider}_{p,d}/10\in[0,1]$, and denote the ground truth by $y_d\in\{0,1\}$. We fit a participant-specific slider mapping on the anchor responses:
$$
\frac{\mathrm{slider}_{p,d}}{10} = a_p\, y_d + b_p + \varepsilon_{p,d},
$$
where $a_p$ captures discriminability, i.e., the slider range used to separate $y_d=1$ from $y_d=0$ days, $b_p$ captures base-rate bias, i.e., the expected response on $y_d=0$ days, and $\varepsilon_{p,d}$ captures residual noise.

We also estimate a participant-specific learning-speed offset from anchor convergence days:
$$
D_{\mathrm{conv}}(p,t) = \mu_t + \alpha_p + \varepsilon_{p,t},
$$
where $D_{\mathrm{conv}}(p,t)$ is the convergence day of participant $p$ on anchor task $t$, $\mu_t$ is the mean convergence day for task $t$ (see~\ref{app:elim:procedure}), and $\alpha_p$ captures whether participant $p$ tends to converge earlier or later than average. We fit both calibration components using hierarchical models, so noisy participant-level estimates are shrunk toward the group mean.

The estimated parameters are then applied to each participant's non-anchor tasks. Each slider response is converted into a calibrated probability:
$$
\hat p_{p,t,d}
=
\mathrm{clip}\big((\mathrm{slider}_{p,t,d}/10-b_p)/a_p,\,0,1\big),
$$
and the day index is adjusted by the participant's learning-speed offset:
$$
\tilde d = d-\alpha_p.
$$
We then average calibrated curves across participants. 
The anchor tasks are excluded from the main human--model comparison.

\subsection{Results}

We present the human study results with representative models in Figures~\ref{fig:baseline_alignment_spearman} and~\ref{fig:baseline_N1-N6}.
In general, human participants perform well on both the intuitive cases (N0 and N1) and the dynamic cases (N4-N7).

\section{Additional Results}
\label{sec:appendix:additional_results}

\begin{figure*}[]
    \centering
    \includegraphics[width=\linewidth]{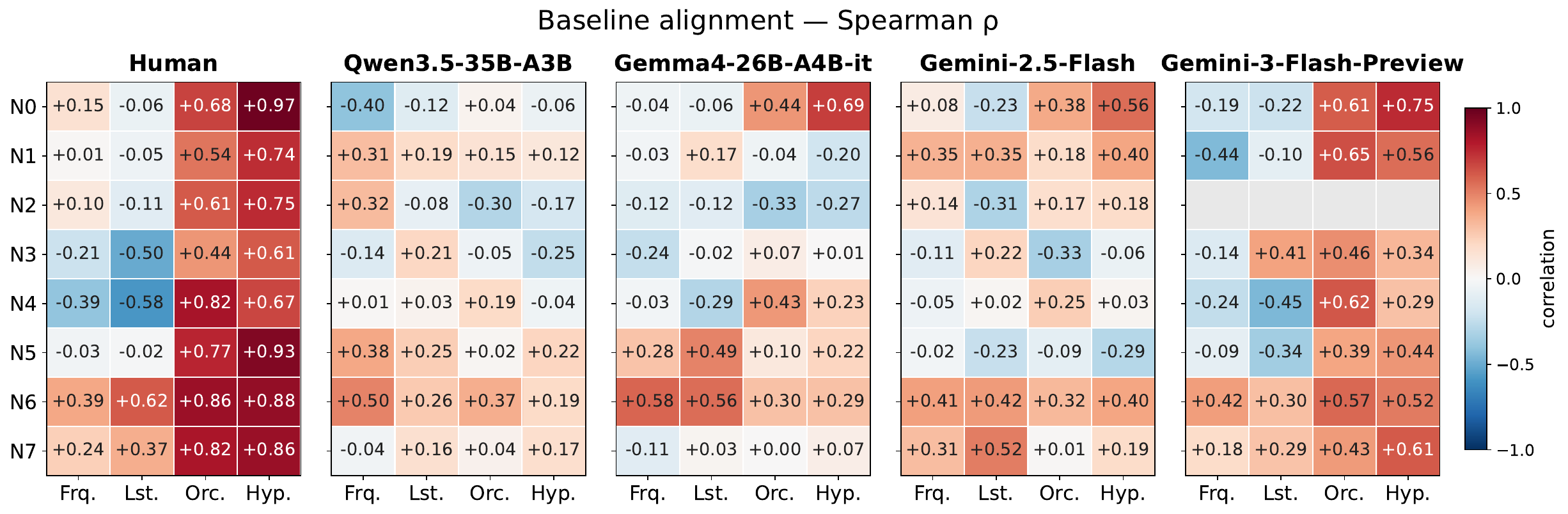}
    \vspace{-20pt}
    \caption{Baseline alignment of day-level prediction trajectories (Spearman $\rho$). Each panel corresponds to one path that produces a daily $P(\text{yes})$ trajectory for the target event: Human (per-person consolidated mean over the 12 participants, anchor-calibrated), Qwen3.5-35B-A3B, Gemma-4-26B-A4B-it, Gemini-2.5-Flash, and Gemini-3-Flash-Preview. Within a panel, each cell is the Spearman correlation between that path's trajectory and one of four reference baselines, computed over the days of a case. Rows N0--N7 are the eight non-anchor cases. Columns are the baselines: Frq.\ = frequency-based, predicting from the historical base rate; Lst.\ = last-day-based, extrapolating from the previous day’s outcome; Orc.\ = ground-truth oracle; Hyp.\ = hypothesis-elimination, pruning candidate rules against accumulated evidence. A cell is undefined (shown in grey) when the source's prediction is constant across that case's days, so the rank correlation has zero variance. Gemini-3-Flash-Preview emits only binary 0/1 probabilities and answers $P(\text{yes})=0$ on every one of the 21 days of N2 (``Father fitness + fishing same day''), never predicting ``yes''; consequently all four Gemini-3 cells in row N2 are undefined and shown grey. Spearman $\rho$ is rank-based (amplitude-free) and robust to the models' tendency to saturate at 0 and 1.
    }
    \label{fig:baseline_alignment_spearman}
\end{figure*}

\begin{figure*}[]
    \centering
    \includegraphics[width=0.9\linewidth]{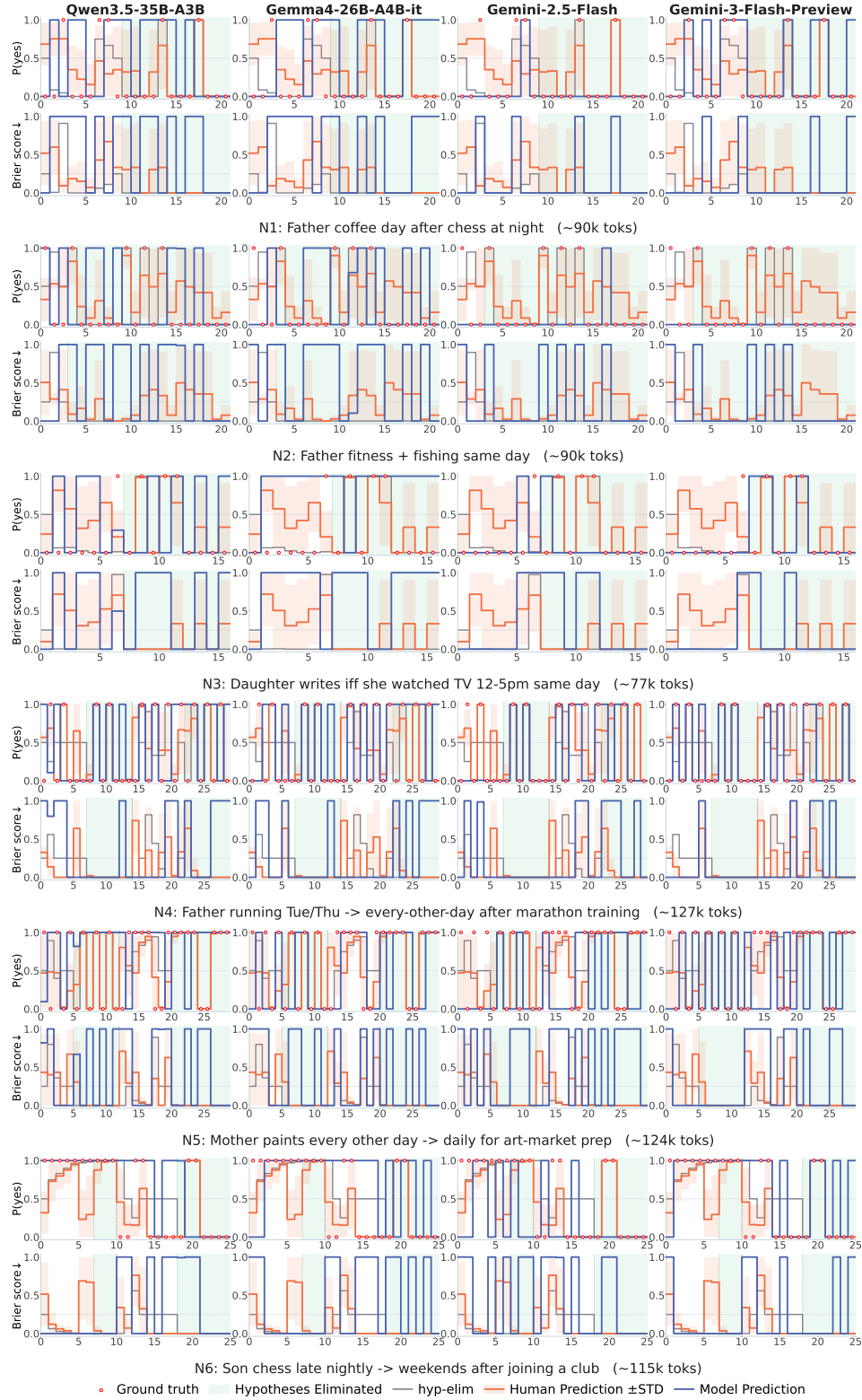}
    \definecolor{hypelimbg}{HTML}{EDF8F2}
    \vspace{-10pt}
    \caption{Additional per-day trajectories of model predictions and human judgments for other non-anchor cases (N1-N6). A rapid convergence of Brier score (bottom) toward zero after \colorbox{hypelimbg}{hypothesis elimination} indicates successful pattern understanding and retention.%
    }
    \label{fig:baseline_N1-N6}
\end{figure*}

We present additional experiment results in this section. Figure \ref{fig:baseline_alignment_spearman} shows Spearman correlation between the non-anchor chains used in the human study and the reference baselines for each model evaluated in RQ2 and the human study responses. Figure \ref{fig:baseline_N1-N6} shows the trajectories for non-anchor cases not mentioned in the body of the paper. Table \ref{tab:rq3_adaptation_full} is the full results for RQ3 on the dynamic understanding video chains. 

For all models used in this work, we strictly follow their licenses and terms of use.

\begin{table*}[]
  \centering
  \setlength{\aboverulesep}{1pt}\setlength{\belowrulesep}{1pt}\setlength{\tabcolsep}{3pt}
  \newcommand{\res}[2]{\ensuremath{#1^{\,\scalebox{0.7}{$\pm#2$}}}}
  \resizebox{\linewidth}{!}{%
  \begin{tabular}{@{}lcccccccccccc@{}}
  \toprule
  \multirow{2}{*}{Model} & \multirow{2}{*}{FPS} & \multirow{2}{*}{Thinking} &
  \multicolumn{3}{c}{Full Hints} & \multicolumn{3}{c}{Partial Hints} & \multicolumn{3}{c}{No Hints} \\
  \cmidrule(lr){4-6} \cmidrule(lr){7-9} \cmidrule(lr){10-12}
  & & & P1 & P2 & Hist & P1 & P2 & Hist & P1 & P2 & Hist \\
  \midrule
  \multicolumn{12}{@{}l}{\textit{Vision + Language (V+L)}} \\
  Qwen3-VL-4B-Instruct & 0.5 & CoT & \res{54.5}{8.1} & \res{56.7}{7.9} & \res{55.1}{6.6} & \res{57.2}{8.1} & \res{55.3}{8.0} & \res{50.9}{6.7} & \res{56.6}{8.1} & \res{54.0}{8.0} & \res{55.6}{6.6} \\
  Qwen3-VL-8B-Instruct & 0.5 & CoT & \res{55.9}{8.1} & \res{47.3}{8.0} & \res{55.6}{6.6} & \res{59.3}{8.0} & \res{60.0}{7.8} & \res{54.2}{6.6} & \res{56.6}{8.1} & \res{54.7}{8.0} & \res{50.5}{6.7} \\
  Qwen3-VL-32B-Instruct & 0.5 & CoT & \res{57.9}{8.0} & \res{54.7}{8.0} & \res{55.6}{6.6} & \res{59.3}{8.0} & \res{60.0}{7.8} & \res{56.5}{6.6} & \res{55.2}{8.1} & \res{56.7}{7.9} & \res{56.9}{6.6} \\
  Qwen3.5-9B & 0.5 & R & \res{49.0}{8.1} & \res{50.7}{8.0} & \res{57.4}{6.6} & \res{55.9}{8.1} & \res{57.3}{7.9} & \res{56.5}{6.6} & \res{49.7}{8.1} & \res{55.3}{8.0} & \res{56.0}{6.6} \\
  Qwen3.5-35B-A3B & 0.5 & R & \res{53.8}{8.1} & \res{48.0}{8.0} & \res{57.9}{6.6} & \res{57.9}{8.0} & \res{56.7}{7.9} & \res{57.9}{6.6} & \res{56.6}{8.1} & \res{55.3}{8.0} & \res{54.2}{6.6} \\
  Gemma-4-26B-A4B-it & 0.5 & R & \res{49.0}{8.1} & \res{54.7}{8.0} & \res{47.2}{6.7} & \res{54.5}{8.1} & \res{49.3}{8.0} & \res{45.8}{6.6} & \res{56.6}{8.1} & \res{52.0}{8.0} & \res{49.5}{6.7} \\
  Gemini-3-flash-preview & 0.5 & R & \res{51.7}{8.1} & \res{48.7}{8.0} & \res{57.4}{6.6} & \res{57.9}{8.0} & \res{56.7}{7.9} & \res{55.6}{6.6} & \res{56.6}{8.1} & \res{55.3}{8.0} & \res{55.6}{6.6} \\
  VideoTree (Mixtral-8x7B) & 1 & CoT & \res{55.9}{8.1} & \res{52.0}{8.0} & \res{57.9}{6.6} & \res{48.3}{8.1} & \res{51.3}{8.0} & \res{59.3}{6.6} & \res{54.5}{8.1} & \res{54.0}{8.0} & \res{55.1}{6.6} \\
  \midrule
  \multicolumn{12}{@{}l}{\textit{Vision + Audio (V+A)}} \\
  Qwen3-Omni-30B-A3B \small{+ Qwen3.5-9B} & 0.5 & R & \res{45.5}{8.1} & \res{46.7}{8.0} & \res{51.9}{6.7} & \res{51.0}{8.1} & \res{51.3}{8.0} & \res{53.7}{6.6} & \res{50.3}{8.1} & \res{50.0}{8.0} & \res{49.5}{6.7} \\
  M3-Agent & 1 & R & \res{49.0}{8.1} & \res{55.3}{8.0} & \res{57.9}{6.6} & \res{60.7}{8.0} & \res{56.7}{7.9} & \res{57.9}{6.6} & \res{53.1}{8.1} & \res{57.3}{7.9} & \res{58.8}{6.6} \\
  \midrule
  \multicolumn{12}{@{}l}{\textit{Language Only (L)}} \\
  Qwen3.5-9B & -- & R & \res{48.3}{8.1} & \res{45.3}{8.0} & \res{50.5}{6.7} & \res{54.5}{8.1} & \res{42.7}{7.9} & \res{51.9}{6.7} & \res{63.4}{7.8} & \res{51.3}{8.0} & \res{53.7}{6.6} \\
  Qwen3.5-35B-A3B & -- & R & \res{71.0}{7.4} & \res{51.3}{8.0} & \res{69.0}{6.2} & \res{63.4}{7.8} & \res{48.7}{8.0} & \res{65.3}{6.3} & \res{62.8}{7.9} & \res{54.0}{8.0} & \res{63.0}{6.4} \\
  Gemma-4-26B-A4B-it & -- & R & \res{64.1}{7.8} & \res{45.3}{8.0} & \res{53.7}{6.6} & \res{66.2}{7.7} & \res{51.3}{8.0} & \res{51.9}{6.7} & \res{68.3}{7.6} & \res{47.3}{8.0} & \res{50.0}{6.7} \\
  Gemini-2.5-flash & -- & R & \res{69.7}{7.5} & \res{53.3}{8.0} & \res{59.7}{6.5} & \res{67.6}{7.6} & \res{47.3}{8.0} & \res{63.9}{6.4} & \res{57.9}{8.0} & \res{52.7}{8.0} & \res{63.4}{6.4} \\
  Gemini-3-flash-preview & -- & R & \res{82.1}{6.2} & \res{63.3}{7.7} & \res{71.8}{6.0} & \res{86.9}{5.5} & \res{61.3}{7.8} & \res{74.5}{5.8} & \res{75.9}{7.0} & \res{64.0}{7.7} & \res{74.5}{5.8} \\
  Gemini-3.1-flash-lite & -- & R & \res{71.7}{7.3} & \res{52.7}{8.0} & \res{65.7}{6.3} & \res{75.2}{7.0} & \res{48.7}{8.0} & \res{59.3}{6.6} & \res{60.7}{8.0} & \res{52.0}{8.0} & \res{64.8}{6.4} \\
  \midrule
  \multicolumn{12}{@{}l}{\textit{Language only, noiseless (L-N)}} \\
  Qwen3.5-9B & -- & R & \res{80.7}{6.4} & \res{56.0}{7.9} & \res{71.8}{6.0} & \res{73.1}{7.2} & \res{58.0}{7.9} & \res{69.0}{6.2} & \res{68.3}{7.6} & \res{50.0}{8.0} & \res{66.7}{6.3} \\
  Qwen3.5-35B-A3B & -- & R & \res{85.5}{5.7} & \res{72.7}{7.1} & \res{79.6}{5.4} & \res{85.5}{5.7} & \res{65.3}{7.6} & \res{77.3}{5.6} & \res{74.5}{7.1} & \res{55.3}{8.0} & \res{72.7}{5.9} \\
  Gemma-4-26B-A4B-it & -- & R & \res{81.4}{6.3} & \res{73.3}{7.1} & \res{72.7}{5.9} & \res{73.1}{7.2} & \res{72.0}{7.2} & \res{72.7}{5.9} & \res{74.5}{7.1} & \res{64.7}{7.6} & \res{72.2}{6.0} \\
  Gemma-4-31B-it & -- & R & \res{84.1}{5.9} & \res{72.7}{7.1} & \res{78.2}{5.5} & \res{82.1}{6.2} & \res{64.7}{7.6} & \res{77.8}{5.5} & \res{84.1}{5.9} & \res{67.3}{7.5} & \res{76.4}{5.7} \\
  Gemini-2.5-flash & -- & R & \res{86.9}{5.5} & \res{68.0}{7.5} & \res{73.6}{5.9} & \res{85.5}{5.7} & \res{57.3}{7.9} & \res{67.6}{6.2} & \res{82.1}{6.2} & \res{62.0}{7.8} & \res{70.8}{6.1} \\
  Gemini-3-flash-preview & -- & R & \res{77.2}{6.8} & \res{62.0}{7.8} & \res{77.8}{5.5} & \res{79.3}{6.6} & \res{65.3}{7.6} & \res{76.4}{5.7} & \res{75.9}{7.0} & \res{62.0}{7.8} & \res{76.9}{5.6} \\
  \bottomrule
  \end{tabular}%
  }
  \caption{Pattern-adaptation breakdown on the \Ours{} dynamic chains (20 chains; question-level accuracy (\%) with 95\% binomial CI half-width $\pm$). \textbf{P1}/\textbf{P2}: normal tasks whose queried day falls before/after the pattern switch (answering about the active regime); \textbf{Hist}: the history tasks, which query Pattern-1 behavior after the model has already observed Pattern~2. Unanswered or overflow responses count as incorrect. Full-context language models (L, L-N) generally show $\text{P1}\!>\!\text{Hist}\!>\!\text{P2}$, whereas vision, retrieval, and memory baselines remain near chance with no such structure.}
  \label{tab:rq3_adaptation_full}
\end{table*}

The detailed version of the main result in Table~\ref{tab:main_results} is shown in Table~\ref{tab:main_results_full}.

\begin{table*}[t]
\centering
\setlength{\aboverulesep}{1pt}
\setlength{\belowrulesep}{1pt}
\setlength{\tabcolsep}{2.5pt}
\definecolor{best}{HTML}{B7D8EF}
\definecolor{second}{HTML}{E2EEF8}
\definecolor{cD}{HTML}{0072B2}
\definecolor{cCF}{HTML}{E69F00}
\definecolor{cNC}{HTML}{009E73}
\definecolor{cIC}{HTML}{CC79A7}
\providecommand{\res}[2]{\ensuremath{#1^{\,\scalebox{0.7}{$\pm#2$}}}}
\resizebox{\linewidth}{!}{%
\begin{tabular}{@{}lcccccccccccccccccccc@{}}
\toprule
\multirow{2}{*}{Model} & \multirow{2}{*}{FPS} & \multirow{2}{*}{Thinking} &
\multicolumn{6}{c}{Full Hints} & \multicolumn{6}{c}{Partial Hints} & \multicolumn{6}{c}{No Hints} \\
\cmidrule(lr){4-9} \cmidrule(lr){10-15} \cmidrule(lr){16-21}
& & & AC & Acc & \textcolor{cD}{Dir.} & \textcolor{cCF}{CF} & \textcolor{cNC}{N-CF} & \textcolor{cIC}{I-CF} & AC & Acc & \textcolor{cD}{Dir.} & \textcolor{cCF}{CF} & \textcolor{cNC}{N-CF} & \textcolor{cIC}{I-CF} & AC & Acc & \textcolor{cD}{Dir.} & \textcolor{cCF}{CF} & \textcolor{cNC}{N-CF} & \textcolor{cIC}{I-CF} \\
\midrule
\multicolumn{21}{@{}l}{\textit{Vision + Language (V+L)}} \\
Qwen3-VL-4B-Instruct & 0.5 & CoT &
  \cellcolor{second}\res{10.5}{3.1} & \res{49.8}{2.6} & 46.6 & 55.1 & 48.6 & 48.1 &
  \cellcolor{best}\res{11.5}{3.2} & \cellcolor{second}\res{50.0}{2.6} & 48.9 & 52.8 & 50.4 & 47.8 &
  \res{9.4}{2.9} & \cellcolor{best}\res{50.3}{2.6} & 50.5 & 53.3 & 49.6 & 47.8 \\
Qwen3-VL-8B-Instruct & 0.5 & CoT &
  \cellcolor{best}\res{12.6}{3.3} & \cellcolor{best}\res{50.1}{2.6} & 46.3 & 54.1 & 48.3 & 51.1 &
  \cellcolor{second}\res{9.7}{3.0} & \cellcolor{second}\res{48.6}{2.6} & 44.3 & 51.4 & 49.1 & 48.9 &
  \res{7.6}{2.7} & \res{48.0}{2.6} & 48.2 & 49.9 & 48.3 & 45.7 \\
Qwen3-VL-32B-Instruct & 0.5 & CoT &
  \res{11.0}{3.1} & \res{51.9}{2.6} & 54.7 & 53.5 & 50.7 & 49.2 &
  \cellcolor{best}\res{13.6}{3.4} & \cellcolor{best}\res{53.2}{2.6} & 53.1 & 53.0 & 53.0 & 53.8 &
  \cellcolor{second}\res{12.1}{3.3} & \cellcolor{second}\res{53.0}{2.6} & 51.8 & 55.6 & 52.8 & 51.4 \\
Qwen3.5-9B & 0.5 & R &
  \res{12.3}{3.3} & \res{48.9}{2.6} & 47.2 & 52.0 & 50.1 & 45.9 &
  \cellcolor{best}\res{14.2}{3.5} & \cellcolor{best}\res{51.4}{2.6} & 51.1 & 53.8 & 52.0 & 48.4 &
  \cellcolor{second}\res{13.4}{3.4} & \cellcolor{second}\res{51.0}{2.6} & 52.4 & 54.1 & 50.7 & 47.0 \\
Qwen3.5-35B-A3B & 0.5 & R &
  \cellcolor{second}\res{13.1}{3.4} & \res{52.2}{2.6} & 50.2 & 56.7 & 53.3 & 48.1 &
  \cellcolor{best}\res{14.2}{3.5} & \cellcolor{best}\res{53.3}{2.6} & 53.7 & 55.6 & 50.7 & 53.2 &
  \res{12.3}{3.3} & \cellcolor{second}\res{52.5}{2.6} & 53.4 & 55.4 & 47.8 & 53.5 \\
Gemma-4-26B-A4B-it & 0.5 & R &
  \cellcolor{second}\res{7.6}{2.7} & \cellcolor{second}\res{48.8}{2.6} & 50.8 & 47.8 & 45.7 & 51.4 &
  \cellcolor{second}\res{7.6}{2.7} & \res{48.1}{2.6} & 50.5 & 50.9 & 45.7 & 45.7 &
  \cellcolor{best}\res{10.8}{3.1} & \cellcolor{best}\res{49.3}{2.6} & 50.2 & 53.0 & 45.7 & 48.4 \\
Gemini-3-flash-preview & 0.5 & R &
  \cellcolor{best}\res{14.7}{3.6} & \res{53.0}{2.6} & 53.7 & 59.3 & 52.2 & 46.5 &
  \res{13.9}{3.5} & \cellcolor{second}\res{54.1}{2.6} & 56.7 & 57.2 & 54.3 & 48.6 &
  \cellcolor{second}\res{14.2}{3.5} & \cellcolor{best}\res{54.8}{2.6} & 53.7 & 56.2 & 55.9 & 53.0 \\
VideoTree (Mixtral-8x7B) & 1 & CoT &
  \cellcolor{best}\res{11.0}{3.1} & \cellcolor{best}\res{54.0}{2.6} & 45.9 & 56.2 & 54.6 & 57.8 &
  \res{9.7}{3.0} & \res{51.7}{2.6} & 46.3 & 53.0 & 56.4 & 50.0 &
  \cellcolor{second}\res{10.8}{3.1} & \cellcolor{second}\res{52.6}{2.6} & 49.8 & 52.0 & 54.6 & 53.5 \\
\midrule
\multicolumn{21}{@{}l}{\textit{Vision + Audio (V+A)}} \\
Qwen3-Omni-30B-A3B \small{+ Qwen3.5-9B} & 0.5 & R &
  \cellcolor{second}\res{10.8}{3.1} & \cellcolor{second}\res{48.1}{2.6} & 45.3 & 50.9 & 49.6 & 45.9 &
  \cellcolor{best}\res{11.0}{3.1} & \cellcolor{best}\res{48.6}{2.6} & 48.9 & 49.9 & 51.7 & 44.1 &
  \cellcolor{second}\res{10.8}{3.1} & \cellcolor{best}\res{48.6}{2.6} & 47.6 & 50.1 & 47.8 & 48.9 \\
M3-Agent & 1 & R &
  \res{10.0}{3.0} & \res{53.4}{2.6} & 46.3 & 55.4 & 50.7 & 60.0 &
  \cellcolor{second}\res{10.2}{3.0} & \cellcolor{best}\res{55.0}{2.6} & 49.2 & 57.0 & 54.3 & 58.4 &
  \cellcolor{best}\res{12.1}{3.3} & \cellcolor{second}\res{54.6}{2.6} & 48.2 & 55.9 & 55.9 & 57.0 \\
\midrule
\multicolumn{21}{@{}l}{\textit{Language Only (L)}} \\
Qwen3.5-9B & -- & R &
  \res{9.2}{2.9} & \cellcolor{second}\res{50.9}{2.6} & 46.9 & 62.5 & 45.4 & 48.1 &
  \cellcolor{second}\res{13.6}{3.4} & \cellcolor{best}\res{54.6}{2.6} & 55.0 & 61.4 & 53.8 & 48.1 &
  \cellcolor{best}\res{14.2}{3.5} & \cellcolor{best}\res{54.6}{2.6} & 50.2 & 61.9 & 53.8 & 51.6 \\
Qwen3.5-35B-A3B & -- & R &
  \cellcolor{second}\res{18.1}{3.9} & \cellcolor{best}\res{61.1}{2.5} & 54.1 & 75.1 & 53.5 & 60.3 &
  \cellcolor{best}\res{18.4}{3.9} & \cellcolor{second}\res{59.3}{2.5} & 59.3 & 71.4 & 51.7 & 54.6 &
  \res{16.0}{3.7} & \res{58.7}{2.5} & 49.5 & 74.0 & 53.5 & 55.9 \\
Gemma-4-26B-A4B-it & -- & R &
  \cellcolor{best}\res{11.3}{3.2} & \res{53.9}{2.6} & 52.8 & 61.7 & 47.8 & 53.0 &
  \cellcolor{second}\res{11.0}{3.1} & \cellcolor{second}\res{54.4}{2.6} & 58.0 & 55.6 & 52.5 & 52.2 &
  \cellcolor{best}\res{11.3}{3.2} & \cellcolor{best}\res{54.8}{2.6} & 56.4 & 62.2 & 52.2 & 48.4 \\
Gemini-2.5-flash & -- & R &
  \cellcolor{second}\res{17.8}{3.8} & \cellcolor{second}\res{58.6}{2.5} & 57.3 & 69.8 & 49.9 & 57.0 &
  \cellcolor{best}\res{21.0}{4.1} & \cellcolor{best}\res{60.2}{2.5} & 59.3 & 70.1 & 54.6 & 56.5 &
  \res{16.3}{3.7} & \res{56.4}{2.6} & 51.1 & 68.5 & 50.1 & 54.9 \\
Gemini-3-flash-preview & -- & R &
  \cellcolor{second}\res{27.8}{4.5} & \cellcolor{second}\res{69.2}{2.4} & 72.3 & 81.1 & 62.5 & \textbf{\underline{61.4}} &
  \cellcolor{best}\res{\textbf{\underline{33.3}}}{4.7} & \cellcolor{best}\res{69.4}{2.4} & 71.0 & 81.1 & 64.8 & \textbf{\underline{60.5}} &
  \cellcolor{second}\res{27.8}{4.5} & \res{66.5}{2.4} & 60.9 & 81.1 & 59.1 & \textbf{\underline{63.8}} \\
Gemini-3.1-flash-lite & -- & R &
  \cellcolor{second}\res{20.7}{4.1} & \cellcolor{second}\res{59.9}{2.5} & 59.9 & 68.0 & 57.5 & 54.1 &
  \cellcolor{best}\res{21.0}{4.1} & \cellcolor{best}\res{61.2}{2.5} & 57.0 & 73.5 & 55.4 & 57.8 &
  \res{18.6}{3.9} & \res{59.1}{2.5} & 52.4 & 71.9 & 53.8 & 56.8 \\
\midrule
\multicolumn{21}{@{}l}{\textit{Language only, noiseless (L-N)}} \\
Qwen3.5-9B & -- & R &
  \cellcolor{best}\res{23.6}{4.3} & \cellcolor{best}\res{65.5}{2.5} & 70.4 & 79.3 & 62.2 & 50.5 &
  \cellcolor{second}\res{18.4}{3.9} & \cellcolor{second}\res{61.2}{2.5} & 64.5 & 76.9 & 57.5 & 46.2 &
  \res{16.0}{3.7} & \res{56.5}{2.6} & 59.0 & 74.0 & 51.2 & 41.9 \\
Qwen3.5-35B-A3B & -- & R &
  \cellcolor{best}\res{34.9}{4.8} & \cellcolor{best}\res{73.0}{2.3} & 76.9 & 84.0 & 70.9 & 60.5 &
  \cellcolor{second}\res{32.0}{4.7} & \cellcolor{second}\res{\textbf{\underline{70.3}}}{2.4} & 71.7 & 84.0 & 66.4 & 58.9 &
  \res{21.5}{4.1} & \res{62.7}{2.5} & 61.2 & 81.9 & 56.2 & 50.8 \\
Gemma-4-26B-A4B-it & -- & R &
  \cellcolor{best}\res{30.2}{4.6} & \cellcolor{best}\res{72.1}{2.3} & 82.4 & \textbf{\underline{86.9}} & 64.8 & 55.7 &
  \cellcolor{second}\res{26.0}{4.4} & \cellcolor{second}\res{68.4}{2.4} & \textbf{\underline{75.6}} & 84.3 & 62.5 & 52.2 &
  \res{22.3}{4.2} & \res{64.1}{2.5} & 69.4 & 82.4 & 54.9 & 50.5 \\
Gemma-4-31B-it & -- & R &
  \cellcolor{best}\res{\textbf{\underline{36.2}}}{4.8} & \cellcolor{best}\res{\textbf{\underline{74.6}}}{2.2} & \textbf{\underline{83.4}} & 86.6 & \textbf{\underline{74.8}} & 54.9 &
  \cellcolor{second}\res{30.2}{4.6} & \cellcolor{second}\res{69.6}{2.4} & 75.2 & \textbf{\underline{86.4}} & 66.4 & 51.1 &
  \res{\textbf{\underline{29.4}}}{4.6} & \res{\textbf{\underline{67.6}}}{2.4} & \textbf{\underline{72.6}} & \textbf{\underline{84.8}} & 61.7 & 51.9 \\
Gemini-2.5-flash & -- & R &
  \cellcolor{best}\res{31.5}{4.7} & \cellcolor{best}\res{71.8}{2.3} & 80.8 & 82.9 & 72.4 & 52.2 &
  \cellcolor{second}\res{22.3}{4.2} & \cellcolor{second}\res{65.3}{2.5} & 70.0 & 80.8 & 64.0 & 46.8 &
  \res{22.0}{4.2} & \res{63.7}{2.5} & 64.8 & 80.3 & 57.0 & 52.4 \\
Gemini-3-flash-preview & -- & R &
  \cellcolor{second}\res{24.1}{4.3} & \cellcolor{best}\res{68.2}{2.4} & 76.9 & 81.1 & 73.5 & 42.2 &
  \cellcolor{best}\res{24.9}{4.3} & \cellcolor{second}\res{67.3}{2.4} & 74.6 & 80.1 & \textbf{\underline{68.5}} & 46.8 &
  \res{23.1}{4.2} & \res{63.9}{2.5} & 65.5 & 80.3 & \textbf{\underline{63.8}} & 45.7 \\
\bottomrule
\end{tabular}%
}
\caption{Full breakdown of the main results (Table~\ref{tab:main_results}) on \OursBench{} (381 tasks / 1,439 questions per hint level). \texttt{AC} is the All-Correct rate and \texttt{Acc} the question-level accuracy, both with 95\% binomial CI half-widths. The remaining columns give the accuracy on each question type, i.e., the values drawn as bars in Table~\ref{tab:main_results}: \textcolor{cD}{Dir.} = Direct ($n{=}307$), \textcolor{cCF}{CF} = Counterfactual ($n{=}381$), \textcolor{cNC}{N-CF} = Noise-Counterfactual ($n{=}381$), \textcolor{cIC}{I-CF} = Inverse-Counterfactual ($n{=}370$), where $n$ is the number of questions of that type per hint level. Shading highlights the within-row best and second-best hint variant for \texttt{AC} and \texttt{Acc}; \underline{\textbf{bold}} marks the overall best in each column.}
\label{tab:main_results_full}
\end{table*}

\section{Case Study Analysis of Model Reasoning Traces}
\label{app:case_studies}

We examine model reasoning traces to understand how models arrive at their predictions. We have full thinking traces for Gemma and Qwen model families, whereas the Gemini models provide summarized reasoning as part of the final output.

\subsection{Hint Levels and Input Modalities}

We start by examining the effects of the three hint levels provided in our evaluation on the reasoning of the models. 

\paragraph{No Hints.} The most common failure, especially for Qwen and Gemini-2.5, is treating the truncated log for the target day as a complete record. Models then simply search that log for the activity. Any use of the history is only to gauge how often it occurs rather than to compare weekdays or find a pattern. As a result, they rarely answer Yes, citing insufficient evidence.

\paragraph{Partial Hints.} These direct attention to the specific weekdays where the target activity happens. Models are more likely to build a table of activities by weekday and check whether the target activity matches the day in question. Partial hints also largely eliminate the truncated-log confusion, so part of their performance gain comes from clarifying which day is being predicted.

\paragraph{Full Hints.} Models treat the hint as a rule to verify. The cleanest reasoning identifies the current weekday, compares it with the activities named in the hint, and verifies the pattern in the history. A new failure appears with two-condition hints: models check only one condition or read an OR as an AND. Gemini-3-flash-preview uses hints most cleanly. It never confuses the hinted day with the target day, keeps false Yes predictions near zero, and never refers to the hint explicitly; the hint's influence shows only in the variables its reasoning attends to.

\paragraph{Interaction with input modality.} Hints will provide the correct variables which determine \emph{which} evidence a model looks for, but the input determines whether that evidence is available:
\begin{itemize}
    \item \textbf{L-N:} The history is complete and structured, so Full Hints reliably produce correct rule application.
    \item \textbf{L:} The evidence exists but is buried in noise. Some models find it (Gemini-3 most consistently); others fall back on a surface reading of the hint.
    \item \textbf{V+L:} The text diaries built from video usually lack the deciding evidence.
\end{itemize}




\paragraph{Case Study: Task 955, "Will Son Sim play sports motion games between 12pm and 10pm?"}

With No Hints, every model answers No in every condition. The Partial Hint ("day of the week + an action from 6am to 12pm") is too vague for the L-N setting and the models latch onto the wrong variable. Gemma, for example, links the activity to morning pushups or sports gaming, when the relevant variable is the \emph{absence} of morning family motion gaming. The Full Hint ("Thu/Fri/Sat + motion games 6am–12pm") names the right variable, and every L-N model, along with Gemini-2.5 and Gemini-3 under L, finds the rule. Under V+L, however, no model succeeds with the Full Hint, because the diaries almost never record the three afternoon sessions, meaning there is not enough evidence to apply the rule. Gemini-3's diary contains the only plausible trace (Son exercising in the gaming pod on Day 4), and the model itself notes that this is the sole recorded instance in 17 days.






\subsection{Trajectory Probing}

Using the RQ2 results, we examine how the reasoning of the four representative models evolves as evidence accumulates.

\paragraph{General patterns.} Models typically answer No on Day 1, reflecting the lack of evidence for a recurring pattern. On Day 2, Gemma, Qwen, and Gemini-3 often switch to Yes based on the previous day alone. These predictions are usually incorrect and suggest an overly aggressive attempt to infer a rule from very limited evidence. During the first week, models entertain multiple hypotheses, often  inferred from the specific wording of the hint (e.g., ``the day before'' or ``a specific date'' prompts every-other-day or date-of-month rules). Once a weekday repeats, reasoning shifts to comparing the current weekday with past occurrences, which lets models identify static patterns.




\paragraph{Failure modes.} We observe three main types of error:
\begin{itemize}
    \item \textbf{Evidence corruption:} inventing observations, missing ones that are present, or attributing another family member's activity to the target Sim.
    \item \textbf{Temporal grounding:} Qwen frequently treats ``today'' as the previous day, so its predictions track the wrong day. Consistent with this, its second-highest Spearman correlation is with the last-day baseline.
    \item \textbf{Rule updating:} On dynamic patterns, most models keep applying the original rule after behavior changes. Only Gemini-3-flash-preview ever cites the dialogue clue as evidence of a change.
\end{itemize}
Models can thus accumulate evidence for a recurring pattern but struggle to revise an established rule when new contextual evidence indicates a change. We examine the examples in Figure~\ref{fig:trajectory_probing_curves} in detail below.

\paragraph{N0: Mother Sim paints every Tuesday.}
Mother paints on Days 4, 11, 18, and 25. As expected, no model predicts Day 4, and all four identify the Tuesday pattern around the second occurrence. The main failure is evidence attribution:
\begin{itemize}
    \item \textbf{Gemini-2.5-flash} finds the pattern after Day 11 but but its reasoning contains several hallucinated or misattributed observations. For example, it incorrectly attributes Father's Day 2 painting to Mother and invents a Day 6 session. It then reverts to searching the current day's log instead of applying the pattern, missing Days 18 and 25, whose logs contain only headers.
    \item \textbf{Gemini-3-flash-preview} finds the pattern by Days 11--12 and applies it through the end, but invokes unsupported priors (e.g., that Tuesday and Friday are commonly paired for hobbies) and attributes the children's painting on Days 15 and 22 to Mother, concluding she paints on both Saturdays and Tuesdays.
    \item \textbf{Gemma-4-26B-A4B-it} initially relies on priors (e.g., predicting painting on Day 1 because it is a weekend), but identifies and maintains the Tuesday pattern from Day 10.
    \item \textbf{Qwen3.5-35B-A3B} attributes Father's painting to Mother and predicts No on Day 11, but corrects itself after the second occurrence.
\end{itemize}

\paragraph{N7: Daughter Sim's motion-game schedule changes.}
For the first 14 days, Daughter plays motion games Monday through Thursday (Days 1--3, 7--10, and 14). A dialogue clue on Day 15 signals a change, after which she plays only on weekends (Days 19, 20, 26, and 27).
\begin{itemize}
    \item \textbf{Gemini-2.5-flash} initially answers No for lack of evidence and later finds a weekday pattern, but focuses heavily on the current day's log, treating an empty log as evidence that the activity did not occur.
    \item \textbf{Gemini-3-flash-preview} is the only model to use the dialogue clue to recognize the change, correctly reasoning that the schedule has changed. After Day 25, however, it begins proposing unsupported alternatives (e.g., alternating schedules) and misattributing evidence from other family members.
    \item \textbf{Gemma-4-26B-A4B-it} relies on the hint rather than the observed evidence and frequently invents implications from it. It also defaults to Yes and describes Daughter as playing ``almost every day,'', while never detecting the shift.
    \item \textbf{Qwen3.5-35B-A3B} treats ``today'' as the previous day on 26 of 29 days, basing predictions on the wrong day.
\end{itemize}
No model consistently identifies and applies both phases. Models either extrapolate the initial pattern, follow the hint over the evidence, or misread the temporal reference.

\section{Vision-based Baseline Implementation Details}
\label{sec:appendix:vision_implementation_details}

For models with a vision input channel (Vision + Language and Vision + Audio in Table~\ref{tab:main_results}), we use a two-phase pipeline for long-video multiple-choice question answering. Each in-game day (spanning 6:00 AM–10:00 PM game time) is divided into four non-overlapping temporal windows of equal game-time duration. In Phase 1, a vision-language model (VLM) generates a natural-language diary summary for each window from sampled frames. In Phase 2, a text-only model answers the question by reasoning over the concatenated diary summaries. No video frames are shown to the model during Phase 2. 

We use vllm 0.19.0 with 1\textasciitilde4 Nvidia A40 GPUs for inference task. It takes between several hours to 7 days to fully run the evaluation for one model. 

\subsection{Temporal Window Definition}

Each simulated day is segmented into four four-hour game-time windows as shown in Table~\ref{tab:game_time_phases}.

\begin{table}[!ht]
\centering
\small
\begin{tabular}{clc}
\toprule
\textbf{Key} & \textbf{Phase Label} & \textbf{Game-Time Range} \\
\midrule
w1 & morning   & 06:00--10:00 \\
w2 & midday    & 10:00--14:00 \\
w3 & afternoon & 14:00--18:00 \\
w4 & evening   & 18:00--22:00 \\
\bottomrule
\end{tabular}
\caption{Game-time phase definitions.}
\label{tab:game_time_phases}
\end{table}

Window boundaries in real-video time are recovered by linear interpolation over a set of (video\_time\_sec, game\_time) anchor pairs extracted from the simulation's event log for each recording. For tasks whose question stop-point falls within a window (i.e., tasks that are cut at a partial day), the cut window is re-summarized independently from frames up to the stop time; windows beyond the stop point are omitted entirely.

\subsection{Phase 1: Per-Window Diary Generation}

\paragraph{Frame extraction.} Frames are sampled at 0.5 fps from the video-time interval corresponding to each window and resized to 224×224 pixels (JPEG quality 85). The resulting JPEG-encoded frames are passed as base64-encoded image\_url entries in the prompt. At 0.5 fps, a four-hour window spanning approximately six minutes of real video time yields around 180 frames, though in practice the pipeline caps this at the frames available in the annotated window interval.

\paragraph{Input composition.} Each Phase-1 call receives:

\begin{itemize}[leftmargin=*, itemsep=3pt, parsep=0pt, topsep=0pt, partopsep=3pt]
    \item A 2×2 cast image showing the four household members' faces (Father Sim, Mother Sim, Son Sim, Daughter Sim), used to ground character identity.
    \item The ordered sequence of sampled frames.
    \item Optionally, a dialogue transcript (in-game HH:MM timestamps) for tasks that require dialogue.
    \item A text trailer specifying the window label, day number, day-of-week, and date.
\end{itemize}

\paragraph{System prompt design.} The model is framed as the observer whose view constitutes the recorded footage, and is asked to narrate its own movements as "I." Family members are referred to exclusively by role name; no personal names are permitted. The prompt mandates two output sections in strict order: [Detailed actions], a moment-to-moment action list (one fragment per line, $\leq$12 words each, action-verb required), and [Window summary], a prose paragraph not exceeding approximately 250 tokens ($\sim$180 words). Fragments in [Detailed actions] must describe what actors do, not what objects or scenery are present. Good examples include "Mother Sim: stirs pot on stove" and "I: open the fridge"; disallowed examples include "I: see cabinets, bar stools, sink area" and "Father Sim: not visible in this segment." Only the [Window summary] section is carried forward to Phase 2.

\paragraph{Sampling parameters.}
Four windows are processed concurrently per video day. The model parameters are described in Table \ref{tab:phase1_decoding}.

\begin{table}[h]
\centering
\small
\begin{tabular}{lc}
\toprule
\textbf{Parameter} & \textbf{Value} \\
\midrule
Temperature & 0.7 \\
Top-p & 0.9 \\
Max output tokens & 800 \\
Frequency penalty & 0.4 \\
Repetition penalty & 1.1 \\
\bottomrule
\end{tabular}
\caption{Phase-1 decoding hyperparameters. Repetition penalty is applied via vLLM's \texttt{extra\_body} parameter.}
\label{tab:phase1_decoding}
\end{table}

\paragraph{Retry strategy.} If the model's output fails, and it lacks the [Window summary] section header — typically because the [Detailed actions] list exceeded the token budget — the call is retried up to two additional times (three total tries). On retry, a brief directive is appended to the end of the user message instructing the model to limit [Detailed actions] to 8–12 lines of at most 8 words each, and to emit the [Window summary] header immediately afterward. Retry delays follow a [0, 10, 30] second schedule.


\subsection{Phase 2: Text-Only QA over Diary Summaries}

\paragraph{Input construction.} The Phase-2 prompt presents the concatenated [Window summary] blocks for all days in the video chain, each prefixed with a structured day header (e.g., \texttt{Day 3 (Wednesday, 2024-03-06)}), followed by per-window labels (\texttt{[morning]}, \texttt{[midday]}, etc.). A calendar preamble listing all days with their dates precedes the diary text. The multiple-choice question and options follow the diaries. No images are included.

\paragraph{System prompt.} The model is instructed to base its answer exclusively on the provided summaries, to pay attention to day boundaries and recurring patterns, and to produce a response in the format:

\begin{tcolorbox}[
    width=\linewidth,
    colback=gray!3!white,
    colframe=gray!55!black,
    fonttitle=\bfseries,
]
\small\ttfamily
\textbf{Thought}: [reasoning, max four sentences]

\textbf{Answer}: [letter]
\end{tcolorbox}



\paragraph{Sampling parameters.} The model parameters are described in Table \ref{tab:phase2_decoding}. The max output tokens is set to be sufficient to cover four sentences of reasoning plus the answer letter.

\begin{table}[h]
\centering
\small
\begin{tabular}{lc}
\toprule
\textbf{Parameter} & \textbf{Value} \\
\midrule
Temperature & 0.7 \\
Top-p & 0.9 \\
Max output tokens (non-thinking) & 512 \\
Max output tokens (thinking enabled) & 8704 \\
Frequency penalty & 0.4 \\
Repetition penalty & 1.1 \\
\bottomrule
\end{tabular}
\caption{Phase-2 decoding hyperparameters.}
\label{tab:phase2_decoding}
\end{table}


\paragraph{Thinking budget (Qwen3/Gemma-4 variants).} For models supporting a hybrid thinking mode (e.g., Qwen3-VL Instruct, Qwen3.5, Gemma-4-IT), Phase 1 disables extended thinking (enable\_thinking=False) to minimize output verbosity during summarization. Phase 2 enables it with a budget of 8,192 thinking tokens per question, allowing the model to reason over the full chain diary before committing to an answer.



\subsection{Multi-Day Chain Handling}
Each evaluation task specifies an ordered chain of video recordings spanning one or more simulated days, along with a stop\_day\_position and a global stop time in seconds marking the task's temporal scope. The pipeline assembles the diary text as follows: for days before the stop day, all four cached [Window summary] blocks are included. For the stop day, windows whose entire interval precedes the stop time are included in full; the window that straddles the stop time is re-summarized using only frames up to the stop time; and subsequent windows are excluded. This ensures the model is never exposed to events beyond the question's designated horizon.



\subsection{Agentic Baselines}

\paragraph{Adapting M3-Agent to \OursBench{}.}
M3-Agent~\citep{long2026seeing} handles long videos by separating memory construction from question answering. Instead of reasoning over the full video directly, it first converts the video into an explicit memory graph. To build this graph, M3-Agent splits the video into short clips, detects faces and speakers with existing tools, transcribes speech, and uses a fine-tuned Qwen2.5-Omni-7B model to generate structured descriptions for each clip. These descriptions use entity placeholders for people and voices, which are linked across clips to form a graph of multimodal memories. At test time, a fine-tuned Qwen3-32B model serves as the reasoning module: it repeatedly retrieves relevant memories from the graph using natural-language queries and answers questions based on the retrieved information.

In our SimLife adaptation, we do not train or fine-tune any M3-Agent models. Instead, we directly use the released M3-Agent models for inference and only modify the memory graph construction process. SimLife consists of reusable Sims video units, and each QA task combines multiple units into a longer chain. Processing every full chain from scratch would therefore be redundant. We first precompute clip-level memories for each reusable unit, including face clusters, speaker information, and Qwen2.5-Omni captions. For each QA task, we then assemble the memory graph by replaying the cached unit memories in the order specified by the task chain, mapping local entity placeholders to global identities, and constructing two graphs: one with audio information and one without. During evaluation, we select the appropriate graph according to whether audio is available in the task setting.

We make three lightweight SimLife-specific changes to improve reliability. First, instead of using ASR-generated transcripts, we use the clean dialogue text provided by SimLife when constructing textual memories. The input remains the original MP4 video, so the multimodal model can still perceive visual content and ambient audio, while the memory graph receives accurate transcript text. Second, we link the four main characters, Father Sim, Mother Sim, Daughter Sim, and Son Sim, to fixed global face identities using reference avatar embeddings. This improves identity consistency across reused video units. Third, we add the date and weekday from the chain calendar to each memory and store them as metadata, enabling more reliable retrieval for time-related questions.

\paragraph{Adapting VideoTree to \OursBench{}.}

VideoTree~\citep{Wang_2025_CVPR} performs long video understanding by building a query-adaptive, hierarchical video representation to aid LLM reasoning in question answering. Rather than uniformly sampling frames, it performs adaptive breadth expansion to identify representative keyframes and then uses relevance-guided depth expansion to further explore the hierarchical structure to yield a tree-based representation of the video where the leaves are a question-focused subset of frames.

In our SimLife adaptation, we precompute frame-level EVA-CLIP-8B features once per reusable video unit and cache them on disk. When constructing a task, we assemble the full feature tensor by concatenating cached unit features in the order specified by the task chain, rather than re-encoding the same video multiple times. Caption resolution is also lazy: instead of loading all captions upfront, we store a narration manifest per chain that records each unit's caption path, global time offset, and calendar metadata (day index, day of week, and date). During tree construction and QA, only the captions for the sparse set of selected tree frames are loaded from disk.

We make a few SimLife-specific changes to the prompts given to the LLM at the breadth-expansion stage and the final question answering stage. 
First, we inject day and wall-clock time annotations into the narration: each caption line is prefixed with its simulated time (mapped from video-local seconds to the 06:00–22:00 simulation day) and grouped under a day header showing the date and weekday. This provides the LLM with explicit temporal grounding for time-sensitive questions. 
For tasks in the audio-available setting, we augment the breadth-expansion prompt with dialogue, inserting speech events that fall within a five-second window of each sampled frame. This allows the model to incorporate conversational context when scoring frame relevance, without modifying the tree-construction algorithm itself.
Finally, VideoTree's original prompts are designed for five-option multiple-choice questions, so we use  \Ours{}-specific prompt templates that format questions with two options (A and B) while preserving the relevance-scoring mechanism needed for breadth expansion. 

\section{Text-Only Baseline Implementation Details}
\label{sec:appendix:text_implementation_details}

We evaluate two families of text-only baselines that receive no video frames — only the structured activity logs extracted during post-processing. These baselines serve two distinct roles: Text-Only (Noisy) measures how much can be inferred from the raw, automatically extracted log stream, and Text-Only (Noiseless) provides an oracle upper bound by supplying only the task-relevant subset of that stream. Both families are evaluated matching the hint variants used in all other baselines to enable direct comparison.

We use vllm 0.19.0 with 1\textasciitilde4 Nvidia A40 GPUs for inference task. It takes between several hours to 7 days to fully run the evaluation for one model. 

\subsection{Input Representation}

\paragraph{Log format.} Activity logs are pre-materialized as plain-text files, one per simulated day. Each line encodes either a character action or a dialogue utterance with its in-game timestamp in HHMM format:

\begin{tcolorbox}[
    width=\linewidth,
    colback=gray!3!white,
    colframe=gray!55!black,
    fonttitle=\bfseries,
]
\small\ttfamily
Action: HHMM Participant1, Participant2: [(category)] action - target

Dialogue: HHMM Speaker (to Others): ``utterance''
\end{tcolorbox}



Lines within a day are sorted first by in-game timestamp, then by original insertion order to break ties. Day sections are delimited by a header of the form \texttt{=== Day N (day\_of\_week, date) ===}, allowing the model to track the calendar structure of the simulation.

\paragraph{Noisy baseline concatenation.} For each task, per-day log files are concatenated in chronological order up to and including the day designated by \texttt{stop\_day\_position}. Two log variants are used depending on the task type: an actions-only file (omitting dialogue) for standard tasks, and a file containing both actions and overlay-side dialogue for tasks that require audio–visual reasoning. For the final day in a chain, if the task specifies a mid-video stopping point (\texttt{stop\_video\_time\_local}), the last day's text is rebuilt from the underlying log file by filtering events whose video timestamp precedes the cut.

When the resulting prompt exceeds the model's input budget, the earliest days are dropped until the truncated log fits. The retained suffix is the most temporally proximate context, which is generally most relevant to questions about behavior up to \texttt{stop\_day\_position.}

\paragraph{Noiseless baseline.} The noiseless variant loads pre-materialized, oracle-filtered log files. Each file contains three sections — one per hint variant — and each section retains only: (i) day-boundary headers, (ii) the condition-side rows that the phenomenon predicate depends on, and (iii) occurrences of the target activity on fully observed prior days. 

\subsection{Prompt Structure}

Both baselines use an identical two-turn prompt structure (system + user). The system message instructs the model to answer only from the provided logs, to attend to day boundaries and day-of-week patterns, and to emit a strictly formatted response:

\begin{tcolorbox}[
    width=\linewidth,
    colback=gray!3!white,
    colframe=gray!55!black,
    fonttitle=\bfseries,
]
\small\ttfamily

Your entire response MUST consist of exactly two lines:

Line 1: Thought: <at most 4 short sentences of reasoning>

Line 2: Answer: <a single capital letter A, B, C, or D>
\end{tcolorbox}




The noiseless system prompt additionally notes that the input is a condensed observation history produced by oracle filtering, and that information beyond the observation cutoff should not be assumed.

The user message presents the concatenated log (or condensed history), followed by the question and lettered answer options.




\subsection{Retry Policy}

Each variant call allows up to three attempts. Attempts beyond the first append a tightening hint to the system message, which: (i) reminds the model that its previous attempt exhausted the token budget without emitting an answer, (ii) caps the Thought to three sentences, (iii) forbids nested counterfactual reasoning and day-by-day re-enumeration — the specific failure modes observed during development on reasoning-capable models. Retry delays are 0 s, 5 s, and 15 s for attempts 1, 2, and 3 respectively. The final attempt's output is persisted regardless of success; an unparseable response is scored as incorrect.

\end{document}